\documentclass[10pt]{article}
\usepackage[preprint]{tmlr}

\usepackage{amsmath,amssymb}
\usepackage{graphicx}
\usepackage{booktabs}
\usepackage{array}
\usepackage{longtable}
\usepackage[hidelinks]{hyperref}
\hypersetup{pdfauthor={Joyjeet Singh},pdftitle={Right Ranking, Wrong Scale: Reproducing the Robotic World Model's Dynamics Model and Testing the Uncertainty of Its Follow-up}}
\usepackage{url}
\usepackage{textcomp}
\def\month{MM}
\def\year{YYYY}
\def\openreview{\url{https://openreview.net/forum?id=XXXX}}
\title{Right Ranking, Wrong Scale: Reproducing the Robotic World Model's Dynamics Model and Testing the Uncertainty of Its Follow-up}
\author{Joyjeet Singh\\Independent Researcher\\Chandigarh, India\\\texttt{joyjeetsingh1@gmail.com}}
\date{}
\begin{document}
\maketitle

\begin{abstract}
We reimplement the dynamics model of the Robotic World Model (RWM) and test the uncertainty its follow-up, RWM-U, uses as a reward penalty in offline policy optimisation. Our rebuild matches the release's gradients. RWM's claim that training on its own rollouts, not one-step targets, lowers long-horizon error holds on held-out data: holding each of ten episodes out once, at the original's training length, it wins on every one at the pre-registered horizon; on short windows at one step it loses. RWM-U's ensemble disagreement ranks errors right but sizes them wrong: it ranks them better than the forecast step (Spearman +0.90 at a hundred steps), yet in every ensemble measured, error exceeds what the spread implies even at one step. The release feeds every member the ensemble mean, so spread does not track accumulated error: at 100 steps on the released checkpoint one-sigma coverage is 4.6\% against 68\%, and rolling each member on its own predictions raises it to 33\%. Each member's variance collapses: the implemented loss, squared error on a sampled prediction, is minimised at zero. Rescaling the released checkpoint's spread per horizon (fitted on nine episodes, scored on the tenth) restores one-sigma coverage, not two-sigma. We fix two pipeline defects. One robot, gait and dataset; no policy is trained.
\end{abstract}

\section{Introduction}

Offline model-based reinforcement learning trains a robot's policy inside a learned simulator. The Robotic World Model (RWM; Li, Krause and Hutter) learns such a simulator for a quadruped from recorded robot data, and its follow-up, RWM-U, adds an ensemble of predictors whose disagreement penalises the reward, in the manner of MOPO (Yu et al., 2020). A policy can be trusted inside the simulator only as far as the simulator's errors are known, so the penalty has two jobs: to rank where the model is more likely to be wrong, and to say by how much. We test whether it does either, on the released code and checkpoint and on models we rebuilt from the papers' descriptions.

\textbf{What we found, in plain words.} The rebuild matches the release's gradients before any training, so later differences belong to the method and not to our wiring (\S{}3.1). RWM's central claim, that training on the model's own rollouts lowers long-horizon error more than training on one-step targets (teacher forcing), holds on episodes the models never saw: with each of ten episodes held out once, rollout training wins on every episode at the pre-registered horizon, and on short windows at one step it loses (\S{}4.1, Figure 1a). The other two claims we tested get only weak tests (\S{}4.2, \S{}4.3).

\begin{figure}[!ht]\centering\includegraphics[width=\linewidth]{\detokenize{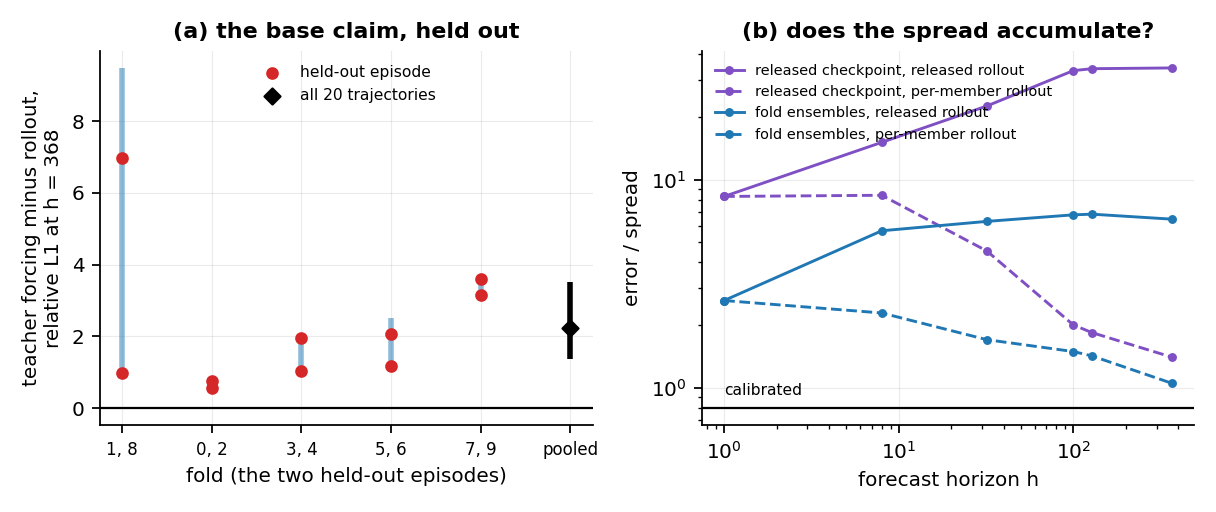}}\caption{(a) Our models, 2,500 iterations: teacher forcing minus rollout, relative L1 at h = 368, by held-out episode (20 trajectories; episode-level exact p 0.0020). (b) Error over spread, released checkpoint and fold ensembles (20 trajectories each); calibrated 0.80; dashed: per-member rollout, not released.}\end{figure}

RWM-U's ensemble disagreement ranks errors correctly and sizes them wrongly, in three separate ways. Even at one step, every ensemble's error exceeds what its spread implies (\S{}5.2). The five members share one trunk, so they can disagree only about what that trunk carries (\S{}5.3). And the release feeds every member the ensemble mean, so the spread does not grow with the horizon as the error does; rolling each member on its own predictions, which is not the released method, partly repairs this (\S{}5.4, Figure 1b). Separately, the per-member variance the method discards collapses because the implemented loss is minimised at zero (\S{}5.5). The spread ranks error better than the forecast step does (\S{}5.6), and rescaling it per horizon restores one-sigma coverage, though not two-sigma, on the released checkpoint; on episodes a model never saw the evidence is mixed (\S{}5.7). We also fix two defects in the released pipeline (\S{}6). Everything is measured on one robot, one gait and one dataset, and no policy is trained, so what the miscalibration costs a policy is not tested (\S{}7).

Figure 1 shows one finding about each original paper. Panel (a) is the base paper's claim that training on rollouts beats teacher forcing, measured on episodes the models never saw: the gap at h = 368, one interval per fold and one point per held-out episode. Panel (b) asks whether the follow-up's spread grows with the horizon as the error does, under the released rollout and under a per-member rollout that is not the released method.

\textbf{Contributions.}

\begin{itemize}
\item \textbf{RWM's claims, tested.} Holding each of the ten episodes out once, rollout training beats teacher forcing at 368 steps on every episode and loses at one step on short windows (\S{}4.1). The other two claims get weak tests (\S{}4.2 and \S{}4.3); two pipeline defects get fixes (\S{}6).
\item \textbf{Three gaps between RWM-U's spread and its error.} Even at one step every ensemble's error exceeds its spread; members sharing one trunk can disagree only about what that trunk carries; feeding them the ensemble mean stops the spread accumulating, which per-member rollout partly repairs (\S{}5.2 to \S{}5.4).
\item \textbf{Why the discarded per-member variance collapses.} The implemented loss is squared error on a sampled prediction, not the Gaussian likelihood, so its optimum is $\sigma$ = 0: derived, shown against known synthetic noise, and approached at an ordinary Adam step per iteration (\S{}5.5).
\item \textbf{What the spread is still good for.} It ranks error better than the forecast step (Spearman). It is also ahead of the model's step size on shorter units at the method's horizon, by a pre-registered rule, and under exploratory ranks on whole trajectories, where the pre-registered linear test could not resolve the margin (\S{}5.6). On the released checkpoint, per-horizon rescaling fixes one-sigma coverage, not the tails; on episodes a model never saw the evidence is mixed (\S{}5.7).
\end{itemize}

\S{}2 places the work; \S{}3 covers the data, metrics and statistics; \S{}4 and \S{}5 test the two papers; \S{}6 describes the pipeline defects; \S{}7 gives lessons and limitations; \S{}8 concludes and states what reproduces.

\section{Background and related work}

\textbf{RWM and RWM-U.} RWM (Li, Krause and Hutter) trains a recurrent dynamics model on a quadruped's proprioceptive data and uses it as a simulator for policy learning. RWM-U's ensemble of five predictors shares one state trunk, as its authors state, and penalises the reward by the members' disagreement. The penalty follows MOPO (Yu et al., 2020), which penalises the reward by an ensemble uncertainty estimate, and MOReL (Kidambi et al., 2020), which detects unknown states by pairwise ensemble disagreement; both branch short rollouts from real states, as MBPO does (Janner et al., 2019). As we read MOPO, the penalty is meant to bound the model's one-step error at the states the policy visits. That is the use against which \S{}5.2 measures the spread.

\textbf{Multi-step training.} Training a dynamics model on its own multi-step predictions rather than on one-step targets is an established remedy for compounding error; Venkatraman et al. (2015), Bengio et al. (2015) for sequences, Talvitie (2017) and Lambert et al. (2021) are examples. That sets the prior for \S{}4: a long-horizon advantage for rollout training is what these methods aim at, so the direction of RWM's claim is expected, and the test is whether it holds on held-out episodes of this dataset and what happens at one step.

\textbf{Ensembles that share a trunk.} Deep ensembles are trained from different initialisations and data orderings (Lakshminarayanan et al., 2017). Reinforcement-learning ensembles with a shared torso include Bootstrapped DQN (Osband et al., 2016) and randomised prior functions (Osband et al., 2018). That heads sharing a trunk under-report disagreement is established (Lee et al. (2015), Fort, Hu and Lakshminarayanan (2019), Wen et al. (2020) and Havasi et al. (2021)). RWM-U documents its shared trunk; \S{}5.3 asks what the sharing costs.

\textbf{Calibration, recalibration and conformal prediction.} Lu et al. (2022) compare uncertainty heuristics for offline model-based reinforcement learning under a protocol built "to capture the specific covariate shift induced by model-based RL", and report rank and bivariate correlation with true model error separately. That is our distinction between ranking and scale, on the same family of penalties, and the closest precedent. We add coverage measurements on a released checkpoint. Kuleshov et al. (2018) recalibrate regression uncertainty, and Malik et al. (2019) do so for a dynamics model inside model-based reinforcement learning, arguing that "good uncertainties must be calibrated". Our per-horizon rescaling (\S{}5.7) fits a scale on some held-out data and scores it on other held-out data, which resembles split conformal prediction (Angelopoulos and Bates (2021), Stankevičiūtė et al. (2021), Lindemann et al. (2023)); we make no claim of the conformal guarantee.

\textbf{Evaluation statistics.} Henderson et al. (2018) and Agarwal et al. (2021) are the reference points for reporting reinforcement-learning results from few runs. Our intervals rest on few seeds and few trajectories, so \S{}3.4 reports the count beside every interval.

\textbf{The PETS lineage.} The bounded log-$\sigma$ head is inherited from PETS (Chua et al., 2018); Appendix J asks which public descendants inherit its objective as well. Seitzer et al. (2022) show failure modes of $\sigma$ heads trained by maximising likelihood; the released state loss is not a likelihood at all (\S{}5.5).

\section{Setup}

\subsection{The reimplementation}

We rebuilt the dynamics model from the papers' descriptions and the released module: a recurrent (GRU) ensemble whose members predict the next proprioceptive state with a mean head and a bounded log-$\sigma$ head, plus auxiliary heads for contact and termination. The paper describes two loss terms; the implementation has 7. Before any training, our outputs match the released module's bitwise, and our losses and gradients match exactly (a maximum difference of 0.000e+00) across those 7 terms and 106 parameter tensors (Appendix A). A later discrepancy therefore belongs to the method or the data, not to our wiring.

\subsection{Data, arenas and independent units}

The released dataset is 10,000 rows of ANYmal D proprioceptive state and policy actions at 50 Hz, 0.13\% of the data budget the reference trains on: ten concatenated 20-second episodes of 1,000 rows (the first has 999), with a termination column that is identically zero. We recover the episode boundaries from the rows at which every joint velocity and action is zero (supplement S16). Table 1 lists the arenas we measure on, each with its count of mutually non-overlapping 400-step trajectories, which is the sample size.

\begin{center}\small\resizebox{\ifdim\width>\linewidth\linewidth\else\width\fi}{!}{%
\begin{tabular}{llll}
\toprule
Arena & Episodes & Independent 400-step trajectories & Used for \\
\midrule
Held-out pair & the two episodes withheld when training our models & 4 & our models' out-of-sample claims \\
Training episodes & the other eight & 16 & our models' in-sample readings \\
All ten episodes & the whole release & 20 & the released checkpoint \\
Fold held-out pairs & five folds, each holding out two episodes, each episode once & 20 (4 per fold) & the fold replication and the fold ensembles \\
\bottomrule\end{tabular}}\end{center}

\emph{Table 1: the arenas; our models are the 2,500-iteration arms. The held-out pair's 4 trajectories come from two episodes.}

Two 400-step trajectories whose spans overlap are not independent evidence, so we count only non-overlapping ones, report the count beside every interval, and resample whole trajectories (\S{}3.4). A bootstrap over 4 trajectories has 256 distinct resamples, so its intervals are coarse, and that count bounds every long-horizon claim about the held-out pair.

\textbf{The released checkpoint's training data is not recorded.} No file in either released repository records which rows trained the released checkpoint, and the checkpoint stores no data statistic, hash or path (its tagged iteration is 5,000). Its one-step error is the same on the eight episodes our own model trained on and the two it did not (0.083 against 0.085), where our model's differs (0.075 against 0.12). That is consistent with its having trained on all ten episodes, on some, or on none, and the data cannot tell which. We therefore call its arena its ten released episodes and never call it held out. Everything here ran on two CPU cores (Appendix B).

\subsection{Metrics and floors}

\textbf{Relative-L1} is the reference's own metric, which keeps our numbers comparable to the upstream's printed ones: at each forecast step, the summed absolute error over the state dimensions divided by the summed absolute true state, on normalised states, averaged over trajectories and over the steps after the first 32, which are teacher-forced. Its denominator can pass through zero, so we also report \textbf{normalised RMSE (nRMSE)}: the root of the squared error pooled over the state dimensions, divided by the dimensions' mean standard deviation over the training episodes. A value of one is not the training-mean floor, because that predictor scores the root-mean-square of the dimensions' standard deviations over their mean, which is at least one; we report three floors: predicting the training mean, holding the last state, and repeating a seasonal pattern (\S{}4.3; Appendix F defines them).

\textbf{Coverage at $\pm$k$\sigma$} is the fraction of (trajectory, step, dimension) triples whose absolute error is within k times the predicted $\sigma$; a calibrated Gaussian gives 68\% at $\pm$1$\sigma$ and 95\% at $\pm$2$\sigma$. \textbf{The error-to-spread ratio} $\rho$ is the mean absolute error over the mean predicted $\sigma$; a calibrated Gaussian has $\rho$ = 0.80, so a ratio of one is not calibration, and we report $\rho$ beside that value. \textbf{"Too small by k"} means one thing: the factor k on $\sigma$ that restores nominal $\pm$1$\sigma$ coverage. \textbf{Ranking statements} use Spearman's rank correlation, because a ranking claim is about order and a rank statistic does not depend on scale or on a linear relation; a rule committed with Pearson's correlation keeps it, and we say so (Appendix L gives Kendall's).

Curves are reported at horizons h = 1, 8, 32, 100, 128 and 368, each cumulative from the first forecast step to h. \textbf{h = 100} is the method's own imagination rollout length. \textbf{h = 368} is the upstream's open-loop diagnostic length, 400 minus the 32-step prefix, 3.7$\times$ longer than the method's rollout and not a deployment horizon.

\subsection{Statistical protocol, pre-registration and deviations}

Intervals are 95\% intervals from a cluster bootstrap over whole trajectories, the unit of dependence, as in the literature on inference with few clusters (Cameron et al., 2008); resampling seed $\times$ trajectory narrows every interval, as an earlier analysis of ours did before we corrected it (supplement S17). Comparisons at four trajectories use exact sign-flip tests. Their smallest attainable two-sided p is 1/8, above every Holm level we use, so no comparison at that size can be called resolved; each is reported with its exact p and as "k of the four". Multiplicity is controlled with Holm's procedure within each pre-registered family, and not across rules. Per-dimension evidence uses permutation nulls that keep the 45 state dimensions together, because they are not independent.

Decision rules were committed to git before the data that tested them, with timestamps a reader can check; Appendix E gives each rule, its lead time and its verdict verbatim. Pre-registration in machine learning has workshops of its own (Bertinetto et al. (2021), Albanie et al. (2022)); we commit each rule before its data so that the analysis cannot follow the result. One rule failed the discipline: the rule for the control that adds exact copies of windows reached git 2.9 hours after its runs finished, so we withdrew its claim to have been pre-registered. Several verdicts went against the original or could not settle the question, and we report each. The ledger keeps 24 superseded entries: nine claims withdrawn on evidence, eight framings withdrawn, and the rest early hypotheses closed as housekeeping. Analyses written after the data are labelled post hoc and never change a verdict. Where our setup departs from the original's or from a rule, Appendix O and Appendix E say how.

\section{The base paper's claims}

\subsection{Training on rollouts against teacher forcing}

RWM's central claim is that training the dynamics model on its own autoregressive rollouts beats training it on one-step targets (teacher forcing) at long forecast horizons. \S{}2 gives the prior: a long-horizon advantage is what multi-step training aims at, so the question is whether it holds on episodes the models never saw, and what happens at one step.

\textbf{Every episode held out once.} We trained both arms on five folds, fixed in advance, each holding out two episodes so that each of the ten episodes is held out exactly once (2,500 iterations, three seeds per arm per fold). Over the 20 independent out-of-sample trajectories of all ten episodes, teacher forcing's relative-L1 error at h = 368 exceeds rollout training's by +2.2 [+1.4, +3.5], a factor of 5.9: the direction replicates across folds. All 10 of the 10 episodes favour rollout training (two-sided sign-flip test over the episodes, p = 0.002), and each fold's own gap, from +0.66 to +4.0, is positive on each of its four trajectories (descriptive at four; Figure 1a). That test treats the ten episodes as exchangeable, although the models of different folds share most of their training episodes, so it is a statement about these ten episodes and this dataset, at its scale (\S{}3.2). On the held-out pair the teacher-forced arm is worse than the training-mean floor at this horizon (nRMSE 3.8 against 0.94) and the rollout-trained arm is better (0.54), so the factor compares a model that works with one that drifts beyond the floors (Appendix F).

\textbf{At one step the order reverses, on short windows.} On 300 non-overlapping 33-row units, the same arms differ by -0.0090 [-0.013, -0.0053] at h = 1, in favour of teacher forcing, which is unsurprising because it trains on exactly that target. At the episode level the sign agrees on only 6 of ten episodes (sign-flip p = 0.20), so the reversal is a short-window result and is not resolved episode by episode. A rule committed before its runs, on the held-out pair's 60 independent 33-row units, finds the same sign, -0.019 [-0.031, -0.0093] (Appendix G).

\textbf{Supporting evidence from longer training.} On the held-out pair, 4 trajectories from two episodes, three seeds trained for 10,000 iterations give a factor of 4.6 at h = 368, with an interval of [+0.70, +2.3] on the gap; all 4 trajectories agree in sign, and the exact p of 1/8 is the smallest attainable at four. A rule committed before those runs, on one seed per arm, found the advantage at h = 368; the three-seed extension was computed afterwards and carries none of its weight. The advantage grows with the horizon and is weakest inside the eight steps the model is trained on (Appendix G gives every horizon).

\subsection{The configuration claim}

The original varies the history length M and the forecast length N and says moderate values trade accuracy against training time, with (M, N) = (32, 8) as its instance. We tested only the accuracy half, one factor at a time, at 2,500 iterations on the held-out pair. At that budget two shorter histories and both longer training forecasts beat the centre, but with 4 trajectories no comparison can be significant after exact tests, and trained longer the centre passes each of them on some reading, so which setting is most accurate depends on the budget. The original's trade-off with training time is not tested, and at our budget the centre is not shown to be best and no alternative is shown to be better (Appendix H).

\subsection{The architecture claim: consistent, but a weak test}

The original says RWM beats an MLP, a recurrent state-space model (RSSM) and a transformer. We built all three to our reading of the original's table of shapes, gave them RWM's settings, and trained each both teacher-forced, as the original does, and autoregressively. Table 2 gives nRMSE at h = 368. Every baseline is worse than RWM on all four trajectories, but at four trajectories even that gives only the smallest attainable exact p, and no baseline beats even the training mean, which is the stronger of the three floors here. The result is consistent with the claim and a weak test of it: the baselines are untuned, their sizes and settings are our reading, and four trajectories cannot reject anything.

\begin{center}\small\resizebox{\ifdim\width>\linewidth\linewidth\else\width\fi}{!}{%
\begin{tabular}{llll}
\toprule
Model & nRMSE, h = 368 & worse than RWM on & exact p \\
\midrule
RWM (our rollout-trained arm) & 0.54 &  &  \\
MLP, teacher-forced † & 110 & 4 of 4 & 1/8 \\
MLP, rollout-trained & 1.4 & 4 of 4 & 1/8 \\
Transformer, teacher-forced † & 13 & 4 of 4 & 1/8 \\
Transformer, rollout-trained & 2.0 & 4 of 4 & 1/8 \\
RSSM, teacher-forced & 5.4 & 4 of 4 & 1/8 \\
RSSM, rollout-trained † & 9.3 & 4 of 4 & 1/8 \\
Floor: predict the training mean & 0.94 &  & not tested \\
Floor: hold the last state & 1.1 &  & not tested \\
Floor: repeat a seasonal pattern & 1.1 &  & not tested \\
\bottomrule\end{tabular}}\end{center}

\emph{Table 2: nRMSE at h = 368 on the held-out pair (4 independent 400-step trajectories from two episodes); our models at 2,500 iterations, every model with RWM's settings. † a diverged row, whose size is not meaningful; only its sign is. A value below the training-mean floor beats predicting the average.}

Matched to RWM's parameter count instead, the rollout-trained MLP is worse than RWM on 4 of 4 trajectories and the transformer on 2 of 4 (exact p 7/8), so the transformer cannot be told apart from RWM; that tests parameter count and nothing else. Our RSSM collapses when rolled out and is not an informative comparison; the failure may be ours rather than the architecture's.

\section{The follow-up's uncertainty}

\subsection{What RWM-U claims and what its code computes}

RWM-U makes three claims about its ensemble that we can test. It says the learned variance "captures aleatoric uncertainty effectively" (its \S{}4.1, unchanged across versions); that the disagreement across members "closely follows the trend of the prediction error" and so serves as a trust metric (its \S{}5.1); and, from its second version on, that uncertainty is "effectively propagated over long horizons".

The released code computes two quantities at each step: the mean across members of each member's predicted $\sigma$, which we call aleatoric, and the standard deviation across the members' mean predictions, which we call epistemic (\texttt{system\_dynamics.py:125-126}). Only the second reaches the policy. It is returned at \texttt{envs/\allowbreak{}base\allowbreak{}.py:158} and applied as a reward penalty with weight $-$1.0 at \texttt{:166}; the aleatoric term is bound to a variable that is never read again (\texttt{:142}). The paper's Eq. 4 specifies a variance where the code computes a standard deviation; the first author confirms that the code's quantity is the intended one and that the aleatoric term is reported for analysis only (supplement S5). We measure the code's quantity. RWM-U does not claim a calibrated interval, and a penalty weight can absorb a constant factor, so what we test is what the spread reports at one step and how that changes with the horizon (\S{}5.2, \S{}5.7). The five members share one recurrent trunk, as RWM-U's authors state, and \S{}5.3 measures what that costs.

As we read MOPO's penalty, its job is to bound the model's error at the states the policy visits, one step at a time. That is the job against which we judge the spread, and it separates three gaps that a single ratio of spread to error runs together: the one-step gap on held-out data (\S{}5.2); the gap between what an ensemble can and cannot see (\S{}5.3); and the gap between a one-step spread and accumulated error (\S{}5.4).

\subsection{At one step, the spread already falls short}

At one step nothing has accumulated: the input is the true state, and $\sigma$ is asked the question it was trained on. Table 3 gives, for each ensemble, the ratio R of its mean squared error to the squared error its spread predicts, the factor k on $\sigma$ that restores nominal coverage at h = 1 and at h = 100, and the $\pm$1$\sigma$ coverage at h = 100 when the members are rolled out on their own predictions (\S{}5.4).

\begin{center}\small\resizebox{\ifdim\width>\linewidth\linewidth\else\width\fi}{!}{%
\begin{tabular}{llllll}
\toprule
Ensemble & Arena (independent trajectories) & One-step R [95\% CI] & k at h = 1 & k at h = 100 & $\pm$1$\sigma$ coverage at h = 100, per-member rollout \\
\midrule
Released checkpoint & all ten released episodes (20) & 174 [82, 255] & 6.6 & 31 & 33\% \\
Shared-trunk ensembles of five, our arms & held-out pair (4) & 7.6 [4.0, 9.7] & 2.6 & 13 & 46\% \\
Shared-trunk ensembles of five, our arms & five folds (20) & not measured & not measured & not measured & 42\% \\
Independently trained members, five & held-out pair (4) & 2.6 [1.6, 2.9] & not measured & not measured & 60\% \\
Independently trained members, three & five folds (20) & not measured & not measured & not measured & 48\% \\
\bottomrule\end{tabular}}\end{center}

\emph{Table 3: our arms at 2,500 iterations, the released checkpoint as released. A cell the artifacts do not hold reads "not measured". Intervals at four trajectories are descriptive: the smallest exact p there is 1/8.}

Every interval for R lies above one: even at one step no ensemble's spread accounts for its error, and the independently trained members come closest. On the folds' 20 independent held-out trajectories, the shared-trunk ensembles' error-to-spread ratio $\rho$ at one step is 3.8, against 0.80 for a calibrated Gaussian. The released checkpoint's disagreement is 8.3$\times$ smaller than its error at h = 1, with 16\% coverage, and the per-member $\sigma$ the method discards is 1,827$\times$ smaller than that error; which rows trained it is not recorded (\S{}3.2), so that figure rests on all ten of its released episodes, 20 independent trajectories. R compares squared errors, so it is larger than the ratio of mean absolute error to spread and than the coverage-matching factor k; the three summarise one shortfall.

\textbf{Does measuring on training data flatter the result?} Not at one step. With every episode held out once and also scored in sample, the ratio of in-sample to held-out error-to-spread at h = 1 is 0.92 [0.81, 1.1]: not resolved in either direction. An earlier comparison on four held-out trajectories found it worse in sample, 1.8 [1.1, 2.7] (descriptive at four), and does not hold up at twenty. At h = 100 in-sample measurement flatters (0.65 [0.58, 0.73]). For each model the two arenas differ in episodes as well as in training status, so the reading depends on the horizon and not on a fixed sign.

\subsection{What an ensemble can see: the shared trunk}

Abbas et al. (2020) separate aleatoric noise, parameter uncertainty and model inadequacy. Spread across members estimates at most the second; to the extent that one-step error is model inadequacy, no spread across members estimates it. That bears on \S{}5.2 before any trunk is considered.

The trunk is shared. The state pathway is one recurrent trunk plus five small heads, and per member 89\% of the parameters are numerically identical to every other member's. An autoregressive rollout feeds the ensemble mean back into one recurrent state, so there is one hidden-state trajectory between them and the spread is that of five heads reading one vector. RWM-U states the sharing; what we add is its measured size and consequence. Such members have correlated errors by construction, so their spread cannot express uncertainty the trunk does not already carry. Randomised prior functions and Bootstrapped DQN are ensembles of the same kind (Osband et al. (2016), Osband et al. (2018)). RWM-U calls its members a bootstrap ensemble; per-head resampling exists in the online training path and not in the offline one our arms follow (Appendix O).

Training five independently initialised full models and using them as an ensemble improves the error-to-spread ratio $\rho$ at h = 100 by 2.0$\times$, against a minimum detectable effect (the smallest gain the comparison could reliably detect) of 1.5$\times$, on the held-out pair's 4 independent trajectories; on the folds' 20 trajectories the gain is 1.8$\times$ with 6.0 points more coverage. That bounds the sharing effect and does not isolate it: independent members also differ in capacity (3.5$\times$ the parameters) and in data order, and at matched capacity the gain, 1.8$\times$, is below what that comparison can resolve. Combining independence with the corrected loss adds nothing beyond independence alone (Appendix K). Neither ensemble is an interval.

\subsection{Does the spread accumulate?}

RWM-U's second version says uncertainty is "effectively propagated over long horizons". Testing that against accumulated error is legitimate. It is not a test of calibration: at depth the input to $\sigma$ is the model's own earlier output, wrong by an amount $\sigma$ never claimed to describe, so a growing error-to-spread ratio is not a failure of the penalty's one-step job (\S{}5.2).

The release feeds every member the ensemble mean through one recurrent state, so its spread at a deep step is a one-step spread at one shared history. We asked whether rolling each member out on its own predictions instead makes the spread track accumulated error. At h = 100 it closes a share 0.75 [0.69, 0.82] of the log gap between spread and error (the log of the error-to-spread ratio, compared before and after the change) on the released checkpoint (its ten released episodes, 20 trajectories), 0.72 [0.67, 0.76] on the shared-trunk ensembles of the five folds (20), and 0.71 [0.67, 0.76] on the independent ensembles of the folds. On the held-out pair's four trajectories the shared-trunk and independent ensembles close 0.80 and 0.88, descriptive only (exact p 1/8). At h = 100 the released checkpoint's $\pm$1$\sigma$ coverage rises from 4.6\% to 33\% and CRPS and NLL improve too, so the effect is not only a wider spread, but coverage stays below the 58--78\% band around the target, so it does not close the gap (Figure 1b).

What this settles: "effectively propagated" fails as the release rolls the ensemble out; most of the log gap closes when the members are rolled out separately, though coverage stays below the band. Per-member propagation is not the released method, and the release's own imagination rollout, which propagates one member per environment, is not tested.

\subsection{Why the per-member variance collapses}

The follow-up reads the low aleatoric value as reflecting small stochasticity in the environment. The observation is right and the reading is not. The implemented state loss is squared error on a sample from the predicted Gaussian, not a likelihood: its expectation is ($\mu$ $-$ y)² + $\sigma$², which is minimised at $\sigma$ = 0 for any $\mu$. Nothing in the loss opposes it. The upper bound is built as the lower bound plus a learned gap, so the lower bound receives no gradient from the bound term and moves only through the state loss. The objective's optimum is $\sigma$ = 0 per step; the fed-back sample is a second pathway we did not isolate, and the teacher-forced arm, which shares the loss, collapses at the same rate.

We tested it before training. On synthetic data whose known noise varies 25$\times$ over the input range, the released head under the implemented loss sits 22$\times$ below the true noise and does not track it, while under the authors' unused likelihood branch it recovers the level (ratio 0.98, averaged over seeds). Across the 28 runs at the released width that the fitted rate uses (Figure 2 draws the 75 logged runs of 2,500 iterations) the log-$\sigma$ range moves linearly in iteration count, falling under sampled MSE and rising under the likelihood (Figure 7, Appendix J).

\begin{figure}[!ht]\centering\includegraphics[width=\linewidth]{\detokenize{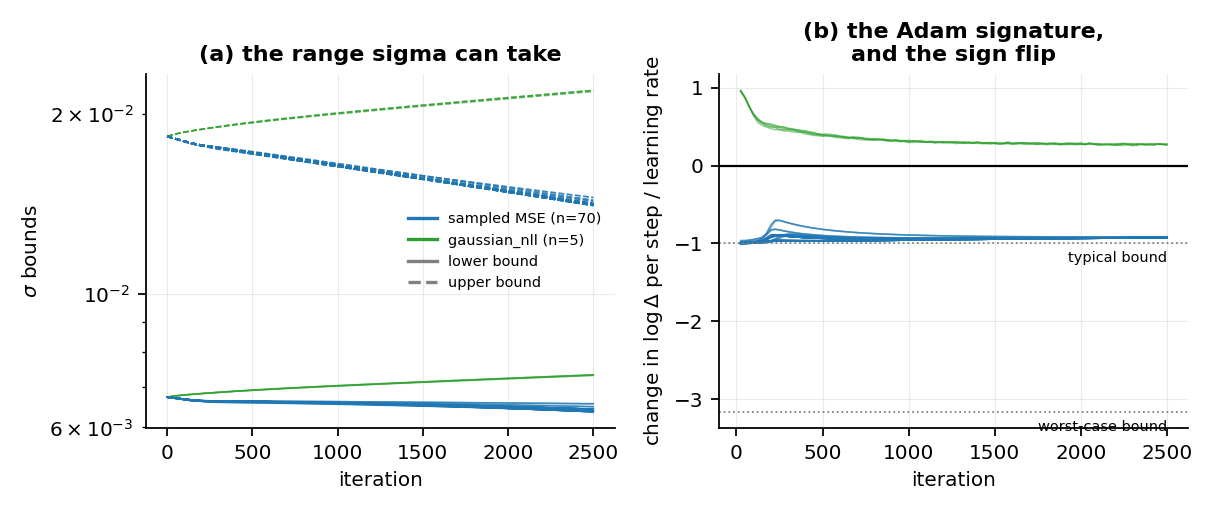}}\caption{The 75 collapse-record runs of 2,500 iterations. (a) The lower (solid) and upper (dashed) bound of $\sigma$ per logged iteration; 342 saved checkpoints reproduce them. (b) Change in $\log\Delta$ per iteration over the learning rate, with Adam's typical and worst-case bounds; the corrected objective flips the sign.}\end{figure}

Figure 2 shows $\sigma$ itself, and the rate is an Adam property. The collapse moves log $\Delta$, the log of the gap between the $\sigma$ bounds, by a median 0.94 learning rates per iteration over 81 runs, the ordinary step of Adam (Kingma and Ba, 2015), and only the sign flip speaks to the objective. Even at the largest step Adam can take (3.2$\times$ the learning rate), the checkpoint's variance state would need 45,737 iterations from the released initialisation, against the 2,500 we trained for. The bounded head is inherited from PETS (Chua et al., 2018); of 13 public repositories inspected, 10 implement it and 1 optionally trains it on sampled squared error (Appendix J).

\subsection{Ranking}

Disagreement ranks error well, and we report it with Spearman's rank correlation (\S{}3.3), taken between the scalar the method applies and total absolute error across the pooled (trajectory, step) points up to the horizon, and at h = 1 across trajectories. On the released checkpoint over all ten released episodes (20 independent trajectories), Spearman is +0.90 at h = 100, and its lead over the forecast step index, a counter that needs no ensemble, is resolved at every horizon from h = 8 on (at h = 1 the index is constant, so no comparison with it is defined); at h = 1 Pearson is the higher (+0.99 against +0.83). The rules committed before these readings used Pearson, and their verdicts keep it.

Does the ensemble beat something free? Against the model's own predicted step size, whole trajectories cannot settle the margin at h = 368 under Pearson (+0.14, below the 0.29 that sample can resolve), though under ranks the margin has an interval above zero (+0.31 [+0.24, +0.40]). A rule committed before its reading put the same comparison to 70 non-overlapping shorter units at h = 100 on the same checkpoint: the margin is +0.28 [+0.20, +0.39], resolved on this checkpoint's ten released episodes and not re-tested at h = 368. Disagreement keeps real information with step size partialled out, so "disagreement ranks error well" is supported, and "it ranks better than a free signal" holds on shorter units and under ranks but not under the pre-registered whole-trajectory test.

On a model we trained, a rule committed before its reading (disagreement against the index on the held-out pair, four trajectories) was inconclusive: the direction held at every horizon, but the paired interval excluded zero at 1 of 4 horizons, short of the majority it required. The same comparison on the folds' 20 trajectories, committed before its data, is resolved: disagreement leads at 4 of 4 horizons and the paired interval excludes zero at 4.

\subsection{Rescaling per horizon}

A constant factor could be absorbed by the penalty's weight, so the one-step shortfall alone does not show the penalty mis-sized; what no constant can absorb is that the factor grows with the horizon. One constant scalar does not repair the spread, because the ratio of error to spread grows: on the released checkpoint the coverage-matching factor k (\S{}3.3) is 6.6 at h = 1 and 42 at h = 368. A multiplier per horizon is a split-conformal construction (Angelopoulos and Bates, 2021): a $\sigma$-normalised score, one quantile per horizon, fitted on calibration units and scored on others. Its finite-sample guarantee needs exchangeable calibration units, which here means episodes and not steps, and then coverage lies between the target and the target plus one over the number of calibration units plus one, so a ten-point band needs at least nine calibration episodes. The units of the nine-episode fit are the released checkpoint's ten released episodes; if it trained on them, which the data cannot rule out, its guarantee holds for in-sample use only.

Fitted on nine episodes and scored on the tenth, the released checkpoint's $\pm$1$\sigma$ coverage over all ten released episodes' 20 independent trajectories is within the band at every horizon. $\pm$2$\sigma$ is not restored: at h = 100 it is 84\% [81, 86] against 95\%, because the errors are heavy-tailed (across all 48 cells of the tails analysis, 30 over-cover at $\pm$1$\sigma$ and under-cover at $\pm$2$\sigma$ once $\sigma$ is rescaled). CRPS improves, from 0.66 to 0.58 at h = 100. Rescaling fixes the scale and not the shape. An earlier reading that every cell is Gaussian was withdrawn (Appendices M and E).

On episodes the model never saw, the evidence is mixed. The multipliers of our rollout-trained models calibrate 17 of 36 cells (a cell is one multiplier scored on one held-out unit at one horizon) on the held-out pair's 4 independent trajectories, and on the five folds a rule committed before its data finds the per-horizon multiplier calibrates 72 of 180 cells and finds that it does not restore calibration, while the folds' pooled $\pm$1$\sigma$ coverage, reported beside it and not as a reading, is 67\% at h = 100, inside the ten-point band around the target. The verdict stands, and both measurements are the result: pooled across folds the multiplier lands near the target, cell by cell it often does not, and its guarantee is for the units that fitted it.

\section{Defects in the released pipeline}

Two defects in the released code and data bear on what precedes; Appendix N gives the rest, and Appendix O the places where the code departs from the paper's description.

\textbf{Unmarked episode boundaries.} The data file concatenates ten episodes with a termination column that is identically zero, so the window builder marks all 9,961 windows valid, including 352 that splice one episode's end onto the next one's start (\S{}3.2). We recover the boundaries from the reset rows and exclude those windows. Training on spliced windows costs nothing measurable in rollout: over 32 cells a contaminated arm is hurt in 0 and helped in 9, consistent with regularisation and not with harm. The defect remains a real one on leakage grounds (Appendix N).

\textbf{The action is one step stale in evaluation.} Row t holds the action that produced state t: at every reset row all twelve actions are exactly zero, which a policy network with biases cannot emit, and which is what Isaac Lab writes at a reset. Training pairs a state with the action in its own row; evaluation feeds the action from the previous row, one step stale. Shifting the two action slices of the evaluation loop (\texttt{model\_training.py:129} and \texttt{:132}) aligns it, and our harness does this with \texttt{action\_offset = 1} (\texttt{src/\allowbreak{}score\_\allowbreak{}reference\allowbreak{}.py:180-190}). On the released checkpoint's all ten released episodes (20 independent trajectories), the stale action raises relative-L1 error by 55\% [26, 93] at h = 1 and 37\% [21, 65] at h = 32; from h = 100 the change is not resolved (8.2\% [-10, 25] there). On the held-out pair's 4 trajectories it raises the error at every horizon reported. Our own models trained under the causal pairing are also hurt by the stale one (Appendix N). We sent this to the first author before submission and had no reply when we submitted.

\textbf{Evaluation on training data.} The released evaluation draws its trajectories from the data the checkpoint may have trained on, and no file records which rows it used (\S{}3.2), so no held-out measurement of the released checkpoint is possible. For our models we keep the two arenas apart throughout.

\textbf{The variance state.} The checkpoint's stored variance state cannot be reached from the released initialisation at a constant rate: even at the largest step Adam can take (Kingma and Ba, 2015), 45,737 iterations would be needed, against the iteration counts that the release, the paper and the checkpoint's tag give. The first author's account, that the release is several revisions removed from the setup that trained it, supplies a mechanism such as a warm start. This is a documentation gap between a release and a run, not an inconsistency (Appendix J). The first author also confirmed Eq. 4's intended quantity and the discard of the aleatoric term (supplement S5).

\section{Lessons and limitations}

\textbf{Lessons for a practitioner.}
\begin{itemize}
\item \emph{Use ensemble disagreement as a ranking signal, price it against free alternatives, and do not read it as a distance.} It ranks error well and better than the forecast step; on the released checkpoint it is also ahead of the model's own predicted step size on shorter units and under ranks, though the linear margin on whole trajectories is not resolved. Its size is wrong at every horizon, including one step (\S{}5.2, \S{}5.6).
\item \emph{If you need an interval, rescale per horizon and refit on your own model.} On the released checkpoint this restores one-sigma coverage and not two-sigma; on episodes a model never saw the evidence is mixed (\S{}5.7).
\item \emph{Roll ensemble members out separately if you want their spread to track accumulated error.} Feeding every member the ensemble mean does not (\S{}5.4).
\item \emph{Check that the implemented loss is the described loss.} Here the variance has an optimum at zero, which a reading of the code explained (\S{}5.5).
\item \emph{Count independent trajectories, report an exact p beside every interval at four, and permute whole trajectories.} Sign counts over coupled state dimensions are not independent trials, and an earlier draft of this paper made its weakest evidence look strongest that way (\S{}3.4).
\item \emph{Say which arena a calibration number comes from.} Measuring on training data neither flatters nor worsens one-step calibration resolvably, and flatters it at long horizons (\S{}5.2).
\item \emph{Anchor a decision rule to the horizon the claim is about.} Our first rule was anchored at the training horizon and pointed at the wrong regime (Appendix G).
\end{itemize}

\textbf{Limitations.} One robot, one gait and one flat terrain, and one dataset, so "generalisation" here means across velocity commands only. No policy is trained, so what the miscalibration costs RWM-U's policies is untested. The reason it may matter is that MOPO's penalty is meant to bound one-step error at visited states, and the spread falls short of the error already at one step (\S{}5.2). A test that needs no policy, of the penalty against the model's error in predicted return, could not be run because the released data do not record the velocity command its reward terms score against. The held-out pair is four trajectories from two episodes, so no comparison there can be significant; the fold replications reach twenty, but their models share most training episodes. The released checkpoint's training rows are not recorded. Our baselines are untuned readings of the original's table of shapes, and our data are a small fraction of the reference's. We apply no correction across our own 35 pre-registered rules, each reported against the threshold it was committed with. Whether other descendants of the PETS head inherit the $\sigma$ = 0 optimum is untested.

\section{Conclusion}

RWM's central training claim reproduces on held-out episodes at long horizons and reverses at one step on short windows; its other two claims get weak tests, and at h = 368 the baselines we built never beat even the training mean. RWM-U's ensemble disagreement ranks error correctly and sizes it wrongly: even at one step every ensemble's error exceeds its spread, members that share a trunk can disagree only about what it carries, and the release's rollout stops the spread accumulating, which rolling members out separately partly repairs. The per-member variance collapses because the implemented loss is minimised at zero. Rescaling per horizon restores one-sigma coverage on the released checkpoint, not two-sigma, and on unseen episodes the evidence is mixed. As a scalar penalty, disagreement is a ranking and not a scale, and a ranking use deserves validation on the deployment distribution. What should travel is the objective's property, which should be checked wherever a bounded $\sigma$ head is trained against sampled squared error; the released artifact's specific factors should be checked rather than assumed, and the limits of one robot's data are limits on what could be resolved, not findings in either direction.

\begingroup\small
\begin{longtable}{>{\raggedright\arraybackslash}p{\dimexpr 0.2503\linewidth-2\tabcolsep\relax}>{\raggedright\arraybackslash}p{\dimexpr 0.3105\linewidth-2\tabcolsep\relax}>{\raggedright\arraybackslash}p{\dimexpr 0.3391\linewidth-2\tabcolsep\relax}>{\raggedright\arraybackslash}p{\dimexpr 0.1001\linewidth-2\tabcolsep\relax}}
\toprule Claim & Evidence & Where it stands & Section \\ \midrule\endfirsthead
\toprule Claim & Evidence & Where it stands & Section \\ \midrule\endhead
\bottomrule\endlastfoot
RWM: training on rollouts beats teacher forcing at long horizons & 20 trajectories from all ten episodes, each held out once & holds at h = 368 on every episode; reverses at one step on short windows & \S{}4.1 \\
RWM: moderate history and forecast lengths trade accuracy against training time & held-out pair, 4 trajectories & centre not shown best at our budget; the ranking depends on training length; the time trade-off is untested & \S{}4.2 \\
RWM: it beats MLP, RSSM and transformer baselines & held-out pair, 4 trajectories & consistent with the claim, a weak test & \S{}4.3 \\
RWM-U: the learned variance captures aleatoric uncertainty & our own runs, a synthetic test with known noise, and the released checkpoint's $\sigma$ (20 trajectories) & not supported: the objective's optimum is zero & \S{}5.5 \\
RWM-U: disagreement is a trust metric & released checkpoint, 20 trajectories; our arms, 4 and 20 & supported as a ranking; wrong as a scale at every horizon, including one step & \S{}5.2, \S{}5.6 \\
RWM-U: uncertainty is effectively propagated over long horizons & released checkpoint and our arms, 20 and 20 trajectories & fails under the release's rollout; per-member rollout partly repairs it & \S{}5.4 \\
\end{longtable}\endgroup

\emph{Table 4: where each tested claim stands. Our arms are at 2,500 iterations; the released checkpoint is as released.}

\section*{Reproducibility and data}

A clean clone reproduces every scientific value: of 18,327 regenerated values, 18,324 are bitwise identical, and the 3 that differ are bookkeeping about the submission's own files (the archive's size and commit count), while 0 are a measurement, a statistic or a verdict (supplement S11). Code, every artifact and the claim ledger accompany this submission as anonymised supplementary material, and a permanent archival identifier follows acceptance. Neither upstream repository is redistributed: \texttt{setup.sh} fetches both at pinned commits and verifies their hashes.

\section*{References}

\begin{enumerate}
\item C. Li, A. Krause, M. Hutter. \emph{Robotic World Model: A Neural Network Simulator for Robust Policy Optimization in Robotics.} arXiv:2501.10100\textbf{v1}, 17 January 2025; current version v5, 14 December 2025, under the same title. Presented as a workshop poster at NeurIPS 2025 (\emph{Embodied World Models for Decision Making}); the venue's page does not say whether the workshop is archival. \emph{Read at v1, whose Roman-numeral sectioning our references follow; v2 (23 April 2025) renumbered to Arabic and moved IV-C into Appendix A.4.1.}
\item C. Li, A. Krause, M. Hutter. \emph{Offline Robotic World Model: Learning Robotic Policies without a Physics Simulator.} arXiv:2504.16680\textbf{v1}, 23 April 2025; current version v3, 8 January 2026, retitled \emph{Uncertainty-Aware Robotic World Model Makes Offline Model-Based Reinforcement Learning Work on Real Robots.} Presented as a workshop poster at ICLR 2026 (\emph{The 2nd Workshop on World Models: Understanding, Modelling and Scaling}); the venue's page does not say whether the workshop is archival. \emph{Read at v1. \S{}3.2 and Eq. 4--5 keep their numbers there; every figure and appendix table has moved, and the model is renamed RWM-O to RWM-U --- the same model, with the letter re-expanded from "Offline Robotic World Model" to "Uncertainty-Aware Robotic World Model". The two names never co-occur: v1 uses RWM-O 39 times and no RWM-U, v3 the reverse. The crosswalk is in \texttt{results/\allowbreak{}original\_\allowbreak{}paper\_\allowbreak{}figures\allowbreak{}.json}.}
\item Z. Abbas, S. Sokota, E. J. Talvitie, M. White. \emph{Selective Dyna-style Planning Under Limited Model Capacity.} ICML 2020. arXiv:2007.02418, 2020.
\item R. Agarwal, M. Schwarzer, P. S. Castro, A. Courville, M. G. Bellemare. \emph{Deep Reinforcement Learning at the Edge of the Statistical Precipice.} NeurIPS 2021. arXiv:2108.13264, 2021.
\item S. Albanie, J. F. Henriques, L. Bertinetto, A. Hernández-García, H. Doughty, G. Varol (eds.). \emph{NeurIPS 2021 Workshop on Pre-registration in Machine Learning.} PMLR 181, 2022. https://proceedings.mlr.press/v181/
\item A. N. Angelopoulos, S. Bates. \emph{A Gentle Introduction to Conformal Prediction and Distribution-Free Uncertainty Quantification.} arXiv:2107.07511, 2021.
\item S. Bengio, O. Vinyals, N. Jaitly, N. Shazeer. \emph{Scheduled Sampling for Sequence Prediction with Recurrent Neural Networks.} NeurIPS 2015. arXiv:1506.03099, 2015.
\item L. Bertinetto, J. F. Henriques, S. Albanie, M. Paganini, G. Varol (eds.). \emph{NeurIPS 2020 Workshop on Pre-registration in Machine Learning.} PMLR 148, 2021. https://proceedings.mlr.press/v148/
\item A. C. Cameron, J. B. Gelbach, D. L. Miller. \emph{Bootstrap-Based Improvements for Inference with Clustered Errors.} Review of Economics and Statistics 90(3), 2008. doi:10.1162/rest.90.3.414
\item K. Chua, R. Calandra, R. McAllister, S. Levine. \emph{Deep Reinforcement Learning in a Handful of Trials using Probabilistic Dynamics Models.} NeurIPS 2018. arXiv:1805.12114, 2018.
\item S. Fort, H. Hu, B. Lakshminarayanan. \emph{Deep Ensembles: A Loss Landscape Perspective.} arXiv:1912.02757, 2019.
\item C. Guo, G. Pleiss, Y. Sun, K. Q. Weinberger. \emph{On Calibration of Modern Neural Networks.} ICML 2017. arXiv:1706.04599, 2017.
\item D. Hafner, T. Lillicrap, I. Fischer, R. Villegas, D. Ha, H. Lee, J. Davidson. \emph{Learning Latent Dynamics for Planning from Pixels.} ICML 2019 (PMLR 97). arXiv:1811.04551, 2019.
\item D. Hafner, T. Lillicrap, M. Norouzi, J. Ba. \emph{Mastering Atari with Discrete World Models.} ICLR 2021. arXiv:2010.02193, 2021.
\item M. Havasi, R. Jenatton, S. Fort, J. Z. Liu, J. Snoek, B. Lakshminarayanan, A. M. Dai, D. Tran. \emph{Training independent subnetworks for robust prediction.} ICLR 2021. arXiv:2010.06610, 2021.
\item P. Henderson, R. Islam, P. Bachman, J. Pineau, D. Precup, D. Meger. \emph{Deep Reinforcement Learning that Matters.} AAAI 2018. arXiv:1709.06560; doi:10.1609/aaai.v32i1.11694, 2018.
\item M. Janner, J. Fu, M. Zhang, S. Levine. \emph{When to Trust Your Model: Model-Based Policy Optimization.} NeurIPS 2019. arXiv:1906.08253, 2019.
\item R. Kidambi, A. Rajeswaran, P. Netrapalli, T. Joachims. \emph{MOReL : Model-Based Offline Reinforcement Learning.} NeurIPS 2020. arXiv:2005.05951, 2020.
\item D. P. Kingma, J. Ba. \emph{Adam: A Method for Stochastic Optimization.} ICLR 2015. arXiv:1412.6980, 2015.
\item V. Kuleshov, N. Fenner, S. Ermon. \emph{Accurate Uncertainties for Deep Learning Using Calibrated Regression.} ICML 2018. arXiv:1807.00263, 2018.
\item B. Lakshminarayanan, A. Pritzel, C. Blundell. \emph{Simple and Scalable Predictive Uncertainty Estimation using Deep Ensembles.} NeurIPS 2017. arXiv:1612.01474, 2017.
\item N. O. Lambert, A. Wilcox, H. Zhang, K. S. J. Pister, R. Calandra. \emph{Learning Accurate Long-term Dynamics for Model-based Reinforcement Learning.} IEEE CDC 2021. arXiv:2012.09156; doi:10.1109/cdc45484.2021.9683134, 2021.
\item S. Lee, S. Purushwalkam, M. Cogswell, D. Crandall, D. Batra. \emph{Why M Heads are Better than One: Training a Diverse Ensemble of Deep Networks.} arXiv:1511.06314, 2015.
\item L. Lindemann, M. Cleaveland, G. Shim, G. J. Pappas. \emph{Safe Planning in Dynamic Environments using Conformal Prediction.} IEEE RA-L 2023. arXiv:2210.10254; doi:10.1109/lra.2023.3292071, 2023.
\item C. Lu, P. J. Ball, J. Parker-Holder, M. A. Osborne, S. J. Roberts. \emph{Revisiting Design Choices in Offline Model-Based Reinforcement Learning.} ICLR 2022 (Spotlight). arXiv:2110.04135, 2022.
\item A. Malik, V. Kuleshov, J. Song, D. Nemer, H. Seymour, S. Ermon. \emph{Calibrated Model-Based Deep Reinforcement Learning.} ICML 2019 (PMLR 97:4314-4323). arXiv:1906.08312, 2019.
\item I. Osband, C. Blundell, A. Pritzel, B. Van Roy. \emph{Deep Exploration via Bootstrapped DQN.} NeurIPS 2016. arXiv:1602.04621, 2016.
\item I. Osband, J. Aslanides, A. Cassirer. \emph{Randomized Prior Functions for Deep Reinforcement Learning.} NeurIPS 2018. arXiv:1806.03335, 2018.
\item Y. Ovadia, E. Fertig, J. Ren, Z. Nado, D. Sculley, S. Nowozin, J. V. Dillon, B. Lakshminarayanan, J. Snoek. \emph{Can You Trust Your Model's Uncertainty? Evaluating Predictive Uncertainty Under Dataset Shift.} NeurIPS 2019. arXiv:1906.02530, 2019.
\item M. Seitzer, A. Tavakoli, D. Antic, G. Martius. \emph{On the Pitfalls of Heteroscedastic Uncertainty Estimation with Probabilistic Neural Networks.} ICLR 2022. arXiv:2203.09168, 2022.
\item K. Stankeviciute, A. M. Alaa, M. van der Schaar. \emph{Conformal Time-series Forecasting.} NeurIPS 2021. https://proceedings.neurips.cc/paper\_files/paper/2021/hash/312f1ba2a72318edaaa995a67835fad5-Abstract.html
\item E. Talvitie. \emph{Self-Correcting Models for Model-Based Reinforcement Learning.} AAAI 2017. arXiv:1612.06018; doi:10.1609/aaai.v31i1.10850, 2017.
\item A. Venkatraman, M. Hebert, J. A. Bagnell. \emph{Improving Multi-Step Prediction of Learned Time Series Models.} AAAI 2015. doi:10.1609/aaai.v29i1.9590
\item Y. Wen, D. Tran, J. Ba. \emph{BatchEnsemble: An Alternative Approach to Efficient Ensemble and Lifelong Learning.} ICLR 2020. arXiv:2002.06715, 2020.
\item T. Yu, G. Thomas, L. Yu, S. Ermon, J. Zou, S. Levine, C. Finn, T. Ma. \emph{MOPO: Model-based Offline Policy Optimization.} NeurIPS 2020. arXiv:2005.13239, 2020.
\end{enumerate}

\emph{Entries 3--35 are generated from \texttt{results/\allowbreak{}t1\_\allowbreak{}bibliography\_\allowbreak{}verified\allowbreak{}.json}. Each was checked against the paper itself: title and full author list from the arXiv record or, for an entry with no arXiv id, from its DOI's Crossref record or its proceedings page; venue from the record or the paper's first page; and every sentence this paper attributes matched verbatim against that paper's own text: 33 of 33 entries and 19 of 19 attributed fragments, 3 of them single common words whose match verifies nothing about the attribution (ledger \texttt{D-35}). No entry was added that was not verified.}

\appendix
\section{Reimplementation and equivalence}

The reimplementation is checked against the released module level by level before any training, which is why a later difference belongs to the method and not to our wiring (\S{}3.1).

What every downstream number rests on. Each level was passed before the next was attempted.

\begin{center}\small\resizebox{\ifdim\width>\linewidth\linewidth\else\width\fi}{!}{%
\begin{tabular}{lll}
\toprule
level & claim & result \\
\midrule
shapes & parameter counts match the reference & exact \\
wiring & inference outputs match the reference module & \textbf{0.000e+00}, bitwise \\
indexing & the harness feeds the actions it claims & bitwise against the raw CSV \\
residual & the zero-delta model is the hold-last floor & 1.192e-07 \\
\textbf{objective} & \textbf{losses and gradients match} & \textbf{0.000e+00 across 7 terms, 106 tensors} \\
trainer & can memorise a single batch & 1,506$\times$ loss reduction \\
\bottomrule\end{tabular}}\end{center}

\section{Data, arenas and compute}

The data file, its windows and the arenas of Table 1 (\S{}3.2), and the commands that rerun the pipeline.

\textbf{Windows and boundaries.} The file has 10,000 rows. The reference window builder marks all 9,961 windows of 40 rows valid, 352 of which splice one episode's end onto the next one's start; the usable, episode-respecting count is 9,609, and the contamination rate of the unfiltered set is 3.53\%. Everything here ran on two CPU cores; per-run times, the training queues and their overlaps are in supplement S7.

\begin{verbatim}
./setup.sh                     # clone upstreams at pinned commits
python3.11 -m venv .venv && . .venv/bin/activate
pip install -r requirements.txt
./reproduce.sh --quick --force # everything except training
\end{verbatim}
\texttt{-{}-force} matters: a clean clone already contains each stage's declared output, so without it every stage skips.

Runtimes, the queues that produced them and the CPU work they overlapped are in supplementary S7.

\section{Statistical methods}

The procedures behind every interval and test in the paper (\S{}3.4), stated once.

\textbf{Cluster bootstrap over trajectories.} Each 95\% interval resamples whole 400-step trajectories, the unit of dependence, with replacement and recomputes the statistic inside every draw; seeds, where pooled, are pooled inside each draw. With 4 trajectories there are 256 distinct resamples, so such intervals are coarse and the percentile method is unreliable; the literature on inference with few clusters is the reason (Cameron et al. (2008)), and we pair each interval at that size with an exact test. Resampling pooled seed $\times$ trajectory values instead of trajectories narrows intervals by a mean 1.42$\times$ and changes 1 of 16 verdicts, in the out-of-sample h = 8 cell at 200-step trajectories and 2,500 iterations; wherever the two units differ we report both.

\textbf{Exact sign-flip tests.} For a paired difference over n trajectories the statistic is the mean difference and the null makes each trajectory's sign equally likely; every sign pattern is enumerated and the p-value is two-sided. At four trajectories the smallest attainable p is 0.125, above the first Holm level of the configuration sweep (0.00625) and of the baseline and size-matched families, so none of those families can reject anything; 154 of the 228 comparisons at four trajectories reach that minimum. At the episode level, at h = 368, the ten episodes give a smallest p of 0.0020.

\textbf{Multiplicity.} Holm's procedure holds the chance of any false rejection at the five per cent level within each pre-registered family. No correction is applied across rules.

\textbf{Permutation nulls that keep dimensions together.} The 45 state dimensions of a robot are coupled and share a forecast-depth trend, so a null that treats them as independent trials is wrong. Where we test a per-dimension correlation, the null permutes whole trajectories, which preserves the coupling between dimensions; Appendix I gives the tests and the binomial version they replace.

\section{Claims of the originals and the version crosswalk}

Each claim of the two original papers, what we did with it and where it stands (\S{}4, \S{}5, Table 4), and how the versions we read differ from the current ones.

Table 4 summarises the first table. It is here in full because the third column --- what the original actually reports --- is the answer to a question a reader of any reproduction should ask, and because "no quantitative figure" is itself a finding that deserves to be checkable row by row. Its verdict cells quote figures from the sections they cite, each on that section's arena, n\_independent and checkpoint; the second table gives them claim by claim.

\emph{Section references follow arXiv:2501.10100\textbf{v1}, which uses Roman-numeral sectioning. v2 renumbered to Arabic and moved IV-C's material into Appendix A.4.1. References to arXiv:2504.16680 follow \textbf{v1}, which is the version we read; it is now at v3 (8 Jan 2026), where \S{}3.2 and Eq. 4--5 keep their numbers but every figure and appendix table has moved --- Figure 2 (right) became Figure 3 (right), and the model was renamed RWM-O to RWM-U, which is a re-expansion of the letter ("Offline Robotic World Model" to "Uncertainty-Aware Robotic World Model") rather than a second variant: no version of the paper contains both names. All locations, and the occurrence counts that establish that, are recorded in \texttt{results/\allowbreak{}original\_\allowbreak{}paper\_\allowbreak{}figures\allowbreak{}.json}.}

\begingroup\small
\begin{longtable}{>{\raggedright\arraybackslash}p{\dimexpr 0.2065\linewidth-2\tabcolsep\relax}>{\raggedright\arraybackslash}p{\dimexpr 0.0894\linewidth-2\tabcolsep\relax}>{\raggedright\arraybackslash}p{\dimexpr 0.1746\linewidth-2\tabcolsep\relax}>{\raggedright\arraybackslash}p{\dimexpr 0.5295\linewidth-2\tabcolsep\relax}}
\toprule claim, and where & tested & what the original reports & verdict \\ \midrule\endfirsthead
\toprule claim, and where & tested & what the original reports & verdict \\ \midrule\endhead
\bottomrule\endlastfoot
RWM-AR consistently outperforms RWM-TF (2501.10100 \S{}IV-D) & \textbf{yes} & \textbf{no quantitative figure.} "significantly outperforms"; the gap is plotted in Fig. 7 and stated nowhere in text, caption or table & \textbf{reproduces} at long horizon (\S{}4.1); retrained on fixed folds with each episode held out once it replicates, +2.2275 [+1.3683, +3.5053] at h = 368 over all ten episodes' 20 trajectories at 2,500 iterations, not 10,000 (rule X3: \textbf{REPLICATES ACROSS FOLDS}, \S{}4.1, Appendix H), and scored with the released evaluation's timing rule M-23's verdict stands (rule X8: \textbf{SAME VERDICT}, Appendix E) \\
Teacher forcing gives "poor autoregressive performance" (\S{}IV-C) & \textbf{yes} & \textbf{quantitative}: Fig. 6 prints e for the teacher-forced N=1 row, 3.99 at (32, 1) against 0.47 at the centre, on unstated data and horizon; the passage itself gives no number & reproduces, and more strongly: on the 400-step unit Arm B is worse than the hold-last floor at every horizon (on M-64's shorter units only at the longest they reach), and the configuration sweep's (32, 1) is 6.35$\times$ the centre's error at h = 368 (Appendix H) \\
M=32, N=8 gives the optimal trade-off between accuracy and training time (\S{}IV-C) & \textbf{yes} & \textbf{quantitative}: Fig. 6 prints the error and the training hours of every cell of its grid, and the centre's error is the tied-lowest (\texttt{docs/\allowbreak{}presubmission/\allowbreak{}ORIGINAL\_\allowbreak{}SPECS\allowbreak{}.md} a.5) & \textbf{NOT OPTIMAL AT OUR BUDGET} on accuracy alone (\S{}4.2): (32, 32), (32, 16), (8, 8) and (2, 8) beat it at our budget (no reading its rule reports puts the centre ahead of any of them), but the ranking depends on training length: trained longer, the centre passes each of them on at least one reading (post hoc, \S{}4.2, Appendix H). The trade-off with training time is not tested; the original's printed hours are consistent with it. Under the released evaluation's timing the rule keeps its verdict (rule X8: \textbf{SAME VERDICT}, Appendix E) \\
Beats MLP, RSSM and transformer baselines (\S{}IV-D) & \textbf{yes} & \textbf{no quantitative figure.} "consistently achieves the lowest prediction errors across all environments"; plotted in Fig. 7, with no number in text, caption or table & \textbf{REPRODUCES} with the baselines teacher-forced, as the original trains them, and \textbf{RWM AHEAD OF ALL THREE} with them trained autoregressively, which compares architectures at one training regime rather than testing this claim (\S{}4.3). Both are the rules' verdicts at h = 368; from h = 32 and h = 128 respectively, every reading the rules report gives them, and at shorter horizons they split. The baselines are built to our reading of Table S7 and trained with RWM's settings; at h = 368 every one is worse than predicting no change, and rule X1's other read-outs and retraining variants did not rescue our RSSM (\textbf{NOT RESCUED BY THE SETTINGS TRIED}), so the RSSM comparison stays uninformative and the claim rests on the MLP and the transformer. Matched to RWM's parameter count, they return \textbf{PARTIAL} (rule X4): the MLP is behind (RWM BETTER) and the transformer cannot be settled (CANNOT BE SETTLED), matching size being one axis only (\S{}4.3). Under the released evaluation's timing both rules keep their verdicts (rule X8: \textbf{SAME VERDICT}, Appendix E). One robot where the original has several \\
Zero-shot hardware transfer (\S{}IV-E) & no & --- & \texttt{[hardware: zero-shot transfer]} no hardware; this is a dynamics-model reproduction \\
Policies transfer to hardware from \textasciitilde{}6M state transitions against \textasciitilde{}250M for the model-free baseline (its Limitations section and table, from v2 on; not in v1) --- the paper's headline sample-efficiency result & no & \textbf{6M against 250M state transitions} at equal real tracking reward (0.90 +- 0.04 against 0.90 +- 0.03), Table 1 from v2 on --- the only table of numbers in either paper & \texttt{[policy, hardware: the sample-efficiency result]} \textbf{not tested.} It is a claim about policy learning and hardware deployment, and requires the RL loop, a simulator and an ANYmal. We reproduce the dynamics model only; no policy is trained anywhere in this work, so no transition count of ours is comparable \\
MBPO-PPO beats SHAC and Dreamer (\S{}IV-E) & no & --- & \texttt{[policy: the comparisons against SHAC and Dreamer]} no policy learning reproduced \\
Generality across quadruped, humanoid, manipulation (\S{}IV-D) & no & plotted in Fig. 7; no numbers in text & \texttt{[model: generality across robot morphologies]} one released dataset, ANYmal D flat \\
Epistemic "closely follows the trend of the prediction error", justifying "its role as a trust metric" (2504.16680v1 \S{}3.2) & \textbf{yes} & \textbf{no quantitative figure.} A "strong correlation" is asserted with no coefficient, interval or sample size; plotted in Fig. 2 (right) & \textbf{supported as a scalar ranking, against a real baseline} --- the applied scalar correlates +0.605 [+0.545, +0.694] with realised error at n\_\allowbreak{}independent = 20, beats the forecast-index counter at every horizon from h = 8 (the index is constant at h = 1) --- though \textbf{not} the free \texttt{step-size} counter over whole trajectories at h = 368, +0.4697 against +0.6053, a margin below the 0.2891 minimum detectable effect (SURVIVES entry-res ONLY: disagreement beats the one-step error before the window, but not the model's predicted step size), which on 70 shorter units at h = 100 it does beat, +0.2816 [+0.1978, +0.3896] (rule X6: \textbf{DISAGREEMENT BEATS STEP SIZE}, Appendix L) --- and survives 5 controls on forecast depth --- the linear partial that keeps +0.596, and four harder ones --- plus a sixth on trajectory difficulty --- the last giving +0.419 [+0.318, +0.576] with both the rollout and the depth held constant (Appendix L, M-45). \textbf{Weaker per-dimension}: at h = 368 the 45-of-45 sign count gives a permutation P of 0.0435 (out-of-sample) and 0.0775 (in-sample), and no cell survives multiplicity correction (Appendix I). \textbf{Not supported as a scale}: 33.4$\times$ overconfident at h = 100, the method's own rollout length, and 34.4$\times$ at the h = 368 diagnostic horizon; a per-horizon rescaling brings the released checkpoint near nominal, but on a model that has not seen the test episodes it does so in only 17 of 36 cells (\S{}5.7) \\
Aleatoric "remains low, reflecting small stochasticity" (2504.16680v1 \S{}3.2) & \textbf{yes} & \textbf{no quantitative figure.} "Low" is relative to the epistemic curve on the same axes of Fig. 2 (right); no absolute value, and no comparison against realised error & the observation holds; the explanation does not (\S{}5.5) \\
Offline MBRL on real robots (2504.16680v1) & no & --- & \texttt{[policy, hardware: offline MBRL on real robots]} not tested \\
Penalising rewards by ensemble disagreement improves the learned policy (2504.16680v1 Eq. 4--5, \S{}4.1) --- the follow-up's core method claim & no & Fig. 3 (right) plots epistemic uncertainty under three penalty weights during training; no numbers & \texttt{[policy: the core claim that penalising rewards by disagreement improves the learned policy]} \textbf{not tested.} We measure the penalty quantity itself --- what it is (\S{}5.1), how well it ranks error (\S{}5.6), whether it is calibrated (\S{}5.2) --- but never train a policy with or without it. Our findings bound what the quantity \emph{reports}, not what it \emph{costs} (\S{}7); a test of the cost needing no policy, rule X5, could not be run (below) \\
\end{longtable}\endgroup

\textbf{What each tested claim rests on.} The second table has one row per headline claim of this paper, with the section that owns it. Each row gives its own arena and number of independent units, its checkpoint, whether that arena is in-sample for the model measured, the verdict as its rule or analysis returned it, and whether it survives multiplicity correction. It is generated from the artifacts each claim is computed from (\texttt{results/\allowbreak{}evidence\_\allowbreak{}summary\allowbreak{}.json}).

\begingroup\footnotesize\setlength{\tabcolsep}{3pt}
\begin{longtable}{>{\raggedright\arraybackslash}p{\dimexpr 0.2335\linewidth-2\tabcolsep\relax}>{\raggedright\arraybackslash}p{\dimexpr 0.1401\linewidth-2\tabcolsep\relax}>{\raggedright\arraybackslash}p{\dimexpr 0.1107\linewidth-2\tabcolsep\relax}>{\raggedright\arraybackslash}p{\dimexpr 0.1009\linewidth-2\tabcolsep\relax}>{\raggedright\arraybackslash}p{\dimexpr 0.2552\linewidth-2\tabcolsep\relax}>{\raggedright\arraybackslash}p{\dimexpr 0.1597\linewidth-2\tabcolsep\relax}}
\toprule claim (\S{}) & arena (n\_\allowbreak{}independent) & checkpoint & in-sample for the model measured? & verdict & survives multiplicity correction? \\ \midrule\endfirsthead
\toprule claim (\S{}) & arena (n\_\allowbreak{}independent) & checkpoint & in-sample for the model measured? & verdict & survives multiplicity correction? \\ \midrule\endhead
\bottomrule\endlastfoot
Autoregressive training beats teacher forcing at h = 368 (\S{}4.1) & out-of-sample (4) & 10,000 iterations & no & gap excludes zero, favouring autoregressive training (under the released evaluation's timing, rule X8: SAME VERDICT) & not applicable at 10,000; the 500- and 2,500-iteration cells survive Holm (Appendix H) \\
The same comparison reverses at h = 1, at the short unit M-64 built (\S{}4.1) & out-of-sample (60) & 10,000 iterations & no & gap excludes zero, favouring teacher forcing & not applicable \\
The same comparison at h = 368, retrained on fixed folds with each episode held out once, at 2,500 iterations (rule X3) (\S{}4.1) & all ten episodes (20) & 2,500 iterations & no & REPLICATES ACROSS FOLDS: gap +2.2275 [+1.3683, +3.5053]; 10 of 10 episodes positive & not applicable \\
(M, N) = (32, 8) is the optimal configuration on accuracy alone (the accuracy half of the trade-off) (\S{}4.2) & out-of-sample (4) & 2,500 iterations & no & NOT OPTIMAL AT OUR BUDGET; 4 of 8 neighbours beat the centre, and no reading the rule reports puts the centre ahead of any of them; the ranking depends on training length (Appendix H) (under the released evaluation's timing, rule X8: SAME VERDICT) & yes \\
RWM beats MLP, RSSM and transformer baselines trained with RWM's settings, at h = 368 (teacher-forced, as the original trains them; the RSSM comparison uninformative, rule X1) (\S{}4.3) & out-of-sample (4) & 2,500 iterations & no & REPRODUCES; every reading the rule reports returns it from h = 32, and below that the readings split; at h = 368 every baseline is worse than predicting no change (under the released evaluation's timing, rule X8: SAME VERDICT) & yes \\
RWM beats MLP, RSSM and transformer baselines trained with RWM's settings, at h = 368 (trained autoregressively: architecture at one training regime, not the original's claim; the RSSM comparison uninformative, rule X1) (\S{}4.3) & out-of-sample (4) & 2,500 iterations & no & RWM AHEAD OF ALL THREE; every reading the rule reports returns it from h = 128, and below that the readings split; at h = 368 every baseline is worse than predicting no change (under the released evaluation's timing, rule X8: SAME VERDICT) & yes \\
RWM beats MLP and transformer baselines matched to its parameter count, at h = 368 (rule X4; size is one axis) (\S{}4.3) & out-of-sample (4) & 2,500 iterations & no & PARTIAL: MLP RWM BETTER, transformer CANNOT BE SETTLED & yes \\
Ensemble disagreement is smaller than realised error, at h = 1 (\S{}5.2) & all ten episodes (20) & released & yes & overconfident; the ratio interval excludes 1 & not applicable \\
Ensemble disagreement is smaller than realised error, at h = 100 (\S{}5.2) & all ten episodes (20) & released & yes & overconfident; the ratio interval excludes 1 & not applicable \\
The aleatoric $\sigma$ head has collapsed and is orders of magnitude smaller than realised error, at h = 1 (\S{}5.2) & all ten episodes (20) & released & yes & overconfident; the ratio interval excludes 1 & not applicable \\
Disagreement ranks realised error better than the forecast step index, at h = 100 (\S{}5.6) & all ten episodes (20) & released & yes & paired difference excludes zero & not applicable \\
Disagreement ranks realised error better than the model's own predicted step size, over whole 400-step trajectories at h = 368 (\S{}5.6) & all ten episodes (20) & released & yes & SURVIVES entry-res ONLY (disagreement beats the one-step error before the window, but not the model's predicted step size): step size not beaten, the margin below the minimum detectable effect; the partial survives & could not at this n \\
The same comparison on shorter units, at h = 100 (rule X6) (\S{}5.6) & all ten episodes (70) & released & yes & DISAGREEMENT BEATS STEP SIZE: margin +0.2816 [+0.1978, +0.3896], against a minimum detectable effect of 0.1713 & not applicable \\
With both the rollout and the depth held constant, disagreement still tracks error (\S{}5.6, Appendix L) & all ten episodes (20) & released & yes & interval excludes zero and clears the minimum detectable effect & not applicable \\
A per-horizon multiplier brings coverage near nominal where a constant one does not (\S{}5.7) & held-out pair (4) & released & yes & every released-checkpoint point estimate unseen by the multiplier within tolerance; on Arm A at 2,500 iterations, unseen by its model too, 17 of 36 epistemic cells; tolerance not resolvable at this arena & could not at this n \\
A per-horizon multiplier fitted on nine episodes and scored on the tenth (rule X7) (\S{}5.7) & all ten episodes (20) & released & yes & WITHIN BAND AT 6 OF 6 HORIZONS for the disagreement; the aleatoric term, alongside, WITHIN BAND AT 3 OF 6 HORIZONS & not applicable \\
An ensemble that shares no trunk is better calibrated than the released topology (\S{}5.3) & out-of-sample (4) & 2,500 iterations & no & MECHANISM SUPPORTED & not applicable \\
The same contrast at matched capacity (\S{}5.3) & out-of-sample (4) & 2,500 iterations & no & UNDER-POWERED --- favours the matched ensemble by less than the MDE & could not at this n \\
Independence and the corrected objective together improve on the released topology (\S{}5.3) & out-of-sample (4) & 2,500 iterations & no & THE COMBINATION IMPROVES CALIBRATION & not applicable \\
The released evaluation pairs states and actions one step stale (\S{}6) & all ten episodes (20) & released & yes & defect confirmed in the code; raises error up to h = 32 on relative-L1; not resolved from h = 100 & not applicable \\
\end{longtable}\endgroup

\textbf{Versions.} We read v1 of the base paper and v1 of the follow-up; their current versions are v5 and v3. The ensemble's rollout, the use of the ensemble mean, the shared trunk and the sentence on the learned variance are word for word the same in every version of the follow-up. What its second version added is the claim that uncertainty is effectively propagated over long horizons, which is therefore tested against accumulated error (\S{}5.4). The base paper's sample-efficiency table is in its second version onward and not in the first. Both training tables moved between versions without changing their values; the follow-up was renamed (RWM-O to RWM-U) with the model unchanged. The workshop versions were not read.

\textbf{Rule X5.} The test of the penalty against the model's error in predicted return could not be run, because the released data do not record the velocity command its reward terms score against; supplement S12 gives the reason in full. It was never pre-registered as a rule, and nothing in the paper infers a command from achieved velocity.

\section{The pre-registration record}

Every pre-registered rule with its plain name, identifier, committed hash, lead time and verbatim verdict, which supports \S{}3.4's account of what was decided in advance; and the re-scoring of four rules under the released evaluation's timing.

\S{}3.4's argument rests on decision rules committed to git before the data that tested them, and the body names those rules by identifier. An identifier with no table behind it is either decoration or an instruction to open a 738 KB ledger, so here is the table. It is generated from \texttt{FINDINGS\_LEDGER.md} and \texttt{results/\allowbreak{}appendix\_\allowbreak{}g\_\allowbreak{}rules\allowbreak{}.json}; nothing in it is typed.

\textbf{Lead time} is the rule's commit timestamp subtracted from the commit that first held the data it tested, never by a stored hash: the first rules' commits are found by \emph{subject}, because the history was rewritten once and two of the commits Figure 3 cites changed hash (supplement S8), and the later rules' by the commit that introduced their ledger heading. Positive means the rule was in git before the data existed. This is the same computation Figure 3 plots. A rule whose data do not exist yet has neither a lead time nor a verdict, and Figure 3 leaves it out until they do.

\begin{figure}[!ht]\centering\includegraphics[width=\linewidth]{\detokenize{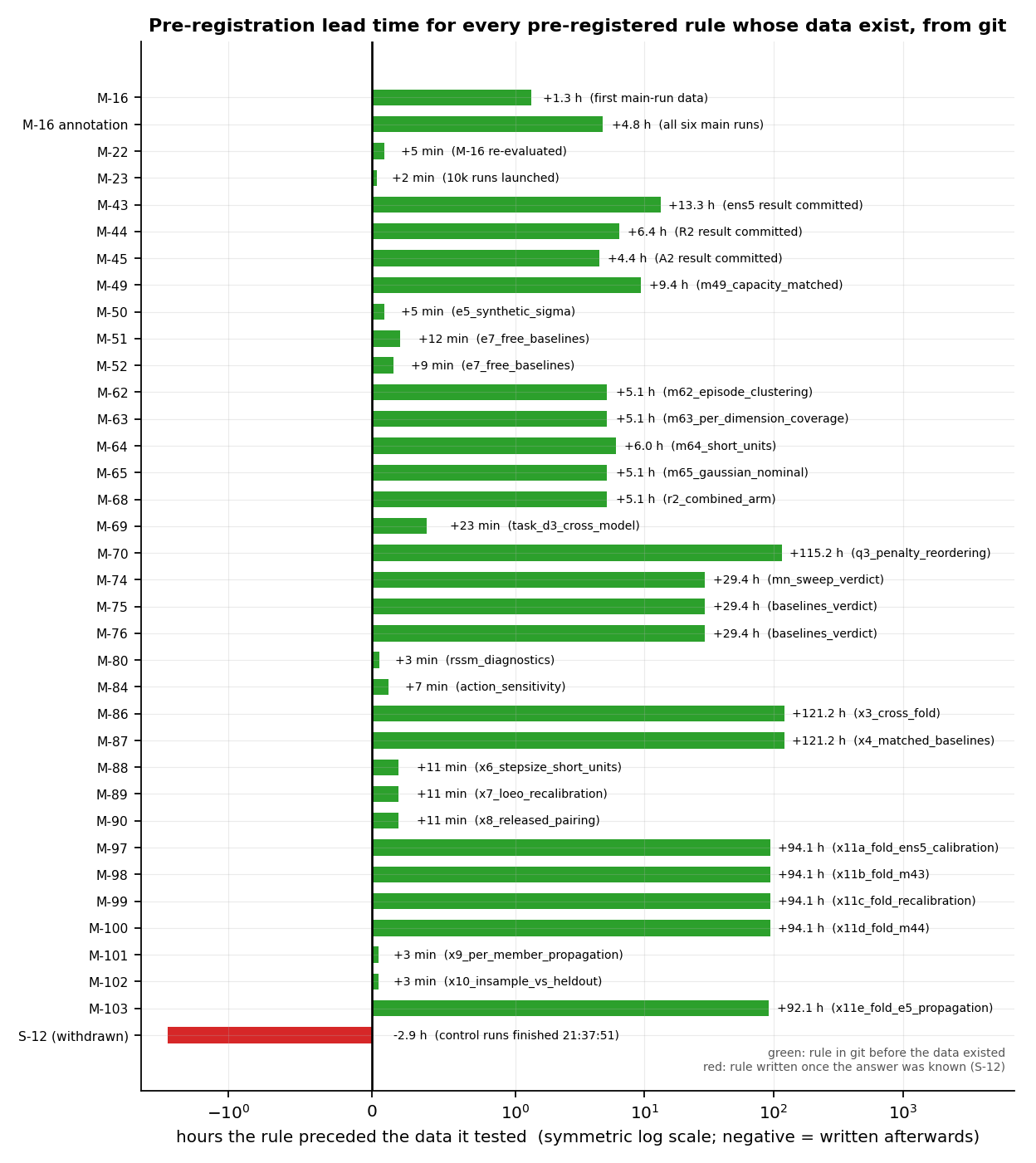}}\caption{Pre-registration lead time for every pre-registered rule whose data exist (Appendix E), with M-16's annotation beside its rule, from git commit timestamps on a symmetric-logarithmic axis. Positive is a rule committed before the data that tested it existed; negative is a rule written afterwards. A lead of a few minutes means the rule was committed just before its runs were launched or its readings computed. The one negative bar, labelled the Task 3 duplication rule, is the duplication control of Appendix N, withdrawn as a pre-registration in this paper (§3.4).}\end{figure}

\begingroup\footnotesize\setlength{\tabcolsep}{3pt}
\begin{longtable}{>{\raggedright\arraybackslash}p{\dimexpr 0.0617\linewidth-2\tabcolsep\relax}>{\raggedright\arraybackslash}p{\dimexpr 0.3261\linewidth-2\tabcolsep\relax}>{\raggedright\arraybackslash}p{\dimexpr 0.2311\linewidth-2\tabcolsep\relax}>{\raggedright\arraybackslash}p{\dimexpr 0.0715\linewidth-2\tabcolsep\relax}>{\raggedright\arraybackslash}p{\dimexpr 0.1401\linewidth-2\tabcolsep\relax}>{\raggedright\arraybackslash}p{\dimexpr 0.1695\linewidth-2\tabcolsep\relax}}
\toprule rule & what it governs & commit & lead time & tested by & verdict \\ \midrule\endfirsthead
\toprule rule & what it governs & commit & lead time & tested by & verdict \\ \midrule\endhead
\bottomrule\endlastfoot
\texttt{M-16} & The Arm A /\allowbreak{} Arm B comparison & \texttt{C004} Step 5: pre-register the decision rule before launching any main run & +1.3 h & first main-run data & CANNOT BE SETTLED AT THIS BUDGET \\
\texttt{M-22} & Whether episode difficulty biases the A/\allowbreak{}B comparison & \texttt{C013} Pre-register the Task 4b difficulty-bias rule, and the two-arena convention & +5 min & M-16 re-evaluated & RESOLVED --- branch 1, 4c not run \\
\texttt{M-23} & The 10,000-iteration comparison & \texttt{C016} 5.1: pre-register M-23, the long-horizon decision rule & +2 min & 10k runs launched & RESOLVED --- reproduces at long horizon \\
\texttt{M-43} & The ensemble-5 replication & \texttt{C081} PRE-REGISTER the ensemble-5 replication rule, before the runs exist & +13.3 h & ens5 result committed & DOES NOT GENERALISE \\
\texttt{M-44} & The trunk-sharing mechanism & \texttt{C090} PRE-REGISTER M-44 and M-45, with the power check M-43 was committed without & +6.4 h & R2 result committed & MECHANISM SUPPORTED \\
\texttt{M-45} & The within-trajectory control & \texttt{C090} PRE-REGISTER M-44 and M-45, with the power check M-43 was committed without & +4.4 h & A2 result committed & DISAGREEMENT CARRIES WITHIN-ROLLOUT INFORMATION \\
\texttt{M-49} & Pre-registered: does trunk-sharing survive capacity matching? & \texttt{C123} M-49: pre-register the capacity-matched test, with the MDE that bounds it & +9.4 h & m49\_\allowbreak{}capacity\_\allowbreak{}matched\allowbreak{}.json & UNDER-POWERED --- favours the matched ensemble by less than the MDE \\
\texttt{M-50} & Pre-registered: is the sigma collapse driven by the objective, on data whose noise is known? & \texttt{C127} M-50 pre-registered, and a different-architecture run stopped contaminating section 6.3 & +5 min & e5\_\allowbreak{}synthetic\_\allowbreak{}sigma\allowbreak{}.json & OBJECTIVE-DRIVEN \\
\texttt{M-51} & Pre-registered: does the ranking claim survive more than one free adversary? & \texttt{C128} M-51 pre-registered: two more free adversaries for the ranking claim & +12 min & e7\_\allowbreak{}free\_\allowbreak{}baselines\allowbreak{}.json & SURVIVES entry-res ONLY \\
\texttt{M-52} & M-51 named a quantity that does not exist, and what replaced it & \texttt{C129} M-52: M-51 named a quantity that does not exist, corrected before it ran & +9 min & e7\_\allowbreak{}free\_\allowbreak{}baselines\allowbreak{}.json & SURVIVES entry-res ONLY \\
\texttt{M-62} & Pre-registered: does any headline verdict change when the bootstrap resamples episodes rather than 400-step trajectories? & \texttt{C154} M-62 to M-65: four rules committed before the data that will test them & +5.1 h & m62\_\allowbreak{}episode\_\allowbreak{}clustering\allowbreak{}.json & NO MOVE \\
\texttt{M-63} & Pre-registered: is the h=1 coverage failure uniform across the 45 state dimensions, or carried by a few? & \texttt{C154} M-62 to M-65: four rules committed before the data that will test them & +5.1 h & m63\_\allowbreak{}per\_\allowbreak{}dimension\_\allowbreak{}coverage\allowbreak{}.json & UNIFORM \\
\texttt{M-64} & Pre-registered: does the horizon-scoped power increase from shorter evaluation units change any verdict at h $\le$ 128? & \texttt{C154} M-62 to M-65: four rules committed before the data that will test them & +6.0 h & m64\_\allowbreak{}short\_\allowbreak{}units\allowbreak{}.json & MOVES, at one cell \\
\texttt{M-65} & Pre-registered: is the Gaussian nominal of 68.27\% defensible, or is the error distribution heavy-tailed? & \texttt{C154} M-62 to M-65: four rules committed before the data that will test them & +5.1 h & m65\_\allowbreak{}gaussian\_\allowbreak{}nominal\allowbreak{}.json & NOT HEAVY-TAILED, in every one of 48 model x horizon cells \\
\texttt{M-68} & The combined arm: independence and the corrected objective together & \texttt{C180} Pre-register M-68: the combined arm's rule, its verdict branches and its minimum detectable effect & +5.1 h & r2\_\allowbreak{}combined\_\allowbreak{}arm\allowbreak{}.json & THE COMBINATION IMPROVES CALIBRATION \\
\texttt{M-69} & The cross-model transfer of the per-horizon multiplier table & \texttt{C182} Pre-register M-69: the cross-model transfer rule, its criterion and its minimum detectable effect & +23 min & task\_\allowbreak{}d3\_\allowbreak{}cross\_\allowbreak{}model\allowbreak{}.json & DOES NOT TRANSFER --- A PROPERTY OF THE MODEL \\
\texttt{M-70} & Whether the per-horizon correction reorders cumulative penalties & \texttt{C206} Pre-register M-70: whether the per-horizon correction reorders cumulative penalties & +115.2 h & q3\_\allowbreak{}penalty\_\allowbreak{}reordering\allowbreak{}.json & DOES NOT REORDER \\
\texttt{M-74} & The configuration claim: is (M, N) = (32, 8) optimal? & \texttt{C225} PRE-REGISTER the M/\allowbreak{}N sweep and architecture-baseline rules, before the runs exist & +29.4 h & mn\_\allowbreak{}sweep\_\allowbreak{}verdict\allowbreak{}.json & NOT OPTIMAL AT OUR BUDGET \\
\texttt{M-75} & The architecture claim, baselines teacher-forced, as the original trains them & \texttt{C225} PRE-REGISTER the M/\allowbreak{}N sweep and architecture-baseline rules, before the runs exist & +29.4 h & baselines\_\allowbreak{}verdict\allowbreak{}.json & REPRODUCES \\
\texttt{M-76} & The architecture claim, baselines trained autoregressively, as RWM is & \texttt{C225} PRE-REGISTER the M/\allowbreak{}N sweep and architecture-baseline rules, before the runs exist & +29.4 h & baselines\_\allowbreak{}verdict\allowbreak{}.json & RWM AHEAD OF ALL THREE \\
\texttt{M-80} & X1: is our RSSM's long-horizon failure a matter of how its forecast is read, of its training settings, or of neither? & \texttt{C301} PRE-REGISTER RSSM diagnostic X1, before its readings exist & +3 min & rssm\_\allowbreak{}diagnostics\allowbreak{}.json & NOT RESCUED BY THE SETTINGS TRIED \\
\texttt{M-84} & X2: do our trained models condition their forecasts on the action they are given? & \texttt{C409} PRE-REGISTER action-sensitivity rule X2, before its readings exist & +7 min & action\_\allowbreak{}sensitivity\allowbreak{}.json & RESPONDS TO THE ACTION at 2,500 and at 10,000 iterations \\
\texttt{M-86} & X3: does the base claim replicate when each episode is held out in turn? & \texttt{C490} PRE-REGISTER X3 cross-fold replication and X4 size-matched baselines, before their runs exist & +121.2 h & x3\_\allowbreak{}cross\_\allowbreak{}fold\allowbreak{}.json & REPLICATES ACROSS FOLDS \\
\texttt{M-87} & X4: is RWM still ahead of baselines matched to its size? & \texttt{C490} PRE-REGISTER X3 cross-fold replication and X4 size-matched baselines, before their runs exist & +121.2 h & x4\_\allowbreak{}matched\_\allowbreak{}baselines\allowbreak{}.json & PARTIAL \\
\texttt{M-88} & X6: does ensemble disagreement beat the model's own predicted step size on more, shorter units? & \texttt{C496} PRE-REGISTER X5--X8, inference-only rules, before their readings exist & +11 min & x6\_\allowbreak{}stepsize\_\allowbreak{}short\_\allowbreak{}units\allowbreak{}.json & DISAGREEMENT BEATS STEP SIZE \\
\texttt{M-89} & X7: per-horizon recalibration, fitted on nine episodes and scored on the tenth & \texttt{C496} PRE-REGISTER X5--X8, inference-only rules, before their readings exist & +11 min & x7\_\allowbreak{}loeo\_\allowbreak{}recalibration\allowbreak{}.json & WITHIN BAND AT 6 OF 6 HORIZONS (epistemic) \\
\texttt{M-90} & X8: would any verdict change under the released evaluation's action timing? & \texttt{C496} PRE-REGISTER X5--X8, inference-only rules, before their readings exist & +11 min & x8\_\allowbreak{}released\_\allowbreak{}pairing\allowbreak{}.json & SAME VERDICT for M-74, M-75, M-76 and the A/\allowbreak{}B comparison \\
\texttt{M-97} & X11a: the ensemble-5 arms' held-out calibration at n = 20, and in-sample against held-out with every episode in both roles & \texttt{C604} [GT1] PRE-REGISTER X11a-X11d (M-97 to M-100) and the X11 rule text, before any X11 run exists & +94.1 h & x11a\_\allowbreak{}fold\_\allowbreak{}ens5\_\allowbreak{}calibration\allowbreak{}.json & MIXED \\
\texttt{M-98} & X11b: M-43's disagreement-against-index comparison on X3's five folds & \texttt{C604} [GT1] PRE-REGISTER X11a-X11d (M-97 to M-100) and the X11 rule text, before any X11 run exists & +94.1 h & x11b\_\allowbreak{}fold\_\allowbreak{}m43\allowbreak{}.json & GENERALISES \\
\texttt{M-99} & X11c: the per-horizon recalibration on episodes the ensemble-5 arms never saw & \texttt{C604} [GT1] PRE-REGISTER X11a-X11d (M-97 to M-100) and the X11 rule text, before any X11 run exists & +94.1 h & x11c\_\allowbreak{}fold\_\allowbreak{}recalibration\allowbreak{}.json & PER-HORIZON SCALAR DOES NOT RESTORE CALIBRATION \\
\texttt{M-100} & X11d: M-44's independent-against-shared-trunk comparison on X3's five folds & \texttt{C604} [GT1] PRE-REGISTER X11a-X11d (M-97 to M-100) and the X11 rule text, before any X11 run exists & +94.1 h & x11d\_\allowbreak{}fold\_\allowbreak{}m44\allowbreak{}.json & MECHANISM SUPPORTED \\
\texttt{M-101} & X9: does propagating each member on its own predictions make the spread track accumulated error? & \texttt{C608} [G3] PRE-REGISTER X9 and X10: ledger entries and scripts (M-101 X9, M-102 X10, M-103 X11e) & +3 min & x9\_\allowbreak{}per\_\allowbreak{}member\_\allowbreak{}propagation\allowbreak{}.json & PROPAGATION NARROWS THE GAP \\
\texttt{M-102} & X10: does measuring on training data flatter an ensemble's calibration? & \texttt{C608} [G3] PRE-REGISTER X9 and X10: ledger entries and scripts (M-101 X9, M-102 X10, M-103 X11e) & +3 min & x10\_\allowbreak{}insample\_\allowbreak{}vs\_\allowbreak{}heldout\allowbreak{}.json & IN-SAMPLE IS WORSE \\
\texttt{M-103} & X11e: X9's per-member propagation on X3's folds' ensemble-5 arms & \texttt{C608} [G3] PRE-REGISTER X9 and X10: ledger entries and scripts (M-101 X9, M-102 X10, M-103 X11e) & +92.1 h & x11e\_\allowbreak{}fold\_\allowbreak{}e5\_\allowbreak{}propagation\allowbreak{}.json & PROPAGATION NARROWS THE GAP \\
\texttt{S-12} & "Task 3's duplication rule was pre-registered" & \texttt{C025} Task 3: the duplication control confirms R-47's mechanism and refutes its statistic & -2.9 h & control runs finished 21:37:51 & RETRACTED \\
\end{longtable}\endgroup

\texttt{M-69}'s discharge commit was amended 2 minutes after it was created, so its lead time is +21 min by that commit's author time and +23 min by its committer time. Both are positive, so the rule reached git before its data on either reading.

35 rules, 35 with a computed lead time, of which 34 are positive and 1 negative. \textbf{The negative one is kept deliberately.} \texttt{S-12} withdraws the claim that the Task 3 duplication rule was pre-registered; the control runs had finished before any threshold reached git. A table that dropped it would be asserting exactly what the ledger retracts.

\textbf{\texttt{M-52} is in the table and that is deliberate.} It is the one mid-flight amendment to a pre-registration in this project: \texttt{M-51} named a baseline that does not exist in the artifact it named --- the residual on the last teacher-forced step of the history window, which the rollout helper never computes because it copies the history rather than predicting it --- and \texttt{M-52} names the replacement, committed before the replacement's statistic was computed. A table of pre-registrations that omitted the one amendment would be a highlights reel, so the table's rows are selected by each entry's \texttt{Status} line and their count is asserted against the set \texttt{scripts/\allowbreak{}ledger\_\allowbreak{}check\allowbreak{}.py} reports.

\textbf{What each rule says, in its own committed words} is in the supplementary material, as \texttt{docs/\allowbreak{}APPENDIX\_\allowbreak{}G\_\allowbreak{}RULES\allowbreak{}.md} --- every rule's text unabridged, generated from the ledger by the same script that generates this table. Quoting all 35 in full here would add pages to an appendix whose job is to be checkable at a glance, and quoting them in part would ship quotations ending mid-sentence. The table is the claim; the supplementary is the evidence.

\textbf{Rule X8.} Rule X8 re-scores four of our rules with the released evaluation's timing, through each rule's own evaluator, and all four keep their verdicts: for the configuration and baseline rules (M-74, M-75 and M-76, three seeds at 2,500 iterations) and the A/B comparison (M-23's statistic, three seeds at 10,000), all on the held-out pair's 4 trajectories, it returns \textbf{SAME VERDICT}, though 22 of their alongside readings change (below); this concerns our models, not the original's printed numbers.

Rule X8 was committed and pushed before its readings existed (ledger M-90; Appendix E gives its lead time) and is discharged in M-93. It is exploratory and re-opens no rule. It re-scores rules M-74, M-75 and M-76 (their models at 2,500 iterations, three seeds each) and the A/B comparison at 10,000 iterations (three seeds; M-23's statistic and that of the by-horizon comparison in Figure 4) with the released evaluation's timing, which pairs the prediction of each state with the action before the one that produced it, through each rule's own evaluator and statistic. Every evaluator first reproduces its committed artifact under the causal pairing. Our models were trained under the causal pairing, so this asks whether any of our verdicts depends on the timing a reader of the released code would score them with; it says nothing direct about the original papers' printed numbers. Every rule returns \textbf{SAME VERDICT}, and 22 of their alongside readings change: each changed cell is in bold below, and the by-horizon trend reading's change is quoted under the last table. The arena is the held-out pair, episodes 1 and 8, 4 non-overlapping 400-step trajectories, n\_independent = 4, except where a row names the in-sample arena (16).

\begin{center}\small\resizebox{\ifdim\width>\linewidth\linewidth\else\width\fi}{!}{%
\begin{tabular}{lll}
\toprule
reading & causal pairing, as committed & released pairing \\
\midrule
verdict (held-out pair, relative-L1 at h = 368) & NOT OPTIMAL AT OUR BUDGET & NOT OPTIMAL AT OUR BUDGET \\
held-out pair, relative-L1 at h = 1 & NOT OPTIMAL AT OUR BUDGET & NOT OPTIMAL AT OUR BUDGET \\
held-out pair, relative-L1 at h = 8 & NOT OPTIMAL AT OUR BUDGET & NOT OPTIMAL AT OUR BUDGET \\
held-out pair, relative-L1 at h = 32 & NOT OPTIMAL AT OUR BUDGET & \textbf{CANNOT BE DISTINGUISHED} \\
held-out pair, relative-L1 at h = 100 & NOT OPTIMAL AT OUR BUDGET & NOT OPTIMAL AT OUR BUDGET \\
held-out pair, relative-L1 at h = 128 & NOT OPTIMAL AT OUR BUDGET & NOT OPTIMAL AT OUR BUDGET \\
held-out pair, relative-L1 at h = 368 & NOT OPTIMAL AT OUR BUDGET & NOT OPTIMAL AT OUR BUDGET \\
held-out pair, nRMSE at h = 1 & CANNOT BE DISTINGUISHED & \textbf{NOT OPTIMAL AT OUR BUDGET} \\
held-out pair, nRMSE at h = 8 & NOT OPTIMAL AT OUR BUDGET & NOT OPTIMAL AT OUR BUDGET \\
held-out pair, nRMSE at h = 32 & NOT OPTIMAL AT OUR BUDGET & \textbf{CANNOT BE DISTINGUISHED} \\
held-out pair, nRMSE at h = 100 & NOT OPTIMAL AT OUR BUDGET & NOT OPTIMAL AT OUR BUDGET \\
held-out pair, nRMSE at h = 128 & NOT OPTIMAL AT OUR BUDGET & NOT OPTIMAL AT OUR BUDGET \\
held-out pair, nRMSE at h = 368 & NOT OPTIMAL AT OUR BUDGET & NOT OPTIMAL AT OUR BUDGET \\
in-sample arena, relative-L1 at h = 100 & NOT OPTIMAL AT OUR BUDGET & NOT OPTIMAL AT OUR BUDGET \\
in-sample arena, relative-L1 at h = 368 & NOT OPTIMAL AT OUR BUDGET & NOT OPTIMAL AT OUR BUDGET \\
in-sample arena, nRMSE at h = 100 & NOT OPTIMAL AT OUR BUDGET & NOT OPTIMAL AT OUR BUDGET \\
in-sample arena, nRMSE at h = 368 & NOT OPTIMAL AT OUR BUDGET & NOT OPTIMAL AT OUR BUDGET \\
(32, 32): episodes where the centre is ahead, of ten & 0 & 0 \\
(16, 8): episodes where the centre is ahead, of ten & 6 & \textbf{9} \\
(32, 16): episodes where the centre is ahead, of ten & 0 & 0 \\
(8, 8): episodes where the centre is ahead, of ten & 2 & \textbf{7} \\
(32, 2): episodes where the centre is ahead, of ten & 10 & 10 \\
(32, 1): episodes where the centre is ahead, of ten & 10 & 10 \\
(2, 8): episodes where the centre is ahead, of ten & 1 & \textbf{2} \\
(1, 8): episodes where the centre is ahead, of ten & 7 & \textbf{9} \\
\bottomrule\end{tabular}}\end{center}

\textbf{Rule M-74, the configuration sweep: nine configurations at 2,500 iterations, three seeds each; arenas as named in each row.} nRMSE as the rule's evaluator averages it, per trajectory. Bold, a reading that changes.

\begin{center}\small\resizebox{\ifdim\width>\linewidth\linewidth\else\width\fi}{!}{%
\begin{tabular}{lll}
\toprule
reading & causal pairing, as committed & released pairing \\
\midrule
verdict (held-out pair, relative-L1 at h = 368) & REPRODUCES & REPRODUCES \\
held-out pair, relative-L1 at h = 1 & CANNOT BE SETTLED & CANNOT BE SETTLED \\
held-out pair, relative-L1 at h = 8 & PARTIAL & PARTIAL \\
held-out pair, relative-L1 at h = 32 & REPRODUCES & REPRODUCES \\
held-out pair, relative-L1 at h = 100 & REPRODUCES & REPRODUCES \\
held-out pair, relative-L1 at h = 128 & REPRODUCES & REPRODUCES \\
held-out pair, relative-L1 at h = 368 & REPRODUCES & REPRODUCES \\
held-out pair, nRMSE at h = 1 & CANNOT BE SETTLED & CANNOT BE SETTLED \\
held-out pair, nRMSE at h = 8 & PARTIAL & PARTIAL \\
held-out pair, nRMSE at h = 32 & REPRODUCES & REPRODUCES \\
held-out pair, nRMSE at h = 100 & REPRODUCES & REPRODUCES \\
held-out pair, nRMSE at h = 128 & REPRODUCES & REPRODUCES \\
held-out pair, nRMSE at h = 368 & REPRODUCES & REPRODUCES \\
in-sample arena, relative-L1 at h = 1 & CANNOT BE SETTLED & \textbf{DOES NOT REPRODUCE} \\
in-sample arena, relative-L1 at h = 8 & PARTIAL & PARTIAL \\
in-sample arena, relative-L1 at h = 32 & REPRODUCES & REPRODUCES \\
in-sample arena, relative-L1 at h = 100 & REPRODUCES & REPRODUCES \\
in-sample arena, relative-L1 at h = 128 & REPRODUCES & REPRODUCES \\
in-sample arena, relative-L1 at h = 368 & REPRODUCES & REPRODUCES \\
in-sample arena, nRMSE at h = 1 & CANNOT BE SETTLED & \textbf{DOES NOT REPRODUCE} \\
in-sample arena, nRMSE at h = 8 & PARTIAL & PARTIAL \\
in-sample arena, nRMSE at h = 32 & REPRODUCES & REPRODUCES \\
in-sample arena, nRMSE at h = 100 & REPRODUCES & REPRODUCES \\
in-sample arena, nRMSE at h = 128 & REPRODUCES & REPRODUCES \\
in-sample arena, nRMSE at h = 368 & REPRODUCES & REPRODUCES \\
\bottomrule\end{tabular}}\end{center}

\textbf{Rule M-75, the baselines teacher-forced as the original trains them, at 2,500 iterations, three seeds each; arenas as named in each row.} The RSSM reads the window's first action row at its first filtering step, because the row before it lies outside the window. Bold, a reading that changes.

\begin{center}\small\resizebox{\ifdim\width>\linewidth\linewidth\else\width\fi}{!}{%
\begin{tabular}{lll}
\toprule
reading & causal pairing, as committed & released pairing \\
\midrule
verdict (held-out pair, relative-L1 at h = 368) & RWM AHEAD OF ALL THREE & RWM AHEAD OF ALL THREE \\
held-out pair, relative-L1 at h = 1 & CANNOT BE SETTLED & CANNOT BE SETTLED \\
held-out pair, relative-L1 at h = 8 & PARTIAL & PARTIAL \\
held-out pair, relative-L1 at h = 32 & PARTIAL & PARTIAL \\
held-out pair, relative-L1 at h = 100 & RWM AHEAD OF ALL THREE & RWM AHEAD OF ALL THREE \\
held-out pair, relative-L1 at h = 128 & RWM AHEAD OF ALL THREE & RWM AHEAD OF ALL THREE \\
held-out pair, relative-L1 at h = 368 & RWM AHEAD OF ALL THREE & RWM AHEAD OF ALL THREE \\
held-out pair, nRMSE at h = 1 & CANNOT BE SETTLED & CANNOT BE SETTLED \\
held-out pair, nRMSE at h = 8 & PARTIAL & PARTIAL \\
held-out pair, nRMSE at h = 32 & PARTIAL & PARTIAL \\
held-out pair, nRMSE at h = 100 & RWM AHEAD OF ALL THREE & RWM AHEAD OF ALL THREE \\
held-out pair, nRMSE at h = 128 & RWM AHEAD OF ALL THREE & RWM AHEAD OF ALL THREE \\
held-out pair, nRMSE at h = 368 & RWM AHEAD OF ALL THREE & RWM AHEAD OF ALL THREE \\
in-sample arena, relative-L1 at h = 1 & PARTIAL & PARTIAL \\
in-sample arena, relative-L1 at h = 8 & PARTIAL & \textbf{A BASELINE AHEAD} \\
in-sample arena, relative-L1 at h = 32 & PARTIAL & \textbf{A BASELINE AHEAD} \\
in-sample arena, relative-L1 at h = 100 & RWM AHEAD OF ALL THREE & RWM AHEAD OF ALL THREE \\
in-sample arena, relative-L1 at h = 128 & RWM AHEAD OF ALL THREE & RWM AHEAD OF ALL THREE \\
in-sample arena, relative-L1 at h = 368 & RWM AHEAD OF ALL THREE & RWM AHEAD OF ALL THREE \\
in-sample arena, nRMSE at h = 1 & PARTIAL & PARTIAL \\
in-sample arena, nRMSE at h = 8 & PARTIAL & \textbf{A BASELINE AHEAD} \\
in-sample arena, nRMSE at h = 32 & PARTIAL & \textbf{A BASELINE AHEAD} \\
in-sample arena, nRMSE at h = 100 & RWM AHEAD OF ALL THREE & \textbf{PARTIAL} \\
in-sample arena, nRMSE at h = 128 & RWM AHEAD OF ALL THREE & RWM AHEAD OF ALL THREE \\
in-sample arena, nRMSE at h = 368 & RWM AHEAD OF ALL THREE & RWM AHEAD OF ALL THREE \\
\bottomrule\end{tabular}}\end{center}

\textbf{Rule M-76, the baselines trained autoregressively, at 2,500 iterations, three seeds each; arenas as named in each row.} The RSSM caveat of the previous table applies. Bold, a reading that changes.

\begin{center}\small\resizebox{\ifdim\width>\linewidth\linewidth\else\width\fi}{!}{%
\begin{tabular}{lll}
\toprule
reading & causal pairing, as committed & released pairing \\
\midrule
M-23's verdict, three seeds (held-out pair, relative-L1 at h = 368) & REPRODUCES AT LONG HORIZON & REPRODUCES AT LONG HORIZON \\
M-23's verdict, seed 1 as discharged & REPRODUCES AT LONG HORIZON & REPRODUCES AT LONG HORIZON \\
M-23's conditions at h = 8 held-out & REPRODUCES AT LONG HORIZON & \textbf{FAILS TO REPRODUCE} \\
M-23's conditions at nRMSE h = 368 held-out & REPRODUCES AT LONG HORIZON & REPRODUCES AT LONG HORIZON \\
M-23's conditions at h = 368 in-sample & REPRODUCES AT LONG HORIZON & REPRODUCES AT LONG HORIZON \\
secondary at h = 8 held-out & A leads, CI excludes 0 & \textbf{spans zero} \\
secondary at h = 368 in-sample & A leads, CI excludes 0 & A leads, CI excludes 0 \\
secondary at h = 8 in-sample & A leads, CI excludes 0 & \textbf{spans zero} \\
by-horizon gap at h = 1 & spans zero & \textbf{excludes zero} \\
by-horizon gap at h = 8 & excludes zero & \textbf{spans zero} \\
by-horizon gap at h = 32 & excludes zero & excludes zero \\
by-horizon gap at h = 100 & excludes zero & excludes zero \\
by-horizon gap at h = 128 & excludes zero & excludes zero \\
by-horizon gap at h = 368 & excludes zero & excludes zero \\
episodes where autoregressive training is ahead at h = 1, of ten & 3 & \textbf{0} \\
episodes where autoregressive training is ahead at h = 8, of ten & 10 & \textbf{4} \\
episodes where autoregressive training is ahead at h = 32, of ten & 10 & 10 \\
episodes where autoregressive training is ahead at h = 100, of ten & 10 & 10 \\
episodes where autoregressive training is ahead at h = 128, of ten & 10 & 10 \\
episodes where autoregressive training is ahead at h = 368, of ten & 10 & 10 \\
\bottomrule\end{tabular}}\end{center}

\textbf{The A/B comparison, Arm A against Arm B at 10,000 iterations, three seeds each; arenas as named in each row, the held-out pair otherwise.} Bold, a reading that changes. The by-horizon trend reading changes too, from "the ratio B/A rises at every step of the horizon grid, so h=368 is the end of a trend rather than a selected point; the gap excludes zero at h=8, h=32, h=100, h=128, h=368 and spans zero at h=1; at h=8, the horizon the model is trained on, the gap excludes zero; the ratio is below one at h=1" under the causal pairing to "the ratio B/A rises at every step of the horizon grid, so h=368 is the end of a trend rather than a selected point; the gap excludes zero at h=1, h=32, h=100, h=128, h=368 and spans zero at h=8; at h=8, the horizon the model is trained on, the gap spans zero; the ratio is below one at h=1" under the released one.

\section{Metrics and floors}

Definitions of the metrics and floors (\S{}3.3), the two nRMSE aggregations and the check on the coverage nominal.

\textbf{The floors (\S{}3.3, Table 2).} Predicting the training mean forecasts every step by each dimension's mean over the arena's training-episode rows; holding the last state repeats the last history row; the seasonal floor repeats the history's last 27 rows in order, a period of 0.54 s fixed from the training episodes' joint-position autocorrelation before any held-out scoring (26 to 27 rows across the folds).

\textbf{The metrics' formulas (\S{}3.3).}

\textbf{Relative-L1} is the reference's own metric, reproduced in behaviour (\texttt{model\_training.py:203}) so that our numbers are comparable to the upstream's printed one. On config-normalised states, per forecast step,

\[r_t \;=\; \frac{\sum_{d=1}^{45}\bigl|\hat{s}_{t,d}-s_{t,d}\bigr|}{\sum_{d=1}^{45}\bigl|s_{t,d}\bigr|}\]

and the reported figure is the flat mean over trajectories and steps,

\[e \;=\; \frac{1}{B\,(T-t_0)}\sum_{b=1}^{B}\sum_{t=t_0+1}^{T} r_{t}^{(b)}\]

with $t_0$ = \texttt{history\_horizon} = 32: the first 32 steps are teacher-forced and excluded. The denominator is recomputed at every step as a 45-term sum in normalised space, so it can pass through zero, which is why this metric goes non-finite on low-dimensional state groups and why a second one exists.

\textbf{Normalised RMSE} fixes the denominator once, over the training episodes only:

\[\mathrm{nRMSE}_t \;=\; \frac{\sqrt{\dfrac{1}{45}\sum_{d=1}^{45}\mathrm{MSE}_{t,d}}}{\dfrac{1}{45}\sum_{d=1}^{45}\sigma^{\mathrm{tr}}_{d}}\quad\text{with}\quad \mathrm{MSE}_{t,d}=\frac{1}{B}\sum_{b=1}^{B}\bigl(\hat{s}^{(b)}_{t,d}-s^{(b)}_{t,d}\bigr)^{2}\]

where the scale constant is

\[\sigma^{\mathrm{tr}}_{d} \;=\; \operatorname{sd}\bigl(\{\,\tilde{s}_{i,d}\;:\; \mathrm{episode}(i)\in\mathcal{E}_{\mathrm{train}}\,\}\bigr)\]

computed once, stored in \texttt{results/\allowbreak{}step4\_\allowbreak{}0a\_\allowbreak{}results\allowbreak{}.json}, and never recomputed per step or derived from held-out data. A value of one is not the training-mean floor (\S{}3.3); the floors are scored directly, above. \textbf{The aggregation is form 1}: pool the per-dimension mean squared errors, then divide, a ratio of means. A mean of per-dimension ratios gives whichever dimension has the smallest scale unbounded leverage; the rest of this appendix gives both forms, and supplement S32 the comparison the second one once inverted.

\textbf{Coverage at $\pm$k$\sigma$} is the fraction of scalar (trajectory, forecast step, state dimension) triples whose absolute realised error falls within k times the $\sigma$ predicted for that same triple:

\[\mathrm{cov}_{\pm k\sigma}(h) \;=\; \frac{1}{B\,h\,45}\sum_{b=1}^{B}\sum_{t=t_0+1}^{t_0+h}\sum_{d=1}^{45}\mathbf{1}\!\left[\;\frac{\bigl|\hat{s}^{(b)}_{t,d}-s^{(b)}_{t,d}\bigr|}{\sigma^{(b)}_{t,d}} \;\le\; k \;\right]\]

It is pooled over all three axes with equal weight per triple. It is \textbf{cumulative} over steps 1..h: coverage "at h" averages the whole rollout up to h and is not the value at step h, and every horizon-indexed quantity in this paper follows the same convention. Because it is built from an \emph{absolute} error, $z \le k$ is the two-sided event, so the calibrated targets are $\mathrm{erf}(k/\sqrt{2})$: \textbf{68.27\%} at $\pm$1$\sigma$ and \textbf{95.45\%} at $\pm$2$\sigma$. \textbf{That nominal is checked rather than assumed.} Rescaling each model's $\sigma$ by the single constant that makes mean$|$error$|$ / mean $\sigma$ equal a calibrated Gaussian's 0.7979 lands coverage above the rule's heavy-tail bound of 63.27\% in every one of 48 model $\times$ horizon cells, with 0 below it (15 within its band around 68.27\% and 33 above it; \texttt{M-65}), so the shortfalls reported below are a property of the scale and not of the tail.

\textbf{The overconfidence factor} is how many times larger the typical realised error is than the typical predicted $\sigma$:

\[\rho(h) \;=\; \frac{\operatorname{mean}_{b,t\le h,d}\bigl|\hat{s}^{(b)}_{t,d}-s^{(b)}_{t,d}\bigr|}{\operatorname{mean}_{b,t\le h,d}\ \sigma^{(b)}_{t,d}}\]

It too is a \textbf{ratio of means}, because a mean of ratios is unbounded whenever a single $\sigma$ approaches zero, which is exactly the regime \S{}5.5 puts these models in. $\rho = 1$ is \emph{not} calibration: a calibrated Gaussian has mean$|$error$|$ / $\sigma$ = $\sqrt{2/\pi}$ = 0.7979. So $\rho$ is reported as a magnitude of miscalibration and coverage as the calibrated reading, and both appear everywhere.

\textbf{The two aggregations.} \S{}3.3 states that this paper uses \textbf{form 1}: pool the per-dimension mean squared errors across the 45 state dimensions, then divide by the pooled scale --- a ratio of means. \textbf{Form 2} is the mean of per-dimension ratios.

The two differ because the state dimensions differ widely in scale. Form 2 gives whichever dimension has the smallest denominator unbounded leverage over the aggregate, and a dimension that is nearly constant in the training episodes has a very small denominator. Form 1 has no such lever: a dimension contributes in proportion to its share of the total squared error.

The definition in force throughout the paper is the one \S{}3.3 gives. Nothing in \S{}4.1, \S{}5 or \S{}6 uses form 2.

\section{Rollout training against teacher forcing in detail}

Detail behind \S{}4.1: the hold-last floor, seeds, the short-horizon reading, the data budget, the long-horizon cells and their multiplicity.

Figure 4 shows the autoregressive-versus-teacher-forcing advantage by forecast horizon, which \S{}4.1 summarises.

\begin{figure}[!ht]\centering\includegraphics[width=\linewidth]{\detokenize{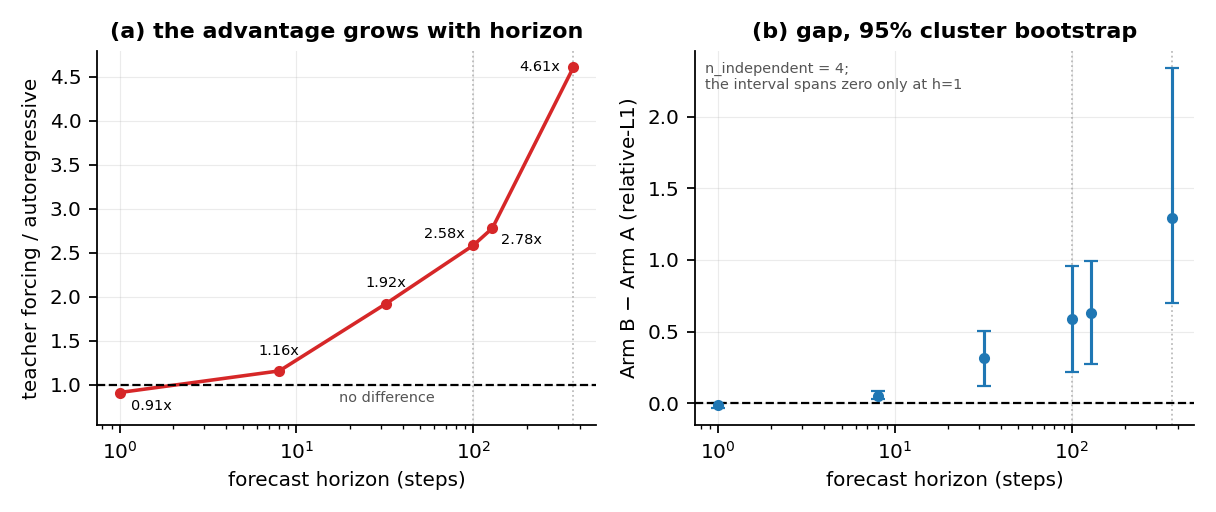}}\caption{The autoregressive-versus-teacher-forcing advantage as a function of forecast horizon, out-of-sample over three seeds at 10,000 training iterations, on the held-out pair's 4 independent 400-step trajectories. (a) the ratio, which grows monotonically with depth: h = 368 is the end of a trend rather than a selected point, and the method's own rollout length of h = 100 sits partway along it. (b) the same comparison as a gap with its 95\% cluster-bootstrap interval over whole trajectories; the interval spans zero only at h = 1, where teacher forcing is ahead -- a lead a shorter evaluation unit resolves as real, not nominal (\S5, M-64). Rule M-23 was run at h = 368 on seed 1 of each arm; these three-seed values, and every other horizon, were computed afterwards.}\end{figure}

\textbf{Three passages behind \S{}4.1.}

\emph{Against a baseline, because neither number means anything without one.} The hold-last floor, predicting that nothing changes, scores \textbf{0.9930} in the h = 368 cell, and autoregressive training beats it by \textbf{2.8$\times$} there and by 2.0$\times$ at h = 100. \textbf{Teacher forcing is 1.66$\times$ worse than assuming nothing changes at all} at h = 368, and 1.27$\times$ worse at h = 100: the arm that reaches a lower training loss predicts the future worse than a model that makes no prediction, at every horizon we measured on the 400-step unit, the sharper statement of what exposure bias costs here; but on M-64's shorter units, each 32 + h rows long and non-overlapping (60 at h = 1, 50 at h = 8, 30 at h = 32, 14 at h = 100 and 12 at h = 128 on these episodes), at the same checkpoint and on the same two episodes, it beats the floor at h = 1, 8, 32 and 100 and loses to it only at the longest horizon those units reach, so that statement holds on the 400-step unit and not on the shorter one. \textbf{The floor is not a weak baseline everywhere}: at h=1 it beats the autoregressive arm as well, scoring 0.0796 at h = 1 against autoregressive training's 0.1447, the only horizon where the autoregressive arm loses to predicting no change on the 400-step unit; on the 33-row unit both arms beat the floor at h = 1 (0.1315 and 0.1121 against 0.1687). At 50 Hz one step is 20 ms and the state barely moves, so that is what one should expect.

\emph{Seeds, and what four trajectories can show.} Seed spread is not symmetric between the arms: Arm A ranges 0.3341--0.3894 across seeds (7.9\% relative), Arm B 1.4241--1.9710 (17.3\%). Teacher forcing is more than twice as variable across seeds as autoregressive training at this horizon, so a single-seed comparison of these two arms is unreliable. For a single seed the bootstrap over trajectories gives the 95\% interval [0.56, 2.05] on n = 4 independent trajectories; by the n = 4 caveat of \S{}3 its tails move in steps of 0.39\%, so it corroborates the sign test rather than being the primary evidence. The four per-trajectory gaps behind \S{}4.1's three-seed comparison from longer training (Arm B minus Arm A, three seeds pooled, at h = 368) are \textbf{+2.8705, +0.8949, +0.7445, +0.6562}. All 4 of 4 are positive, the direction of the sign test's 10 of 10 episodes, but one trajectory carries +2.8705 against a smallest of +0.6562, which no interval on four units shows. At h = 100 the four are +1.0925, +0.8170, +0.2800, +0.1549; this appendix notes where other sections store theirs.

\textbf{What is small, and where it resolves.} At h = 8, the horizon the model is trained on, the advantage is small, and it resolves only with the longer training and all 3 seeds. The same comparison by horizon (Figure 4), at 10,000 iterations with 3 seeds pooled, gives an h = 8 gap of +0.0485 [+0.0271, +0.0844], which excludes zero, a factor of 1.16$\times$. The single seed the rule was run on (seed 1) gives an h = 8 gap of 0.008 at the same 10,000-iteration checkpoint, and its interval includes zero. At the 500 and 2,500-iteration checkpoints, with all 3 seeds pooled, the out-of-sample gap excludes zero in \textbf{0 of 4} h = 8 cells (both trajectory lengths crossed with both checkpoints). An earlier rule of ours, anchored at h = 8 and evaluated at those same checkpoints (rule M-16, Appendix E), returned \textbf{CANNOT BE SETTLED AT THIS BUDGET}. \textbf{The advantage is small at the training horizon and large beyond it.}

\textbf{The data budget (\S{}3.2).} \S{}3.2 states the result in one clause.

The base paper's headline is a sample-efficiency result: policies transfer to hardware from 6,000,000 state transitions of world-model pretraining against \textasciitilde{}250M for the model-free baseline (its Table 1, from v2 on). That is a claim about policy learning and hardware, which we cannot test, but its \emph{world-model} half is a quantity we can count exactly.

\textbf{Our arms consume 7,991 distinct state transitions}: the 7,999 rows of the eight training episodes less one per episode (8 of them), since an episode's first row has no predecessor, a transition being a consecutive pair of rows inside one episode. It is deliberately not the 7,687 training windows, which overlap almost completely (consecutive 40-row windows start one row apart), nor the 640,000 window draws a 2,500-iteration run makes, which resample the same data with replacement. Against the reference's 6,000,000, that is \textbf{751$\times$ less data, 0.133\% of its world-model budget}.

A dynamics model trained on 0.133\% of the reference's data still reproduces the training result, 4.61$\times$ at h = 368 and 2.58$\times$ at h = 100 over 3 seeds (the rule ran on seed 1), and still beats the hold-last floor, by 2.8$\times$ and 2.0$\times$ at those two horizons. That is what this paper can add to the sample-efficiency question without training a policy.

\textbf{Three limits.} It is not a reproduction of the 6,000,000-against-250M comparison, which is about policy learning. It says nothing about whether a policy trained inside our model would transfer to hardware, or anywhere. And our model is evaluated on the narrow distribution it trained on (one robot, one gait, one terrain, velocity commands from a single bounded box), where the reference's 6,000,000 transitions span considerably more; a smaller data budget buys less than it appears to when the evaluation distribution shrinks with it.

\section{Configuration and architecture claims in detail}

Detail behind \S{}4.2 and \S{}4.3: the baselines and their diverged rows, the RSSM diagnostic, the reimplementation beside the released checkpoint, and the settings sweep at equal compute.

\textbf{Two passages behind \S{}4.3.}

On the in-sample arena's 16 trajectories, with nRMSE pooled as \S{}3.3 defines it (computed afterwards, post hoc; ledger R-77), two readings differ from the per-trajectory average the rules' evaluator used. Teacher-forced, at h = 1 all three baselines are ahead of RWM: D, signed as the rules sign it (positive favours RWM), is -0.160 [-0.286, -0.002] for the MLP, -0.186 [-0.328, -0.013] for the RSSM and -0.052 [-0.089, -0.005] for the transformer, so M-75's reading there is DOES NOT REPRODUCE where the average gave CANNOT BE SETTLED: the same one-step lead for teacher forcing as \S{}4.1's. Trained autoregressively, the MLP's deficit at h = 100 is no longer resolved (+0.017 [-0.017, +0.044]), so M-76's reading there is PARTIAL where the average gave RWM AHEAD OF ALL THREE. Neither changes a verdict; both rules govern at h = 368 on relative-L1.

\textbf{Our RSSM is not an informative comparison.} Teacher-forced, it is the most accurate of \S{}4.3's models one step ahead, 0.0761 against RWM's 0.1232 and the floor's 0.0796, though not resolvably, and it has collapsed open-loop by h = 32, where its 1.0182 is above the floor's 0.5950. The original adds that an RSSM trained autoregressively performs comparably to RWM; ours does not, 10.6405 against RWM's 0.5856 at h = 368. A diagnostic committed before its readings existed (rule X1, ledger M-80; exploratory, it re-opens neither rule) asks why. Its Part A reads the forecast differently, feeding the prior's expected or sampled latent in place of its most likely one: the error at h = 32 falls to 0.7165 and 0.6914, still above the floor, so it returns \textbf{NOT EVALUATION-LIMITED} (M-81). Its Part B is descriptive: over the history, the teacher-forced RSSM's one-step error from its prior is 1.16$\times$ to 1.27$\times$ its error from its posterior at the same recurrent state (the autoregressive one's at most 1.02$\times$), with 16.4 to 20.0 nats of KL divergence between them per step. Its Part C retrains the teacher-forced RSSM at seed 0 for 2,500 iterations, once with PlaNet's KL settings and once with DreamerV2's layer-normalised recurrent cell. Neither rescues it: read from its most likely latent at h = 32, they score 1.2588 and 1.4685 against the floor's 0.5950, with 3 and 2 of the 4 trajectories above their own floor value; a rescue needs the mean below the floor's and every trajectory below its own. So rule X1 returns \textbf{NOT RESCUED BY THE SETTINGS TRIED} (ledger M-83). The failure may still be our RSSM rather than the architecture: Table S7's latent is ambiguous, we read it in DreamerV2's naming (\texttt{BASELINE\_SPECS.md}, the RSSM rows), and X1 tried only the two settings above, on one seed, neither of them PlaNet's latent overshooting (Hafner et al., ICML 2019), which trains the prior's multi-step predictions in latent space and which PlaNet's own final RSSM agent did not need. So the architecture claim rests on the MLP and the transformer.

\textbf{Teacher forcing's two definitions.}

On the original's heatmap, for the teacher-forced (32, 1) it prints 3.99 against 0.47 at the centre, 8.5$\times$ worse, on evaluation data and at a horizon it does not state; our sweep's (32, 1) is 6.35$\times$ worse on relative-L1 at h = 368 and 2,500 iterations. Our 4.61$\times$, at 10,000, uses a different definition of teacher forcing (\texttt{docs/\allowbreak{}presubmission/\allowbreak{}ORIGINAL\_\allowbreak{}SPECS\allowbreak{}.md} \S{}2); the next paragraph relates the two.

Arm B predicts each of the window's 8 forecast targets from true inputs, where the original's teacher forcing is N = 1; the sweep's (32, 1), trained that way, is 6.35$\times$ worse than the centre at h = 368 at 2,500 iterations (the sweep behind \S{}4.2), so the direction holds under both definitions.

\textbf{The diverged baseline rows (\S{}4.3).} † marks a diverged row, one where any seed's mean relative-L1 at h = 368 exceeds 10 times the hold-last floor's (\texttt{results/\allowbreak{}pooled\_\allowbreak{}nrmse\_\allowbreak{}rescore\allowbreak{}.json}, post hoc). Run-away rollouts dominate those three rows' means; per seed they are MLP, teacher-forced: 182, 16.0 and 238; transformer, teacher-forced: 11.2, 10.4 and 23.9; RSSM, autoregressive: 9.87, 10.8 and 11.3. No verdict depends on that magnitude, only on the sign: each such row's 4 per-trajectory differences from RWM are all positive, so every bootstrap resample favours RWM whatever their size.

\textbf{The long-horizon cells.} At long horizons \S{}4.1's pattern is consistent across the design. Under the cluster bootstrap, the out-of-sample gap excludes zero in \textbf{4 of 4} long-horizon cells, both trajectory lengths crossed with the 500 and 2,500-iteration checkpoints. These figures are relative-L1; the nRMSE aggregation is reported separately and does not change the direction.

\textbf{Multiplicity.} Those 4 cells sit in a family of 8 out-of-sample comparisons. All 4 of 4 still exclude zero at a Bonferroni level of 0.05/8, and Holm--Bonferroni rejects \textbf{4 of 4}. \S{}4.1's sign test is unaffected either way.

\textbf{How good the reimplementation is as a model, next to the artifact it reimplements.} \S{}4.1 compares two training rules with each other and \S{}5.2's table compares calibration, so neither puts the released checkpoint and our arms side by side on absolute accuracy. Both aggregations, for these models and for \S{}4.3's architecture baselines, on one arena:

\begin{center}\small\resizebox{\ifdim\width>\linewidth\linewidth\else\width\fi}{!}{%
\begin{tabular}{lllllllll}
\toprule
model & nRMSE h = 1 & rel-L1 h = 1 & nRMSE h = 8 & rel-L1 h = 8 & nRMSE h = 100 & rel-L1 h = 100 & nRMSE h = 368 & rel-L1 h = 368 \\
\midrule
released checkpoint & 0.0544 & 0.0563 & 0.0697 & 0.0808 & 0.5028 & 0.3304 & 0.9051 & 0.6041 \\
Arm A --- autoregressive, faithful MSE & 0.1262 & 0.1232 & 0.3029 & 0.3289 & 0.4913 & 0.4798 & 0.5425 & 0.5856 \\
Arm A --- autoregressive, \texttt{gaussian\_nll} & 0.1204 & 0.1195 & 0.2867 & 0.3154 & 0.4603 & 0.4413 & 0.5245 & 0.5608 \\
Arm B --- teacher-forced & 0.0879 & 0.0929 & 0.3064 & 0.3392 & 1.0363 & 1.0849 & 3.7927 & 4.5684 \\
MLP, teacher-forced (\S{}4.3) † & 0.1393 & 0.1539 & 0.3624 & 0.4086 & 1.1399 & 1.2057 & 112.5344 & 145.1650 \\
MLP, autoregressive (\S{}4.3) & 0.1364 & 0.1470 & 0.3016 & 0.3492 & 0.5452 & 0.5400 & 1.4383 & 1.4628 \\
RSSM, teacher-forced (\S{}4.3) & 0.0707 & 0.0761 & 0.3393 & 0.3686 & 2.0376 & 2.2128 & 5.3890 & 6.3718 \\
RSSM, autoregressive (\S{}4.3) † & 0.1223 & 0.1354 & 0.6004 & 0.6211 & 3.1128 & 3.2660 & 9.2508 & 10.6405 \\
transformer, teacher-forced (\S{}4.3) † & 0.1202 & 0.1239 & 0.3773 & 0.3969 & 2.0195 & 2.0607 & 13.4730 & 15.1548 \\
transformer, autoregressive (\S{}4.3) & 0.1329 & 0.1384 & 0.2562 & 0.3025 & 0.5821 & 0.5821 & 2.0110 & 1.8771 \\
hold-last floor & 0.0989 & 0.0796 & 0.4117 & 0.3298 & 0.9537 & 0.7558 & 1.0897 & 0.9930 \\
\bottomrule\end{tabular}}\end{center}

\textbf{Arena, stated once for the whole table: out-of-sample held-out pair, episodes 1 and 8, 4 non-overlapping 400-step trajectories, n\_independent = 4; possibly in-sample for the released checkpoint (\S{}3.2).} Arm rows are the mean over 3 seeds at 2,500 training iterations (the \texttt{weights\_2500.pt} checkpoint), with per-seed values in \texttt{results/\allowbreak{}head\_\allowbreak{}to\_\allowbreak{}head\_\allowbreak{}accuracy\allowbreak{}.json}; nRMSE is form 1 (\S{}3.3), and both metrics are cumulative over forecast steps 1..h. Every RWM row is read from the stored rollouts behind Appendix I's held-out calibration tables, so no model is run to build it. The architecture-baseline rows (\S{}4.3) come from their own evaluator on the same four trajectories, whose relative-L1 reproduces the Arm A row exactly; their nRMSE is pooled as \S{}3.3 defines it, recomputed afterwards with the same evaluator (post hoc, \texttt{results/\allowbreak{}pooled\_\allowbreak{}nrmse\_\allowbreak{}rescore\allowbreak{}.json}), and † marks a diverged row (\S{}4.3). \textbf{This table is at 2,500 iterations and \S{}4.1's longer-training comparison (Figure 4) at 10,000}, which is why Arm A's relative-L1 at h = 368 reads 0.5856 here and 0.3582 there: the same arm, trained longer.

Both metrics put the released checkpoint first at h = 1 and h = 8, they name different leaders at h = 100, and at h = 368 both put an Arm A variant ahead of it: the reimplementation is behind the artifact it reimplements at short horizons and ahead of it at the longest horizon we measure. That reading flatters the released checkpoint, because the split is ours: the arena is out-of-sample for our arms and in-sample for it, which trained on all ten episodes (the in-sample caveat of \S{}3).

\emph{Stored per-trajectory values elsewhere.} \S{}5.3's and supplementary S2's paired contrasts at the held-out pair's n = 4 store their four per-trajectory values in \texttt{results/\allowbreak{}r2\_\allowbreak{}independent\_\allowbreak{}ensemble\allowbreak{}.json} and \texttt{results/\allowbreak{}m49\_\allowbreak{}capacity\_\allowbreak{}matched\allowbreak{}.json}; Appendix I's two held-out tables (the four models and the ensemble-5 arms), at n\_independent = 4, give intervals only, coarse for the same reason (the n = 4 caveat of \S{}3).

\textbf{Rule X3, fold by fold (\S{}4.1).} Rule X3 was committed and pushed before any of its runs existed (ledger M-86; Appendix E gives its lead time) and is discharged in M-94. It is exploratory and re-opens no rule: rule M-23 stands. The gap is Arm B's seed-mean relative-L1 minus Arm A's at h = 368. Each fold's interval is exact over all 256 ordered resamples of its trajectories; the pooled one is M-23's three-seed bootstrap, 20,000 draws (seed 0), whole trajectories, every seed carried.

\begin{center}\small\resizebox{\ifdim\width>\linewidth\linewidth\else\width\fi}{!}{%
\begin{tabular}{llll}
\toprule
fold (episodes held out) & trajectories & gap at h = 368 [95\% interval] & each episode's mean gap \\
\midrule
\{1, 8\} (section 5) & 4 & +3.9828 [+0.9907, +9.4848] & 1: +6.9749, 8: +0.9907 \\
\{0, 2\} & 4 & +0.6554 [+0.5172, +0.8630] & 0: +0.5512, 2: +0.7596 \\
\{3, 4\} & 4 & +1.4966 [+1.0403, +1.9528] & 3: +1.0403, 4: +1.9528 \\
\{5, 6\} & 4 & +1.6282 [+1.1379, +2.4999] & 5: +2.0714, 6: +1.1850 \\
\{7, 9\} & 4 & +3.3743 [+3.1063, +3.5926] & 7: +3.1561, 9: +3.5926 \\
all five, pooled & 20 & \textbf{+2.2275 [+1.3683, +3.5053]} & 10 of 10 positive, P = 0.002 \\
\bottomrule\end{tabular}}\end{center}

\textbf{Arena: each fold's held-out pair, out-of-sample for both arms, 4 non-overlapping 400-step trajectories per fold and 20 pooled (n\_independent = 20). Checkpoints: Arm A and Arm B at 2,500 iterations, 3 seeds per arm per fold.} Bold, the reading, \textbf{REPLICATES ACROSS FOLDS}. The per-episode sign test is part of the rule and carries no verdict of its own: 10 of 10 episodes positive, exact two-sided P = 0.002. Alongside: at h = 100 the pooled gap is +0.3573 [+0.2624, +0.4678], and Arm B's mean error is 5.85$\times$ Arm A's at h = 368. The folds are fixed, not drawn: two of the new ones hold both their episodes from the same speed half, unlike the held-out pair, which is stratified by speed.

Post hoc (ledger R-78; \texttt{results/\allowbreak{}mn\_\allowbreak{}compute\_\allowbreak{}matched\allowbreak{}.json}); it re-opens no rule, and rule M-74's verdict, NOT OPTIMAL AT OUR BUDGET, stands as a statement about the 2,500-iteration budget (\S{}4.2). Each row sets one of the four configurations that beat the centre at 2,500 iterations against the centre trained to 2,500, 5,000 or 10,000 iterations: twelve rows of four readings each. A longer forecast costs more per iteration than the centre, a shorter history less, so the centre's longer runs ask whether a configuration's win survives giving the centre at least as much training computation.

\textbf{How \S{}4.2's alongside readings were computed, and how far its runs had trained.}

The rules' evaluator averaged nRMSE per trajectory, where \S{}3.3 pools it. The nRMSE readings quoted alongside the rules here and in \S{}4.3 are pooled, recomputed afterwards (post hoc, \texttt{results/\allowbreak{}pooled\_\allowbreak{}nrmse\_\allowbreak{}alongside\allowbreak{}.json}), and, for \S{}4.2's rule (M-74), return the same verdicts as the averaged ones (the passages behind \S{}4.3, above, give the two readings of the baseline rules that differ).

All 48 runs at 2,500 iterations, the sweep's, the baselines' and both arms', are still lowering their training loss at the end, with slopes from -1.99 to -0.09 per thousand iterations (\texttt{results/\allowbreak{}training\_\allowbreak{}tail\_\allowbreak{}slopes\allowbreak{}.json}, post hoc), so the ranking is at this budget, not at convergence, and it changes with training length (below).

\textbf{\S{}4.2's reading, in full.}

\textbf{Accuracy at equal compute (post hoc; ledger R-78).} A longer training forecast costs more per iteration, 1.35$\times$ the centre's for (32, 16) and 1.81$\times$ for (32, 32); the shorter histories that win cost less, 0.40$\times$ for (8, 8) and 0.26$\times$ for (2, 8) (\texttt{results/\allowbreak{}mn\_\allowbreak{}compute\_\allowbreak{}matched\allowbreak{}.json}; the table below gives every reading). The shorter histories win on all four readings at 2,500 iterations while using well under half the centre's computation, so that part of the verdict needs no compute correction. The longer forecasts' wins do. Trained to 5,000 iterations, the centre has had 1.11$\times$ the computation of (32, 32), and the readings split: (32, 32) is still ahead on the held-out pair at h = 368 (-0.2031 [-0.5033, -0.0275]), the centre is ahead on the in-sample arena at h = 100 (+0.0454 [+0.0356, +0.0563]), and the other two are not resolved. At 10,000 iterations, 2.21$\times$ the computation, the centre is ahead on both in-sample readings and neither held-out difference is resolved. Trained that long, it also passes both shorter histories on three of the four readings: on the held-out pair at h = 368, (2, 8) by +0.1135 [+0.0983, +0.1357] and (8, 8) by +0.1270 [+0.0987, +0.1589], with 15$\times$ and 10$\times$ their computation, and on both in-sample readings; on the held-out pair at h = 100 neither difference is resolved. Which setting is most accurate therefore depends on how long each is trained. None of this re-opens rule M-74.

\begingroup\small
\begin{longtable}{>{\raggedright\arraybackslash}p{\dimexpr 0.1655\linewidth-2\tabcolsep\relax}>{\raggedright\arraybackslash}p{\dimexpr 0.1533\linewidth-2\tabcolsep\relax}>{\raggedright\arraybackslash}p{\dimexpr 0.1958\linewidth-2\tabcolsep\relax}>{\raggedright\arraybackslash}p{\dimexpr 0.1213\linewidth-2\tabcolsep\relax}>{\raggedright\arraybackslash}p{\dimexpr 0.1213\linewidth-2\tabcolsep\relax}>{\raggedright\arraybackslash}p{\dimexpr 0.1213\linewidth-2\tabcolsep\relax}>{\raggedright\arraybackslash}p{\dimexpr 0.1213\linewidth-2\tabcolsep\relax}}
\toprule configuration (cost per iteration ÷ the centre's) & centre trained to (iterations) & compute ratio (centre's total ÷ configuration's) & h = 100, held-out pair (4) & h = 100, in-sample (16) & h = 368, held-out pair (4) & h = 368, in-sample (16) \\ \midrule\endfirsthead
\toprule configuration (cost per iteration ÷ the centre's) & centre trained to (iterations) & compute ratio (centre's total ÷ configuration's) & h = 100, held-out pair (4) & h = 100, in-sample (16) & h = 368, held-out pair (4) & h = 368, in-sample (16) \\ \midrule\endhead
\bottomrule\endlastfoot
(2, 8) (0.26) & 2,500 & 3.78 & \textbf{-0.1274 [-0.2175, -0.0373]} & \textbf{-0.0219 [-0.0341, -0.0107]} & \textbf{-0.1139 [-0.2988, -0.0088]} & \textbf{-0.0219 [-0.0364, -0.0081]} \\
(2, 8) (0.26) & 5,000 & 7.55 & -0.0742 [-0.1586, +0.0101] & \textbf{+0.0619 [+0.0448, +0.0813]} & -0.0301 [-0.1976, +0.0711] & \textbf{+0.0927 [+0.0681, +0.1230]} \\
(2, 8) (0.26) & 10,000 & 15.11 & -0.0176 [-0.0921, +0.0475] & \textbf{+0.1085 [+0.0846, +0.1367]} & \textbf{+0.1135 [+0.0983, +0.1357]} & \textbf{+0.1403 [+0.1075, +0.1802]} \\
(8, 8) (0.40) & 2,500 & 2.50 & \textbf{-0.0755 [-0.1410, -0.0269]} & \textbf{-0.0142 [-0.0239, -0.0053]} & \textbf{-0.1004 [-0.2449, -0.0167]} & \textbf{-0.0208 [-0.0462, -0.0009]} \\
(8, 8) (0.40) & 5,000 & 4.99 & -0.0223 [-0.0782, +0.0244] & \textbf{+0.0696 [+0.0538, +0.0870]} & -0.0166 [-0.1434, +0.0719] & \textbf{+0.0937 [+0.0762, +0.1123]} \\
(8, 8) (0.40) & 10,000 & 9.99 & +0.0343 [-0.0198, +0.0719] & \textbf{+0.1162 [+0.0924, +0.1436]} & \textbf{+0.1270 [+0.0987, +0.1589]} & \textbf{+0.1414 [+0.1163, +0.1710]} \\
(32, 16) (1.35) & 2,500 & 0.74 & \textbf{-0.1254 [-0.2054, -0.0454]} & \textbf{-0.0438 [-0.0675, -0.0258]} & \textbf{-0.2509 [-0.5218, -0.0956]} & \textbf{-0.1003 [-0.1355, -0.0719]} \\
(32, 16) (1.35) & 5,000 & 1.48 & -0.0723 [-0.1466, +0.0020] & \textbf{+0.0400 [+0.0326, +0.0481]} & \textbf{-0.1671 [-0.4301, -0.0117]} & \textbf{+0.0142 [+0.0068, +0.0205]} \\
(32, 16) (1.35) & 10,000 & 2.96 & -0.0157 [-0.0707, +0.0393] & \textbf{+0.0866 [+0.0729, +0.1005]} & -0.0235 [-0.1158, +0.0388] & \textbf{+0.0618 [+0.0531, +0.0706]} \\
(32, 32) (1.81) & 2,500 & 0.55 & \textbf{-0.1450 [-0.2490, -0.0409]} & \textbf{-0.0384 [-0.0637, -0.0186]} & \textbf{-0.2869 [-0.5951, -0.1005]} & \textbf{-0.1062 [-0.1422, -0.0768]} \\
(32, 32) (1.81) & 5,000 & 1.11 & -0.0918 [-0.1901, +0.0065] & \textbf{+0.0454 [+0.0356, +0.0563]} & \textbf{-0.2031 [-0.5033, -0.0275]} & +0.0084 [-0.0006, +0.0161] \\
(32, 32) (1.81) & 10,000 & 2.21 & -0.0352 [-0.1142, +0.0438] & \textbf{+0.0920 [+0.0770, +0.1068]} & -0.0594 [-0.1903, +0.0340] & \textbf{+0.0560 [+0.0483, +0.0639]} \\
\end{longtable}\endgroup

\textbf{Arenas: the held-out pair (episodes 1 and 8, 4 non-overlapping 400-step trajectories, n\_independent = 4) and the in-sample arena (the training episodes' 16 non-overlapping 400-step trajectories, n\_independent = 16). Checkpoints: each configuration at 2,500 iterations, the centre at the iterations in the second column; three seeds each.} Each cell is D, the configuration's relative-L1 minus the centre's, the mean over trajectories of the three-seed means, with its 95\% interval: exact over all 256 ordered resamples on the held-out pair, 20,000 Monte Carlo resamples (seed 0) in-sample. Negative favours the configuration; bold where the interval excludes zero. Cost per iteration is steady training time per iteration, timed without contention; the compute ratio is the centre's total training computation over the configuration's.

\section{Calibration in detail}

Calibration behind \S{}5.2 and \S{}5.6: every horizon, the held-out-pair version, the ensemble-5 arms, the checks on how the numbers are read, and the per-dimension tests.

\textbf{Scope of the per-member readings.} Per-member propagation is not the released method. MEAN tests "effectively propagated" as the release's evaluation path does it; PER-MEMBER asks whether the ensemble's information would propagate if the members were rolled out separately. The release's imagination path, which propagates one member per environment, is not tested.

Figure 5 shows the four models' calibration on the held-out pair, the arena of the four-model table below.

\begin{figure}[!ht]\centering\includegraphics[width=\linewidth]{\detokenize{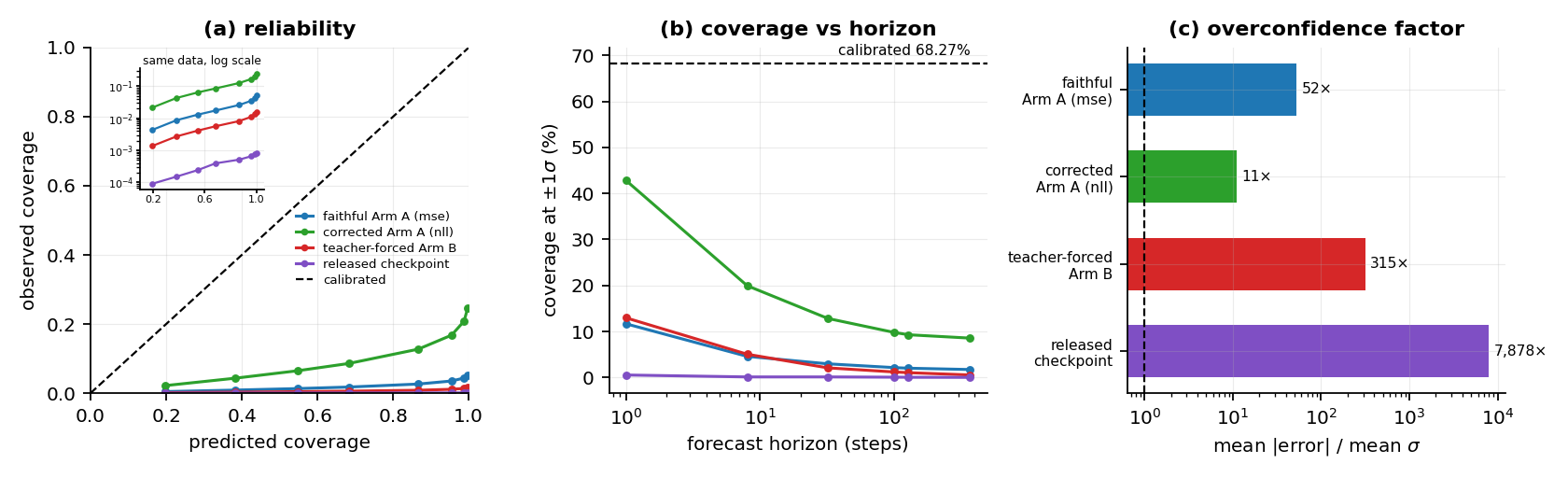}}\caption{Calibration of all four models on the held-out pair's 4 independent 400-step trajectories, our arms at 2,500 training iterations. The released checkpoint trained on these episodes, so for it this arena is in-sample (\S3). (a) reliability: observed against predicted coverage, with the calibrated diagonal; the inset repeats the same points with observed coverage on a log scale, where the four models separate. (b) coverage at $\pm1\sigma$ against forecast horizon, log scale, against the 68.27\% a calibrated Gaussian gives. Every curve sits far below the diagonal and falls further with horizon. (c) the overconfidence factor for each model: mean absolute error divided by mean predicted $\sigma$, log scale, with the dashed line where error equals $\sigma$.}\end{figure}

\textbf{The held-out-pair version (\S{}5.2).} \S{}5.2 measures the released checkpoint on all ten episodes: restricting it to the held-out pair would buy no independence, since its training rows are not recorded (\S{}3.2), and would cost four fifths of the sample. That version, at n\_independent = 4 400-step trajectories, is in the supplementary material (\texttt{results/\allowbreak{}task\_\allowbreak{}b2\_\allowbreak{}epistemic\allowbreak{}.json}). Its epistemic column agrees in direction with the all-ten-episodes table below at all 6 of 6 horizons; its aleatoric column agrees at 4 of 6 and differs at h=8 and h=100, and the aleatoric $\sigma$ is in any case 1,827$\times$ (h = 1) to 20,669$\times$ (h = 368) too small for its ordering to be the interesting quantity.

\textbf{The four-model table.} On the held-out pair, the four models stood as follows.

\begin{center}\small\resizebox{\ifdim\width>\linewidth\linewidth\else\width\fi}{!}{%
\begin{tabular}{llll}
\toprule
model & mean $|$error$|$ / mean $\sigma$, whole 368-step rollout & coverage at $\pm$1$\sigma$, h=1 & coverage at $\pm$1$\sigma$, h=100 \\
\midrule
faithful Arm A (sampled MSE) & 52.2$\times$ [39.6, 69.7] & 11.67\% [7.96, 15.19] & 2.17\% [1.59, 2.75] \\
corrected Arm A (\texttt{gaussian\_nll}) & 10.9$\times$ [7.8, 14.7] & 42.78\% [24.44, 62.04] & 9.8\% [6.80, 12.81] \\
teacher-forced Arm B & 315$\times$ [177, 509] & 12.96\% [7.04, 20.93] & 1.22\% [0.84, 1.61] \\
released checkpoint & 7,878$\times$ [5,410, 9,934] & 0.56\% [0.00, 1.67] & 0.08\% [0.06, 0.11] \\
\bottomrule\end{tabular}}\end{center}

\emph{Our arms are at 2,500 training iterations; the released checkpoint is as released. Every cell carries a 95\% interval from a cluster bootstrap over whole trajectories, n\_independent = 4, quantised as the n = 4 caveat of \S{}3 describes; where three seeds contribute, seeds are pooled inside each draw rather than resampled, because seeds are not trajectories (\S{}3.4).}

\textbf{All four rows are the held-out pair} --- the two episodes withheld from our own arms, n\_independent = 4 400-step trajectories --- because that is the only arena on which our arms can be scored fairly. On the aleatoric head every model is overconfident by between one and four orders of magnitude (Figure 5); that is the quantity \S{}5.1 shows the method discards. The released checkpoint's 7,878$\times$ is its whole 368-step rollout on those same 4 trajectories, for comparability with the arms; its best-sampled figure is the 11,683$\times$ below, cumulative to h = 100 at n\_independent = 20. \textbf{Both are correct, on a different arena and a different horizon}, and neither is out-of-sample for that checkpoint (the in-sample caveat of \S{}3).

\textbf{The released checkpoint at every horizon.} Over all 10 episodes, n\_independent = 20 non-overlapping 400-step trajectories, the released checkpoint as released:

\begin{center}\small\resizebox{\ifdim\width>\linewidth\linewidth\else\width\fi}{!}{%
\begin{tabular}{llllllll}
\toprule
h & aleatoric err/$\sigma$ [95\% CI] & aleatoric $\pm$1$\sigma$ & epistemic err/$\sigma$ [95\% CI] & epistemic $\pm$1$\sigma$ [95\% CI] & epistemic $\pm$2$\sigma$ & dims r$>$0 & permutation P \\
\midrule
1 & 1,827$\times$ [915, 3,110] & 0.11\% & \textbf{8.3$\times$} [6.1, 9.7] & 16.22\% [13.44, 18.89] & 30.11\% & 44/45 & 0.0056 \\
8 & 3,034$\times$ [1,638, 4,849] & 0.08\% & 15.1$\times$ [10.3, 19.2] & 9.99\% [8.60, 11.32] & 19.76\% & 45/45 & 0.0069 \\
32 & 4,525$\times$ [2,492, 7,196] & 0.07\% & 22.6$\times$ [15.5, 28.7] & 6.95\% [5.99, 7.85] & 13.68\% & 45/45 & 0.0344 \\
\textbf{100} & 11,683$\times$ [7,970, 15,846] & 0.04\% & \textbf{33.4$\times$} [28.7, 39.0] & 4.61\% [4.12, 5.09] & 9.27\% & 45/45 & 0.2769 \\
128 & 14,934$\times$ [10,564, 19,700] & 0.03\% & 34.2$\times$ [30.6, 38.2] & 4.37\% [3.96, 4.81] & 8.75\% & 45/45 & 0.3762 \\
368 & 20,669$\times$ [15,666, 25,688] & 0.02\% & 34.4$\times$ [29.8, 40.3] & 3.59\% [3.27, 3.92] & 7.19\% & 45/45 & 0.0804 \\
\bottomrule\end{tabular}}\end{center}

\textbf{The larger sample changes one thing materially.} At n\_independent = 4 the epistemic ordering looked like chance at short horizon, 23 of 45 dimensions at h=1. At n\_independent = 20 it is 44 of 45 at h=1, with mean r = +0.662, the \emph{strongest} mean correlation of any horizon, and the in-sample permutation test agrees (below). The short-horizon "chance" result was an artifact of four trajectories, not a property of the model.

\textbf{The ensemble-5 arms (\S{}5.2).} \textbf{The released checkpoint is no longer the only ensemble measured.} Three Arm A arms at ensemble size 5 (3 seeds, out-of-sample, n\_independent = 4 400-step trajectories, 2,500 training iterations) give, averaged over seeds:

\begin{center}\small\resizebox{\ifdim\width>\linewidth\linewidth\else\width\fi}{!}{%
\begin{tabular}{lllll}
\toprule
h & epistemic err/$\sigma$ [95\% CI] & $\pm$1$\sigma$ [95\% CI] & $\pm$2$\sigma$ & dims r$>$0 \\
\midrule
1 & 2.1$\times$ [1.7, 2.5] & 35.93\% [29.63, 42.22] & 60.37\% & 35.0/45 \\
8 & 6.3$\times$ [4.4, 7.5] & 14.10\% [11.02, 17.18] & 26.62\% & 37.3/45 \\
32 & 8.0$\times$ [5.9, 9.8] & 10.79\% [8.53, 13.71] & 20.50\% & 34.3/45 \\
100 & \textbf{10.5$\times$} [9.0, 11.5] & 8.19\% [6.80, 9.59] & 16.11\% & 44.0/45 \\
128 & 11.0$\times$ [9.5, 11.9] & 7.79\% [6.81, 8.77] & 15.34\% & 44.0/45 \\
368 & \textbf{13.0$\times$} [11.9, 13.9] & 6.30\% [5.84, 6.77] & 12.51\% & 40.7/45 \\
\bottomrule\end{tabular}}\end{center}

Our arms are \textbf{better calibrated than the released checkpoint and fail the same way}: 10.5$\times$ overconfident at h = 100 against its 25.7$\times$ on the same held-out pair (33.4$\times$ over all ten episodes), with 8.19\% coverage where a calibrated Gaussian gives 68.27\%. Appendix K establishes that the two are the same architecture in the respect that matters here, so this is a comparison of like with like. Being closer to calibrated is not being calibrated.

\textbf{Two pre-registered checks on how the calibration and ranking numbers are read.} The first (rule M-62, Appendix E) asks whether any verdict depends on resampling 400-step trajectories rather than whole episodes, which two trajectories share. It returns \textbf{NO MOVE}: in the one arena with power at that level, all ten episodes (n = 20 falling to 10), the pooled correlation's interval widens by 27\% and the double-demeaned one's width changes by a factor of 0.98, and neither crosses zero. The other 4 cells it names are out-of-sample, where an episode bootstrap has n = 2 and three distinct resamples, or, for the per-horizon multiplier cell, one episode per direction and no interval at all; the rule said so in advance, and they are reported as uninformative rather than as intervals.

\textbf{The second (rule M-63, Appendix E) asks whether the one-step failure is a few bad channels or all of them}, since a pooled coverage is the unweighted mean of 45 per-dimension ones. It returns \textbf{UNIFORM}: the interquartile range across dimensions is 10.0 points against a 15-point threshold committed in advance, and the five worst dimensions carry well under half of the shortfall. \textbf{No channel is exempt.} The reading is coarse by construction: at h = 1 on 20 trajectories a per-dimension coverage moves in 5-point steps, a limit the rule fixed before the run.

The last column of the released-checkpoint table above gives permutation P-values over whole trajectories, not binomial ones, on the same 20 trajectories as the counts beside them; Appendix C explains why a binomial null is inadmissible here. h = 100 is tested too, because it carries the abstract's headline figure. These are six tests on one family and none survives Holm--Bonferroni across the arena's 30 cells: the smallest is faithful (mse) h=368 at 0.0037 against a threshold of 0.001667. Read the column as a consistency check on direction, not as six independent findings.

Adding the teacher-forced arm, trained for \S{}4.1, sharpens the finding (our arms at 2,500 training iterations; the released checkpoint as released):

\begin{center}\small\resizebox{\ifdim\width>\linewidth\linewidth\else\width\fi}{!}{%
\begin{tabular}{lllll}
\toprule
model & $\sigma$ variation across inputs (CoV) & dims with r($\sigma$, error) $>$ 0 at h=368, held-out pair & perm P at h=368, held-out pair & perm P at h=368, training episodes \\
\midrule
faithful Arm A & 0.0076 & 39/45 & 0.0417 & 0.0240 \\
corrected Arm A & 0.0059 & 21/45 & 1.0000 & 0.8746 \\
\textbf{teacher-forced Arm B} & \textbf{0.1188} & \textbf{45/45} & \textbf{0.2609} & \textbf{0.5565} \\
released checkpoint & 0.0177 & 20/45 & 0.6957 & 0.9753 \\
\bottomrule\end{tabular}}\end{center}

\textbf{The CoV column is the aleatoric $\sigma$ in every row}, the only $\sigma$ the ensemble-size-1 arms have. Our ensemble-5 arms (above) have both: their aleatoric CoV is comparable to the other arms', and their \emph{epistemic} term is far more input-dependent than any aleatoric head here, at 0.379--0.395 against the released checkpoint's 0.0177. \textbf{The count column is the out-of-sample arena} (n\_independent = 4), so that all four models are compared on trajectories none of our arms trained on. For the released checkpoint's aleatoric head it is not the most informative arena: at h = 368 and n\_independent = 20 over all ten episodes that head is 0/45, negatively correlated with error on \emph{every} dimension, against 20/45 here; supplementary S3 quotes the larger arena.

Arm B's $\sigma$ is 15.6$\times$ more input-dependent than the faithful arm's, and it has the largest mean correlation of the four (r = 0.257). It is still 315$\times$ overconfident.

\textbf{The P column is a permutation P}, which pairs each trajectory's $\sigma$ with a different trajectory's realised error and so keeps the coupling between dimensions and the growth of error with depth that every trajectory shares. The binomial P-values an earlier draft attached to these counts are withdrawn (\texttt{S-15}), and the correction is large (below).

So $\sigma$ \emph{collapsing in magnitude} is objective-driven, and $\sigma$ \emph{becoming input-independent} is not. The teacher-forced arm collapses in magnitude exactly like the autoregressive ones --- same objective, same rate --- while retaining 15.6$\times$ more variation across inputs. Input-independence is a property of the autoregressive arms and the released checkpoint, and input-dependence and correct ranking are both achievable without the interval becoming meaningful. One candidate mechanism, stated as an untested hypothesis: autoregressive feedback narrows the input distribution toward the model's own manifold, leaving a heteroscedastic head less variation to key on.

\textbf{The same pattern holds for the quantity the method uses, and this is where the correction bites hardest.} At h=128 and h=368 the epistemic term correlates positively with realised error on \textbf{45 of 45} dimensions, matching the best aleatoric head here on the sign count, while being 39.7$\times$ overconfident at h = 368. \textbf{Figures in this paragraph are the held-out arena (n\_independent = 4) unless labelled otherwise}, so the epistemic term and the four aleatoric heads are compared on identical trajectories; the released-checkpoint table above gives 34.4$\times$ for the same ratio at h = 368 and n\_independent = 20. Nor does it beat Arm B's head on strength: its mean correlation at h=368 is +0.151 against 0.257. The two rank comparably, and neither is close to an interval. Under the permutation null that count gives P = 0.0435 on the held-out pair and 0.0775 on the training episodes, both possibly in-sample for the checkpoint (\S{}3.2), against 5.68e-14 from the independent-trials test. It still fails the horizon test the same way: $\sigma$ grows 1.59$\times$ from h=1 to h=368 while error grows 13.33$\times$.

\textbf{Across arenas, and after correction for multiplicity.} The two larger arenas, which are nested (every in-sample trajectory is among the all-ten ones), put the epistemic ordering's strength at \emph{short} horizon, because at long horizon the forecast-depth trend every trajectory shares lifts the null until a full count is close to chance, which is what motivates \S{}5.6's index control. Nothing here survives Holm--Bonferroni in any of the three arenas, and out of sample at the 400-step unit nothing could, since its smallest attainable P already exceeds the smallest Holm threshold (below).

\textbf{The structural excuse, with its table.} \textbf{The structural excuse does not survive.} One could argue that a model trained on an 8-step horizon cannot be expected to report calibrated uncertainty about step 368. It cannot report it about step 8 either. Inside the trained horizon, $\sigma$ is flat while error grows (Figure 6; out-of-sample held-out pair, n\_independent = 4 400-step trajectories, possibly in-sample for the released checkpoint (\S{}3.2); our arms at 2,500 training iterations, the released checkpoint as released):

\begin{figure}[!ht]\centering\includegraphics[width=\linewidth]{\detokenize{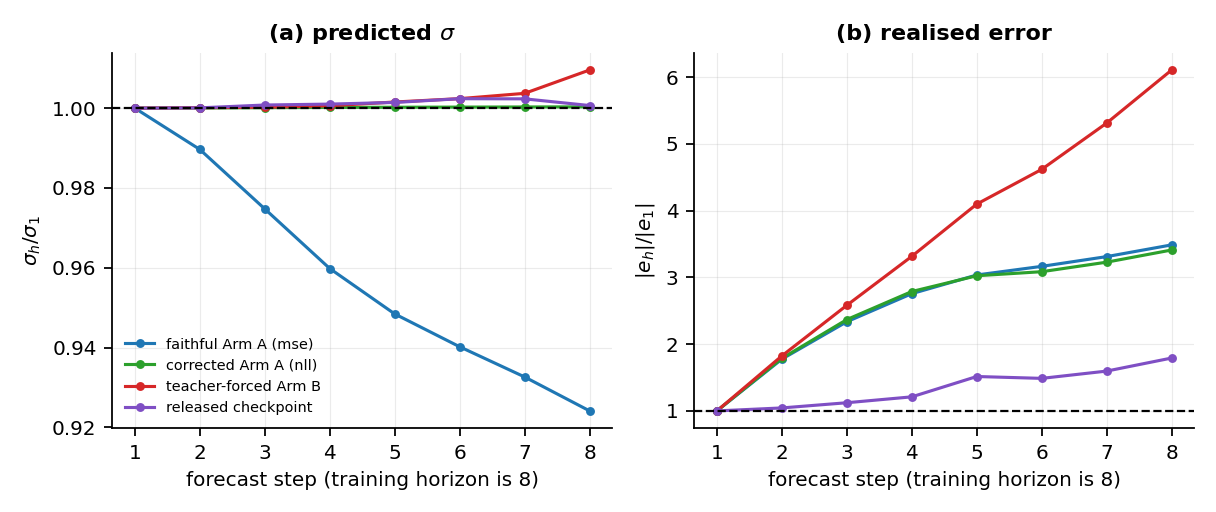}}\caption{Why the coverage collapse is a horizon effect. Both panels are normalised to forecast step 1, on the out-of-sample held-out pair, n\_independent = 4 400-step trajectories; our arms at 2,500 training iterations, the released checkpoint as released; it trained on these episodes, so for it this arena is in-sample (\S3). (a) predicted $\sigma$ barely moves, and for the faithful arm it declines. (b) realised error grows 1.79× to 6.11× over the same steps, across the four models. The gap between the panels is the collapse.}\end{figure}

\begin{center}\small\resizebox{\ifdim\width>\linewidth\linewidth\else\width\fi}{!}{%
\begin{tabular}{lll}
\toprule
model & $\sigma$ growth, step 1 → 8 & error growth, step 1 → 8 \\
\midrule
faithful Arm A & 0.9241$\times$ & 3.49$\times$ \\
corrected Arm A & 1.0003$\times$ & 3.41$\times$ \\
teacher-forced Arm B & 1.0096$\times$ & 6.11$\times$ \\
released checkpoint & 1.0007$\times$ & 1.79$\times$ \\
\bottomrule\end{tabular}}\end{center}

The faithful arm's $\sigma$ \emph{declines} (0.9241$\times$) while its error grows 3.49$\times$. The coverage collapse in Figure 5(b) is therefore driven entirely by growing error against a fixed $\sigma$.

\textbf{The permutation tests.} \textbf{The P columns above are permutation P-values, not binomial ones}, and the binomial P-values an earlier draft attached to their counts are withdrawn (\texttt{S-15}). A binomial null treats the 45 state dimensions as independent trials, and they are not: position, velocity and torque for the same joint are physically coupled, and base linear and angular velocity are coupled through the gait. More importantly, error grows with rollout depth in every trajectory, so \emph{any} $\sigma$ that also grows with depth correlates with \emph{any} trajectory's error, including one it was never paired with.

We therefore permute whole trajectories. The null pairs each trajectory's $\sigma$ with a different trajectory's realised error, which keeps both marginal distributions and the entire cross-dimension dependence structure and destroys only the association under test. The correction is large, and largest exactly where we leaned hardest. The worst-affected cell is teacher-forced armB at h=368, in the in-sample arena. At h = 368 it moves from 5.68e-14 to 0.5565, a factor of about 10\textsuperscript{13}, because under a null that keeps the dependence a random re-pairing already yields 43.8 of 45 dimensions positive on average, so observing 45 of 45 is close to unremarkable. A fair coin centres the count at 22.5 of 45; the dependence-preserving null centres it between 5.1 and 43.8 depending on model, horizon and arena.

\textbf{The larger arenas agree with each other against the smallest.} At n\_independent = 4 400-step trajectories on the held-out pair, possibly in-sample for the checkpoint (\S{}3.2), the epistemic ordering looks strongest at long horizon (0.0417 at h=128, 0.0435 at h=368) and unremarkable at short (0.4348 at h=1). Both larger arenas invert that. In sample (n\_independent = 16): 0.0052 at h=1, 0.0070 at h=8, against 0.3794 at h=128. Over all ten episodes (n\_independent = 20): 0.0056, 0.0069 and 0.3762, with h = 100 at 0.2769, between h=32's 0.0344 and h=128's 0.3762. Two larger arenas at four and five times the sample say the effect is strongest at \emph{short} horizon, though they are not independent: all 16 in-sample trajectories are among the 20.

The null means explain why, and the explanation is the one that motivates \S{}5.6. At long horizon the shared forecast-depth trend lifts the null to 41.6 of 45 at h = 128, so a count of 45 is close to what chance alone delivers; at h = 1 the null sits near 15.4 and the same count is genuinely surprising. The out-of-sample arena is not wrong so much as blind: at 4 trajectories its smallest attainable P-value is 0.04167, so it cannot tell a strong effect from a marginal one at any horizon. \textbf{That blindness belongs to the 400-step unit, not to the arena}: a 33-row unit gives 60 independent units on the same two episodes, and 14 at h = 100 (rule M-64, Appendix E). We did not recompute this permutation family at the shorter unit, so what the short units change is the scope of the design claim, not any verdict: the arena is underpowered at h = 368, which needs the full 400 rows, and is no longer demonstrably so at h = 128 and below. Running it there is one pass over stored rollouts, and we did not do it. The small arena's numbers are reported because it is the only arena out-of-sample for our own arms, not because it is the better measurement.

\textbf{Nothing here survives multiplicity correction, in any of the three arenas.} Holm--Bonferroni over each arena's 30 model $\times$ horizon cells at $\alpha$ = 0.05 rejects 0 out of sample, 0 in sample and 0 over all ten episodes. Out of sample that is a property of the design: with 4 independent trajectories the smallest attainable P-value is 0.04167, which already exceeds the smallest Holm threshold 0.001667, so no effect of any size could have been rejected there. In sample the miss is real: the smallest P in the family is teacher-forced armB h=128 at 0.0027 against a threshold of 0.001667.

\section{Why the per-member $\sigma$ collapses}

Evidence behind \S{}5.5: the synthetic-noise test, the term-by-term gradient check, the variance-state arithmetic and the PETS lineage with the survey counts.

\textbf{The corrected objective under the permutation test.} Under Appendix I's trajectory permutation test, at h = 368, the faithful and corrected arms' sign counts (Appendix I's table of $\sigma$ against error) give P = 0.0417 and 1.0000 out of sample, 0.0240 and 0.8746 in sample. The faithful arm's ordering is the one result in this family that points the same way in both arenas; it is also the weakest effect of the three, and it does not survive multiplicity correction either.

\textbf{\S{}5.5's derivation says the collapse happens on any dataset, and that is testable.} It matters because it is what answers the follow-up's own explanation: on the released CSV, "small stochasticity in the environment" and our reading are observationally identical, since the data may simply be nearly deterministic. So under a rule committed before the runs (rule M-50, Appendix E) we built data where it is not.

Synthetic data whose true noise level is \textbf{known} and varies by a factor of 25 across the input range, with a non-constant true mean; the \textbf{same} bounded log-$\sigma$ head as the released model --- \texttt{MLPStateHead} unmodified, including the double-softplus clamp, the learnable \texttt{state\_min\_logstd} and \texttt{state\_log\_delta\_logstd}, and the bound loss at its configured weight --- trained under each objective in turn on 4,000 points for 12,000 iterations at 3 seeds. Nothing else differs between the arms.

\emph{Two ways this is not the released setting.} The head is built here over a \textbf{one-dimensional} state with no recurrent trunk in front of it, where the released one predicts 45 dimensions from a GRU. The trunk's absence is deliberate: the question is about the head's objective, and a GRU would add a confound. The dimensionality matters because the state loss sums over state dimensions, so at one dimension it is roughly 45$\times$ smaller relative to the bound term than in the released path. That makes this setting \emph{more} favourable to $\sigma$ surviving, and the collapse happens anyway.

\begin{center}\small\resizebox{\ifdim\width>\linewidth\linewidth\else\width\fi}{!}{%
\begin{tabular}{llll}
\toprule
objective & median $\hat{\sigma}$ / $\sigma$\_true & $\hat{\sigma}$ spread across the input range & slope of log $\hat{\sigma}$ on log $\sigma$\_true \\
\midrule
\texttt{mse} --- the implemented branch & \textbf{0.0460} & 1.002--1.003$\times$ & +0.000493 \\
\texttt{gaussian\_nll} --- the authors' unused branch & 0.9779 & 1.02--3.67$\times$ & +0.1330 \\
\bottomrule\end{tabular}}\end{center}

\emph{Ratios and slopes are means over 3 seeds; spreads are the range across them, because the mean of a spread hides which seeds recovered.}

\textbf{Under the implemented objective $\sigma$ sits 21.7$\times$ below the true noise and does not track it at all}: a spread of 1.003$\times$ where the truth spans 25$\times$, and a slope below the 0.00309 the design can detect. Under the authors' own branch, same data and same head, $\sigma$ recovers the true level to a median ratio of 0.9779, and every one of the 3 seeds the rule was discharged over clears the slope threshold. \textbf{Twenty seeds show that clearance is not general}: 11 of 20 clear it, so the all-seeds criterion would not have held at that sample. The rule's verdict stands as returned over its own 3 and is not re-opened by more seeds (\S{}3.4); what twenty establish is that the hedge below was necessary. \textbf{The recovering arm is seed-variable, and the rule said so before the runs}: its slopes span 0.0078--0.3537, a factor of 45, and its 3 seeds recover $\sigma$ spreads of only 1.02--3.67$\times$ against the truth's 25$\times$. So what this experiment establishes is the \textbf{contrast}, that one objective tracks the noise at all and the other does not, and not the magnitude of the recovery, which this training budget does not pin down. \textbf{OBJECTIVE-DRIVEN}, which is the verdict the rule names for that pattern.

\emph{The statistic is the slope, not the correlation, because a correlation is scale-free: a $\hat{\sigma}$ that is essentially constant still returns a large one off its own numerical noise. Under a permutation null, the same data with the input-to-noise pairing destroyed, a head whose $\sigma$ spanned 1.0004$\times$ returned correlations as large as $\pm$0.24, while its slope was 2e-05. The detection threshold is set at the slope corresponding to a 1.01$\times$ spread rather than at that noise floor, and the measured false-positive rate at zero signal is 0\%.}

\textbf{What this does that \S{}5.5's derivation alone could not.} It removes the competing explanation rather than arguing against it: the stochasticity here is large, known and input-dependent, and the collapse happens anyway. The design's limit, stated in the rule, holds: the dilution ladder detects the signal at full strength and at no dilution below it, so this establishes that $\sigma$ does not track the noise \textbf{at all}, not the magnitude of how badly.

\textbf{\S{}5.5's derivation covers two terms, and the objective has 7.} Its completeness rests on the other 5 being inert with respect to $\sigma$, so each term is computed alone on one real batch (64 windows from the training episodes, at freshly initialised weights of the released architecture) and back-propagated alone, and the gradient reaching the log-$\sigma$ tower, \texttt{state\_log\_delta\_logstd} and \texttt{state\_min\_logstd} is recorded. A term that cannot move $\sigma$ produces exactly zero on all three.

\begin{center}\small\resizebox{\ifdim\width>\linewidth\linewidth\else\width\fi}{!}{%
\begin{tabular}{lllllll}
\toprule
loss term & live? & weight & where the reference computes it & $\partial$/$\partial$ log-$\sigma$ tower & $\partial$/$\partial$ \texttt{log\_delta\_logstd} & $\partial$/$\partial$ \texttt{min\_logstd} \\
\midrule
\texttt{state} & live & 1.00 & \texttt{system\_dynamics.py:270-284} & 0.000325 & 0.0509 & 0.0703 \\
\texttt{sequence} & \textbf{dead} & 1.00 & \texttt{system\_dynamics.py:272-279} & 0 & 0 & 0 \\
\texttt{bound} & live & 1.00 & \texttt{system\_dynamics.py:301-302} & 0 & 0.2 & 0 \\
\texttt{kl} & \textbf{dead} & 0.10 & \texttt{system\_dynamics.py:205} & 0 & 0 & 0 \\
\texttt{extension} & \textbf{dead} & 1.00 & \texttt{system\_dynamics.py:233-268} & 0 & 0 & 0 \\
\texttt{contact} & live & 1.00 & \texttt{system\_dynamics.py:233-268} & 0 & 0 & 0 \\
\texttt{termination} & live & 1.00 & \texttt{system\_dynamics.py:233-268} & 0 & 0 & 0 \\
\bottomrule\end{tabular}}\end{center}

4 of the 7 configured terms are live at all under the released configuration: \texttt{sequence\_loss} is dead code, guarded by a \texttt{prediction\_type} the reference sets to \texttt{"single"} on both paths, and \texttt{extension} is zero because its dimension is, while \texttt{kl} belongs to the RSSM branch only. Of the 7, exactly 2 reach $\sigma$: \texttt{state} and \texttt{bound}. The remaining 5 produce a gradient of exactly zero, not merely a term the code suggests is irrelevant (\texttt{results/\allowbreak{}e4\_\allowbreak{}sigma\_\allowbreak{}gradients\allowbreak{}.json}).

We predicted the collapse from \S{}5.5's algebra before training, then observed it. Three run counts appear below and they are not the same set. This project trained 33 runs for \S{}4.1--\S{}6, besides the 42 of \S{}4.2 and \S{}4.3 (supplement S7), of which 28 are at the released \texttt{rnn\_hidden\_size} of 256 and form the collapse family; the remaining 5 are the capacity-matched arm of rule M-49 (Appendix E) at width 124, a different architecture, excluded from every rate \S{}5.5 quotes (supplement S7). Across all 28 runs of that family the log-$\sigma$ range moves linearly in iteration count at a nearly identical rate within each objective, falling under sampled MSE and rising under the corrected one (Figure 7a). Rates are fitted on 22 of those 28: the 6 10,000-iteration runs continue seeds already counted at 2,500 and would double-weight them. Figure 7(a) shows all 28 runs of the collapse family and Figure 7(b) only the 22 the rate is fitted on, so the scatter and \S{}5.5's quoted statistic describe the same set. The 33 runs, with the width column separating the collapse family from the capacity-matched arm:

\begin{figure}[!ht]\centering\includegraphics[width=\linewidth]{\detokenize{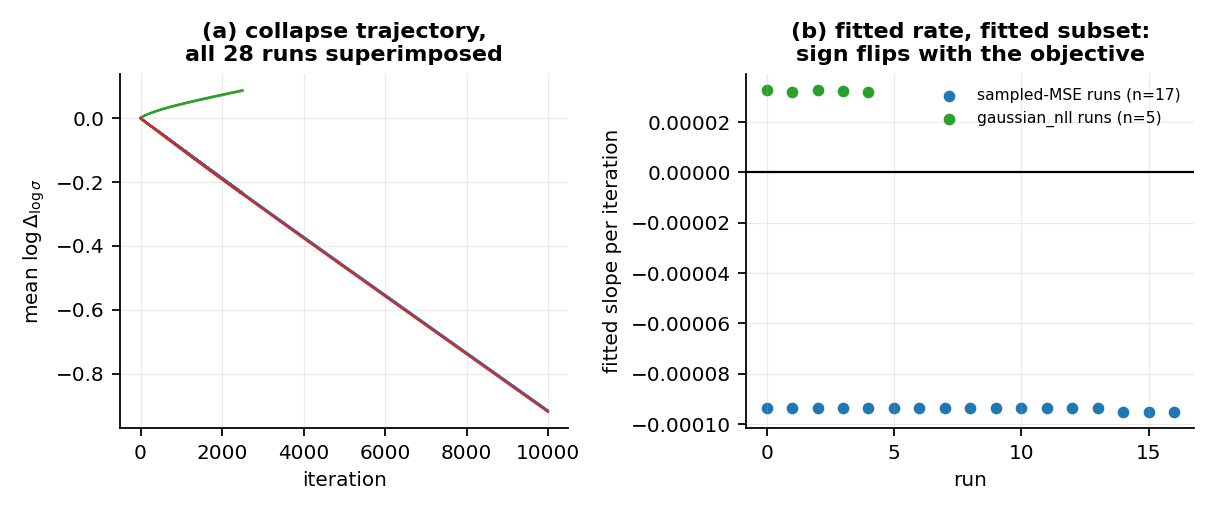}}\caption{The variance collapse is objective-driven. (a) mean $\log\Delta_{\log\sigma}$ against training iteration for each of the 28 runs of the collapse family, which is every run of §5–§7 at the released width outside §4.2's sweep (Appendix H). The runs are drawn individually but are visually coincident within each objective, so the 28 read as two lines, one falling and one rising -- which is the point: the trajectory does not vary visibly from run to run. (b) the fitted per-iteration slope for each run, grouped by objective: negative and tightly clustered under sampled MSE, positive under \texttt{gaussian\_nll}. The sign flip is the evidence that the objective, not the optimiser or the data, produces it.}\end{figure}

\begin{center}\small\resizebox{\ifdim\width>\linewidth\linewidth\else\width\fi}{!}{%
\begin{tabular}{llllllll}
\toprule
arm & iterations & ensemble & objective & dataset & width & seeds & seed ids \\
\midrule
Arm A & 2,500 & 1 & gaussian\_nll & clean & 256 & 5 & 0, 1, 2, 3, 4 \\
Arm A & 2,500 & 1 & mse & clean & 124 & 5 & 0, 1, 2, 3, 4 \\
Arm A & 2,500 & 1 & mse & clean & 256 & 5 & 0, 1, 2, 3, 4 \\
Arm A & 2,500 & 1 & mse & contaminated & 256 & 3 & 0, 1, 2 \\
Arm A & 2,500 & 1 & mse & duplicated & 256 & 3 & 0, 1, 2 \\
Arm A & 2,500 & 5 & mse & clean & 256 & 3 & 0, 1, 2 \\
Arm A & 10,000 & 1 & mse & clean & 256 & 3 & 0, 1, 2 \\
Arm B & 2,500 & 1 & mse & clean & 256 & 3 & 0, 1, 2 \\
Arm B & 10,000 & 1 & mse & clean & 256 & 3 & 0, 1, 2 \\
\bottomrule\end{tabular}}\end{center}

\textbf{The variance-state arithmetic.} \S{}6 states the conclusion. The arithmetic and the five assumptions it rests on are in \texttt{docs/\allowbreak{}APPENDIX\_\allowbreak{}G\_\allowbreak{}VARIANCE\_\allowbreak{}ARITHMETIC\allowbreak{}.md}, shipped as supplementary, kept out of the body because the numbered claim it once supported is retracted (\texttt{S-19}) and because a forensic case the section then defuses with the author's own reply is not what a reader needs in the body of a reproduction.

So the finding is not that the release is internally inconsistent. It is that \textbf{the released artifacts do not reproduce the released checkpoint's variance state, and the author's account is that the released repository is not the one that trained it.} That is a \textbf{documentation gap between a release and a run}: common, worth recording, and much less interesting than an inconsistency. The $\sigma$ collapse is linear in iteration count and its rate is nearly identical across our runs (\S{}5.5), which makes it a clock, and read as a clock it puts the checkpoint's variance state out of reach of a constant-rate run from the released initialisation at the configured learning rate, at every iteration count the release, the paper and the checkpoint tag state. The first author's account is that the released repository is several revisions removed from the setup that trained the checkpoint, which supplies a mechanism, a warm start or a different \texttt{log\_delta\_logstd} initialisation, that would explain it with no inconsistency at all. We report the arithmetic because it is what let us detect the gap at all, not as a charge against the work.

\textbf{Where the parameterisation comes from.} The bounded log-$\sigma$ head that \S{}5.5 shows has its optimum at $\sigma$ = 0 is largely not this codebase's invention: its clamp is inherited, line for line, from the probabilistic ensembles of Chua, Calandra, McAllister and Levine (PETS, NeurIPS 2018). PETS's Appendix A.1 gives

\begin{verbatim}
logvar = max_logvar - softplus(max_logvar - logvar)
logvar = min_logvar + softplus(logvar - min_logvar)
\end{verbatim}

and \texttt{architectures/\allowbreak{}mlp\allowbreak{}.py:92-93} is those two lines in log-standard-deviation rather than log-variance, with \texttt{system\_dynamics.py:302} supplying PETS's regulariser on the bounds.

What is \textbf{not} inherited is the objective, and one line of the bounds. PETS uses "the negative log prediction probability as our loss function", and that likelihood's log-$\sigma$ term opposes $\sigma$ → 0; \texttt{system\_dynamics.py:283} substitutes squared error on a reparameterised sample, which has no such term. And \texttt{architectures/\allowbreak{}mlp\allowbreak{}.py:91} builds the upper bound as the floor plus a learned positive gap, where PETS keeps the two bounds independent. \S{}5.5 needs both changes: the substitution removes the log-$\sigma$ term, and the tie is why the bound regulariser does not take its place, since the floor cancels out of it. In a descendant that kept PETS's independent bounds, the same regulariser pushes the floor up while the objective pulls $\sigma$ down onto it, a case \S{}5.5 does not derive and we have not tested (ledger \texttt{M-72}). So \textbf{a descendant of this lineage that replaced the likelihood with a sampled squared error, and left nothing pushing its variance floor back up, would inherit the same optimum.} That is a hypothesis about mechanism, untested in any other descendant (\S{}7).

\textbf{How often the substitution is made, which narrows the hypothesis.} Of 10 public repositories examined on 20 September 2026 under the protocol in \texttt{results/\allowbreak{}q1\_\allowbreak{}search\_\allowbreak{}protocol\allowbreak{}.md}, 10 carry the construction and 1 of those trains it against a sampled squared error, and that one only in an optional value-aware mode; the other 9 keep PETS's likelihood (\texttt{results/\allowbreak{}q1\_\allowbreak{}pets\_\allowbreak{}descendants\allowbreak{}.json}). The protocol counts a repository only if it has the bounded head, learnable bounds, and a loss that squares the error of a \emph{sampled} prediction; squaring the error of the predicted \emph{mean}, an option several of these repositories offer, leaves $\sigma$ untrained rather than driving it to zero, and does not count. We read "learnable" as "trainable by the code, by default or not", a reading settled after the survey, because in \texttt{mbrl-lib} and in \texttt{va\_mbpo}, the one repository that inherits, the bounds train only when a caller switches that on (ledger \texttt{M-73}). The survey did not check how each repository treats its variance floor, so 1 of 10 is an upper bound on how often both conditions hold. The survey looked at 13 in all; the other 3 lack the construction, so the protocol does not count them as examined, among them mainline \texttt{rsl\_rl}: the bounded head exists in the fork this paper pins, not in the library it forks, and we do not count the fork, since counting what \S{}5.5 measured as evidence that the result travels would be circular. The protocol capped the survey at 25 repositories; it stopped at 10 examined, 13 looked at, and its notes give no reason. We wrote the protocol before the search, but neither it nor its hash reached git before the results did, so that order rests on our own record; it is a search protocol, not one of Appendix E's decision rules. Search engines rank and truncate, so this is a sample of convenience, but on it the substitution is rare, and a reader who takes the hypothesis to reach widely infers more than the evidence supports.

\section{The shared trunk and independent ensembles}

The shared trunk behind \S{}5.3, from the source and for our own arms, and the two pre-registered comparisons that test it: independent members, and independent members with the corrected objective.

\textbf{Source walkthrough.} \texttt{system\_dynamics.py:34} builds \textbf{one} \texttt{state\_base}. \texttt{system\_dynamics.py:35-41} replicates the \emph{heads} \texttt{ensemble\_size} times, and only the heads. In the forward pass \texttt{system\_dynamics.py:87} evaluates the trunk \textbf{once} and \texttt{:90} hands the identical feature vector to every head; \texttt{system\_dynamics.py:126} then computes the epistemic term as the standard deviation across those heads.

\textbf{Our own arms.} \textbf{This applies to our own arms identically, which is why comparing them with the released checkpoint (Table 3, Appendix I) is fair.} Our ensemble-5 arms build one trunk the same way (\texttt{src/\allowbreak{}rwm\_\allowbreak{}model\allowbreak{}.py:164-167}), evaluate it once (\texttt{:182}), hand the same vector to every head (\texttt{:185}) and compute the same spread (\texttt{:200}). Their tensor names and parameter counts match the released checkpoint exactly --- 636,672 shared, 77,492 per head, 1,995,569 in total, on all 3 arms checked (\texttt{results/\allowbreak{}v1\_\allowbreak{}ensemble\_\allowbreak{}topology\allowbreak{}.json}). So Appendix I's finding that our arms "fail the same way", 10.5$\times$ overconfident at h = 100, compares two instances of one architecture, not two architectures.

\textbf{The table for rule M-44.}

\begin{center}\small\resizebox{\ifdim\width>\linewidth\linewidth\else\width\fi}{!}{%
\begin{tabular}{lllll}
\toprule
h & independent err/$\sigma$ & shared-trunk err/$\sigma$ & independent $\pm$1$\sigma$ & shared-trunk $\pm$1$\sigma$ \\
\midrule
1 & 1.4$\times$ & 2.1$\times$ & 50.00\% & 35.93\% \\
8 & 3.4$\times$ & 6.3$\times$ & 22.64\% & 14.10\% \\
32 & 4.4$\times$ & 8.0$\times$ & 17.92\% & 10.79\% \\
\textbf{100} & \textbf{5.2$\times$} & \textbf{10.5$\times$} & \textbf{15.31\%} & \textbf{8.19\%} \\
128 & 5.3$\times$ & 11.0$\times$ & 15.03\% & 7.79\% \\
368 & 5.1$\times$ & 13.0$\times$ & 15.00\% & 6.30\% \\
\bottomrule\end{tabular}}\end{center}

\emph{Same trajectories (the held-out pair), same harness, n\_independent = 4, every model at 2,500 training iterations. The shared-trunk column is the mean over 3 seeds; the comparison below is paired against each of them separately.}

\S{}5.3 establishes the topology as a fact and the mechanism as a hypothesis. Rule M-44 tests the hypothesis, under a rule committed to git before any of the artifacts below existed (rule M-44, Appendix E), with a power check estimating what it could detect at the sample size it would face.

\textbf{The contrast, and why it is affordable.} Training 5 genuinely independent models from scratch costs about 4.8 h of wall clock on two cores at these runs' iteration count, against the 49.8 h supplement S7 gives for the 33 runs of \S{}4.1--\S{}6 it times (\S{}4.2's sweep and \S{}4.3's baselines are counted apart). Arm A at ensemble size 1 already existed at seeds 0, 1 and 2; we added two more at about 0.9 h each, 1.7 h in total, and scored the 5 together as an ensemble \textbf{at evaluation time}. There is no new training code and no new architecture, and the disagreement across 5 independently initialised \emph{full models} is exactly the contrast \S{}5.3 asks for. The rollout protocol mirrors the shared-trunk one in every respect except the one under test: each member sees the same input state, each keeps \textbf{its own} recurrent hidden state, and the ensemble mean is fed back to all of them.

\textbf{The rule returns MECHANISM SUPPORTED.} All 6 of its 6 conditions hold, against every one of the 3 shared-trunk seeds. At h = 100 the independent ensemble's overconfidence factor is 0.485--0.503$\times$ the shared-trunk arms', a \textbf{2.03$\times$} improvement against a pre-registered minimum detectable effect of 1.45$\times$, and its $\pm$1$\sigma$ coverage is 6.69 to 7.33 points higher, a mean of +7.12 against an MDE of 2.26. Every paired interval excludes zero. \textbf{Members that share a feature extractor do produce a smaller spread, and the effect is large enough to matter.}

\textbf{Where the improvement comes from, which is not all one thing.} The overconfidence factor is error over $\sigma$, so it improves if $\sigma$ grows \emph{or} if error shrinks, and only the first is the trunk-sharing mechanism; five independent models also denoise better than five heads on one trunk, an ordinary ensembling effect. Splitting the improvement into its two multiplicative parts:

\begin{center}\small\resizebox{\ifdim\width>\linewidth\linewidth\else\width\fi}{!}{%
\begin{tabular}{llllll}
\toprule
h & $\sigma$ larger by & error smaller by & total & share from $\sigma$ & share from accuracy \\
\midrule
1 & 1.56$\times$ & 0.97$\times$ & 1.52$\times$ & 106\% & -6\% \\
8 & 1.81$\times$ & 1.02$\times$ & 1.84$\times$ & 97\% & 3\% \\
32 & 1.68$\times$ & 1.08$\times$ & 1.83$\times$ & 87\% & 13\% \\
\textbf{100} & \textbf{1.65$\times$} & 1.23$\times$ & 2.03$\times$ & \textbf{71\%} & 29\% \\
128 & 1.62$\times$ & 1.29$\times$ & 2.09$\times$ & 65\% & 35\% \\
368 & 1.49$\times$ & 1.70$\times$ & 2.53$\times$ & 43\% & 57\% \\
\bottomrule\end{tabular}}\end{center}

\emph{On the held-out pair's 4 independent trajectories, independent against shared-trunk ensembles, all at 2,500 iterations. Shares are of the log improvement, so they add to 100\%. A share above 100\% at h=1 means the independent ensemble is very slightly the \textbf{worse} predictor there and the $\sigma$ gain more than covers it.}

\textbf{The reading, at the horizon the rule is stated over.} $\sigma$ is larger at every horizon, by 1.49$\times$ at its weakest (h = 368) and 1.81$\times$ at its strongest (h = 8), the direction trunk-sharing predicts, and at h = 100 it is 1.65$\times$, \textbf{71\%} of the improvement. So the mechanism is supported, and it is the larger part of the effect where the method operates. At the open-loop diagnostic horizon of h = 368 the split reverses: 57\% of the improvement there is the ensemble simply predicting better, so a reader who takes the 2.53$\times$ figure at that horizon as a measure of the architectural effect would overstate it.

\textbf{What this does and does not license.} It licenses saying that \textbf{the released ensemble's disagreement understates epistemic uncertainty partly because its members are not independent models}, with a measured size at the horizon that matters. It does not license attributing the whole gap to trunk-sharing: independently seeded runs differ in \emph{both} initialisation and data ordering (supplementary S2), so this comparison \textbf{bounds} the architectural effect rather than isolating it, and the bound is generous to the mechanism by construction.

\textbf{And it does not repair the interval.} The independent ensemble is 5.2$\times$ overconfident at h = 100 with 15.31\% coverage where a calibrated Gaussian gives 68.27\%. Better by a factor of 2.03, and still not an interval. Building the ensemble properly is worth doing and it is not sufficient; \S{}5.7's per-horizon multiplier remains the only correction in this paper whose estimates all land within the band, though only on the released checkpoint and no single cell is resolvable at this arena. On Arm A, whose model never saw the test episodes, it manages 17 of 36 epistemic cells (\S{}5.7).

\textbf{The table for rule M-68.}

\begin{center}\small\resizebox{\ifdim\width>\linewidth\linewidth\else\width\fi}{!}{%
\begin{tabular}{lllllll}
\toprule
h & combined err/$\sigma$ & shared-trunk err/$\sigma$ & combined $\pm$1$\sigma$ & shared $\pm$1$\sigma$ & combined $\pm$2$\sigma$ & shared $\pm$2$\sigma$ \\
\midrule
1 & 1.4$\times$ & 2.1$\times$ & 51.11\% & 35.93\% & 78.89\% & 60.37\% \\
8 & 3.5$\times$ & 6.3$\times$ & 20.62\% & 14.10\% & 39.72\% & 26.62\% \\
32 & 4.4$\times$ & 8.0$\times$ & 17.97\% & 10.79\% & 32.66\% & 20.50\% \\
\textbf{100} & \textbf{5.4$\times$} & \textbf{10.5$\times$} & \textbf{15.53\%} & \textbf{8.19\%} & 28.27\% & 16.11\% \\
128 & 5.5$\times$ & 11.0$\times$ & 15.39\% & 7.79\% & 27.97\% & 15.34\% \\
368 & 5.4$\times$ & 13.0$\times$ & 14.44\% & 6.30\% & 27.63\% & 12.51\% \\
\bottomrule\end{tabular}}\end{center}

\emph{Same trajectories (the held-out pair, n\_independent = 4 400-step trajectories), same harness, the same bootstrap unit, whole trajectories, as M-44's table above, every model at 2,500 training iterations. The shared-trunk columns are the mean over 3 seeds; the comparison below is paired against each of them separately.}

\S{}5.3 changes the \textbf{topology} and holds the objective at \texttt{mse}; \S{}5.5's arms change the \textbf{objective} and hold the topology at ensemble size 1, where disagreement is zero by construction. Neither answers the practitioner's question of what the two give together, and adding two effects never measured on the same model is the arithmetic this paper criticises elsewhere. Rule M-68 runs the combination, under a rule committed to git before either new member existed (rule M-68, Appendix E), with a minimum detectable effect estimated in advance as well.

\textbf{The arm.} 5 independently initialised full models, sharing no trunk and no hidden state, trained under \texttt{gaussian\_nll}, every setting identical to \S{}5.3's arm except the loss type. Three of the five, seeds 0, 1 and 2, already existed under that objective; two more were trained, and the 5 are scored together as an ensemble at evaluation time under M-44's rollout protocol, on the same held-out arena of non-overlapping 400-step trajectories at n\_independent = 4, over the same six horizons.

\textbf{Which $\sigma$ the coverage is against.} The combined arm is the first arm in this paper carrying \textbf{both} an aleatoric head that has not collapsed to zero and an across-member epistemic spread. The figures below are the \textbf{epistemic} one, the standard deviation across the five members' mean predictions, taken per state dimension, rather than the aleatoric head's output: the quantity the rule names and M-44's table above reports, so the two tables are comparable line for line. The aleatoric head does not enter here.

\textbf{The rule returns THE COMBINATION IMPROVES CALIBRATION}, branch 2 of the four it names. All 5 of its 5 conditions hold, against every one of the 3 shared-trunk seeds. At h = 100 the combined arm's overconfidence factor is 0.506--0.525$\times$ the shared-trunk arms', a \textbf{1.94$\times$} improvement against a minimum detectable effect of 1.218$\times$ fixed in advance (results/p3\_combined\_arm\_power.json, binding\_mde at h = 100 (largest of the four calibrations)); its $\pm$1$\sigma$ coverage is 6.92 to 7.56 points higher, a mean of +7.34 against an MDE of 2.50 points; and every paired interval excludes zero.

\textbf{Condition (e), and how we read it.} The rule's fifth condition asks that both statistics move in the improving direction at at least four of the six horizons, "against every shared-trunk seed", and the rule states globally that a condition holds only if it holds against all three. We applied the strict reading: a horizon counts only when \textbf{both} statistics improve there against \textbf{all three} seeds. It holds at 6 of 6 horizons, so the looser per-seed reading would not have changed the verdict; we record which we applied because the rule does not spell the difference out.

\begin{center}\small\resizebox{\ifdim\width>\linewidth\linewidth\else\width\fi}{!}{%
\begin{tabular}{llllll}
\toprule
h & $\sigma$ larger by & error smaller by & total & share from $\sigma$ & share from accuracy \\
\midrule
1 & 1.51$\times$ & 1.00$\times$ & 1.51$\times$ & 101\% & -1\% \\
8 & 1.75$\times$ & 1.05$\times$ & 1.83$\times$ & 93\% & 7\% \\
32 & 1.63$\times$ & 1.11$\times$ & 1.81$\times$ & 82\% & 18\% \\
\textbf{100} & \textbf{1.55$\times$} & 1.25$\times$ & 1.94$\times$ & \textbf{66\%} & 34\% \\
128 & 1.51$\times$ & 1.32$\times$ & 2.00$\times$ & 60\% & 40\% \\
368 & 1.35$\times$ & 1.77$\times$ & 2.39$\times$ & 34\% & 66\% \\
\bottomrule\end{tabular}}\end{center}

\emph{On the held-out pair, n\_independent = 4 400-step trajectories, every model at 2,500 iterations: the same $\sigma$-versus-accuracy split this appendix uses for rule M-44, so the two arms' improvements can be compared part by part. Shares are of the log improvement, so the two columns add to one.}

\textbf{The objective's own contribution is below what this design can resolve.} The rule scores the combined arm against the \textbf{shared-trunk} arms, so its verdict measures the two fixes together. The comparison that isolates the objective is against \S{}5.3's independent arm, which differs from this one in the loss type and nothing else, from the same script on the same trajectories. At h = 100 switching \texttt{mse} for \texttt{gaussian\_nll} multiplies the overconfidence factor by \textbf{1.044$\times$} (95\% interval from the same cluster bootstrap over whole trajectories, [1.029, 1.054]), the wrong direction, slightly; moves $\pm$1$\sigma$ coverage by +0.22 points ([-0.31, +1.09], which spans zero); and multiplies the epistemic $\sigma$ by 0.940$\times$. Both effects sit under the minimum detectable effect fixed in advance, 1.218$\times$ on the factor and 2.50 points on coverage (results/p3\_combined\_arm\_power.json, binding\_mde at h = 100 (largest of the four calibrations)), so although the factor's interval lies wholly on the worse side of no change, the objective's separate effect is \textbf{below the size this design was built to resolve at n\_independent = 4}, and the point estimate points away from an improvement. It does not establish that the objective contributes nothing: reading a null out of an effect smaller than its own floor is the error M-24 and M-43 record, and the rule's fourth branch exists to say underpowered rather than null. What the arm does settle is the combination, which is better calibrated than the shared-trunk arms by essentially the amount \S{}5.3 measured for independence alone (2.03$\times$ there, 1.94$\times$ here). \textbf{The two fixes do not add. A reader who added them would have been wrong, which is why the arm was run.} That the objective's separate effect is at most small is not a surprise once stated: the objective governs the \textbf{aleatoric} head, and the epistemic term is a spread across members that the loss never sees.

\textbf{And it still does not repair the interval.} The combined arm is 5.4$\times$ overconfident at h = 100 with 15.53\% coverage where a calibrated Gaussian gives 68.27\%. Better than the shared-trunk arms, no better than independence alone, and still not an interval. \S{}5.7's per-horizon multiplier remains the only correction in this paper whose estimates all land within the band, though only on the released checkpoint and no single cell is resolvable at this arena. On Arm A, whose model never saw the test episodes, it manages 17 of 36 epistemic cells (\S{}5.7).

\textbf{What the arm does not separate.} It differs from the shared-trunk arms on three axes at once: trunk sharing, the objective, and capacity (a factor of 3.49 in state-pathway parameters, as \S{}5.3's contrast also does; supplementary S2), and independently seeded runs also differ in data ordering. The rule said so in advance: it bounds the combination and attributes nothing. The one further comparison here, the \texttt{mse}/\texttt{gaussian\_nll} pair above, which holds every other axis fixed, sits outside the rule's governing statistics and under the design's own minimum detectable effect, so it bounds the objective's separate contribution rather than measuring it.

\section{Ranking against free signals}

Detail behind \S{}5.6: the correlation at each horizon, the robustness checks, the free baselines and their cost, and the rule on shorter units.

\textbf{Why \S{}5.6's paired test, not overlapping intervals.} Comparing the two marginal intervals for overlap is the wrong comparison: both correlations are measured on the \emph{same} trajectories, so their sampling errors move together and the marginal intervals are needlessly conservative. The paired difference, resampling whole trajectories and recomputing \emph{both} correlations inside each draw, is the appropriate and more powerful test.

\textbf{The shorter unit at h = 128.} At the shorter unit of rule M-64 (Appendix E), 60 non-overlapping units over the same ten episodes give [+0.148, +0.351] for the same paired difference, so that horizon carries more weight than it did.

\textbf{r\_dd per horizon.} \textbf{Per horizon, on the same 20 trajectories (the released checkpoint over all ten episodes) and the same kind of cluster bootstrap:} it is a separate run, so its h = 368 interval differs from M-45's in the table below, [+0.318, +0.576], in the third decimal, as this appendix's pooled interval does from Appendix D's [+0.545, +0.694].

\begin{center}\small\resizebox{\ifdim\width>\linewidth\linewidth\else\width\fi}{!}{%
\begin{tabular}{llll}
\toprule
h & r\_dd, double-demeaned & 95\% CI & excludes zero \\
\midrule
1 & --- & --- & --- \\
8 & +0.306 & [-0.387, +0.713] & no \\
32 & +0.186 & [-0.032, +0.636] & no \\
\textbf{100} & \textbf{+0.385} & \textbf{[+0.283, +0.701]} & \textbf{yes} \\
128 & +0.453 & [+0.359, +0.713] & yes \\
368 & +0.419 & [+0.315, +0.573] & yes \\
\bottomrule\end{tabular}}\end{center}

\emph{(h = 1 has one forecast step per trajectory, so there is nothing within a rollout to demean against and r\_dd is undefined rather than zero.)}

The within-rollout effect is materially smaller than the pooled figure and not established at short horizon; at h = 1 the correlation ranks whole rollouts, on 20 trajectory-level points (below).

\textbf{Rule M-43 (\S{}5.6).} \textbf{Does it hold on a model we trained?} Three Arm A arms at ensemble size 5 (2,500 training iterations, scored on the held-out pair), under a rule committed before the runs (rule M-43, Appendix E), lead the index in \textbf{12 of 12} seed-horizon cells, but the paired difference excludes zero at only 1 of 4 horizons, so the rule returns \textbf{DOES NOT GENERALISE}. At n\_independent = 4 400-step trajectories it was under-powered, which we measured after the fact and should have checked before committing it (below).

\textbf{The expectation.} \textbf{We ran the baseline test expecting it to go the other way.} A counter matching disagreement would have been the more consequential result, making the trust metric close to vacuous, since a counter is free. We record that as an expectation only: it was not committed to git before the data existed, so by this paper's own standard (\S{}3.4) it is not a pre-registration. It did not go that way against the counter. It went that way against something else.

\textbf{The two free baselines and their cost (\S{}5.6).} A claim that beats exactly one competitor is a claim about that competitor. Under a rule committed before either was computed (rule M-51, Appendix E, corrected by rule M-52), we added two baselines needing no ensemble and no second model: \texttt{step-size}, the magnitude of the model's own predicted state change $\lVert \mu_t - \mu_{t-1} \rVert$, which costs nothing because the rollout has already made those predictions; and \texttt{entry-res}, its one-step error at the step \emph{before} the forecast window opens, which costs \textbf{one extra rollout in this harness} and nothing in deployment, where a model consumes the history to build its recurrent state anyway. The first rule called both free without distinguishing those, and the second records the correction.

\textbf{X6's alongside readings (\S{}5.6).} On the released checkpoint's 70 units of 132 rows over its ten released episodes, at h = 100 the margin also exceeds the 0.1713 minimum detectable effect computed before the reading by rule M-51's method. Disagreement correlates +0.6808 with error there against step size's +0.3992, and keeps +0.6361 [+0.5567, +0.7466] with step size partialled out; alongside, on 150 units at h = 32 the margin is +0.2393 [+0.0950, +0.3131].

\textbf{What survives removing each confound.} A linear partial is not much of a control: error does not grow linearly with depth, and a control that under-fits the index leaves index-driven variance in the residual and flatters disagreement. The first five rows after the pooled baseline below are \textbf{depth controls}, each partialling out a different model of how far into the rollout you are; the last four \textbf{decompose} the pooled figure into its between-trajectory and within-trajectory parts, and the last of those was pre-registered before it was computed (rule M-45, Appendix E). \textbf{One row is not comparable to the others}: the rank partial is a correlation of ranks rather than of values, so its being larger than the pooled figure says nothing about how much depth explains. Every row is the released checkpoint over all ten episodes, n\_independent = 20 400-step trajectories.

\begin{center}\small\resizebox{\ifdim\width>\linewidth\linewidth\else\width\fi}{!}{%
\begin{tabular}{llll}
\toprule
what is removed & correlation & 95\% CI & what survives it \\
\midrule
nothing (pooled) & +0.605 & [+0.545, +0.695] & --- \\
the step index, linearly & +0.596 & --- & depth explains 0.010 of it, from +0.605 \\
log(1 + index) & +0.589 & [+0.512, +0.688] & a non-linear model of depth \\
a cubic in the index & +0.582 & [+0.495, +0.676] & any polynomial trend in depth \\
any monotone function of depth (rank partial) & +0.906 & [+0.856, +0.921] & depth in any monotone form \\
depth exactly, within each forecast step & \textbf{+0.739} & [+0.711, +0.852] & positive at 368 of 368 steps, median +0.737\textsuperscript{1} \\
forecast-step means only & +0.589 & --- & the within-step control, pooled \\
everything within a trajectory & +0.470 & [+0.389, +0.604] & --- \\
the 20 trajectory means alone & +0.878 & [+0.817, +0.956] & do harder rollouts disagree more? \\
\textbf{both, additively (r\_dd)} & \textbf{+0.419} & \textbf{[+0.318, +0.576]} & \textbf{at a given depth, in a given rollout, does disagreement know?} \\
\bottomrule\end{tabular}}\end{center}

\textsuperscript{1} Adjacent forecast steps on the same 20 trajectories are heavily dependent --- structurally the same problem that the trajectory permutation null corrects for across the 45 coupled state dimensions (Appendices C and I). The count is descriptive; the interval [+0.711, +0.852] is the statistic, and no P-value attaches to 368/368.

The weakest figure across the 5 depth controls is +0.582, so disagreement is not re-encoding the clock: at a fixed depth it still knows which rollouts are going wrong. But depth was never the only confound. Per-episode difficulty, the released checkpoint's error on each episode scored with the causal action pairing (\S{}6), spans 0.523 to 1.585 and is barely correlated with commanded speed (+0.088 across the ten episodes), so if harder trajectories simply have both larger error and larger disagreement, the pooled correlation would look exactly as it does with disagreement carrying no within-rollout information at all. Two things pointed there: the 20-point h = 1 figure of +0.994, which is not the shape of a genuine per-step signal, and the within-step control coming out \emph{above} the pooled figure, the signature of a between-unit effect.

\textbf{The verdict.} The between-trajectory correlation is +0.878 and the two components contribute 51.7\% and 48.3\% of the pooled covariance, so a large part of the pooled correlation is a between-rollout effect. That reinterprets the within-step control: it is a mean of 368 between-trajectory correlations, which is why it reads +0.739 against the pooled +0.605 rather than below it. The decisive statistic is the double-demeaned one, and it survives: +0.419 [+0.318, +0.576] at n\_independent = 20, against the rule's pre-committed threshold that the interval exclude zero and a minimum detectable effect of 0.183 from the bootstrap's standard error; a dilution study missed an effect of 0.086 and caught one of 0.172. \textbf{The rule returns DISAGREEMENT CARRIES WITHIN-ROLLOUT INFORMATION}: with both the rollout and the depth held constant, disagreement still tracks error rather than merely reporting which episode is hard.

Two qualifications go with that. The within-rollout effect is \textbf{materially smaller than the pooled figure}, +0.419 against +0.605, so a practitioner should expect disagreement to separate \emph{rollouts} better than it separates \emph{moments within a rollout}. And it is \textbf{not established at short horizon}: r\_dd's interval excludes zero at h=100, h=128, h=368 and spans zero at h=8 and h=32, where too few steps exist to demean against. At h = 100 it is established, at +0.385 [+0.283, +0.701].

\textbf{The h = 1 figure survives the same test, but it is not what it looked like.} At h = 1 the panel has one column, so +0.994 is a correlation over 20 \emph{trajectory-level} points and nothing within a rollout is being tested. Disagreement correlates +0.006 with commanded speed and +0.027 with per-episode difficulty, and partialling both out of the disagreement--error correlation leaves +0.995: it does not move, so the figure is real and not a difficulty artifact. It is nevertheless a statement about \textbf{ranking whole rollouts at one step ahead}, on 20 points, and \S{}5.6 states it that way.

\textbf{Does it hold on a model we trained?} Apart from this rule and its replication on the fold ensembles, \S{}5.6 is measured on the released checkpoint, because our main arms run at ensemble size 1 where the epistemic term is identically zero. We therefore trained three Arm A arms at \textbf{ensemble size 5}, at 2,500 training iterations and identical in every other setting, under a rule committed to git before the runs existed (rule M-43, Appendix E). It asked that disagreement lead the index at every horizon, and that the paired difference exclude zero at a majority of them.

\textbf{It returns DOES NOT GENERALISE.} The first condition passes completely: disagreement leads the index in \textbf{12 of 12} seed-horizon cells, every paired estimate positive, +0.204 to +0.545. The second fails: the paired difference excludes zero at 1 of 4 horizons, not a majority. We report the verdict the rule returns and do not rewrite the rule, nor its denominator: it was committed over 4 horizons, and adding h = 100 after the fact would change what "a majority" means in a discharged rule. The released checkpoint's comparison with the index (\S{}5.6) follows the six-horizon grid, because no pre-registration is stated over it, so the two counts are deliberately different.

\textbf{What separates the two conditions is sample size, and we measured that rather than asserting it.} Our own arms can only be scored out-of-sample on the held-out pair, n\_independent = 4, where the released checkpoint's finding used 20. Subsampling four trajectories at a time from a twenty-trajectory pool, the rule's criterion fires on 75\% of draws on average and on only 24\% at h=8. That estimate is an \textbf{upper bound}, because the pool it subsamples is in-sample for these arms, where the effect is +0.425 to +0.790 against +0.204 to +0.545 on the held-out pair. So the rule was under-powered at the sample size it faced, decisively at one horizon. We do not claim it could not have passed, only that it was committed without checking what it could detect: a failure of ours, the same one the ledger records as M-24, a rule anchored without regard to the regime it would be applied in.

\emph{Reported as a companion and not as a discharge:} on all ten episodes (n\_independent = 20 400-step trajectories, \textbf{in-sample} for these arms, which trained on eight of them) the same measurement excludes zero at 4 of 4 horizons and would have satisfied both conditions. It cannot discharge the rule, which is stated over the out-of-sample arena; we record it only so the comparison with the released checkpoint's 20 is like for like.

\emph{Why the within-step control is undefined for the forecast index.} The within-step control holds the forecast step fixed and correlates across trajectories, so it annihilates any quantity that is constant across trajectories at a fixed step --- which is the forecast index, whose within-step correlation is therefore undefined rather than zero. It does \textbf{not} annihilate a per-trajectory scalar like \texttt{entry-res}, which varies across exactly the axis the control varies over. Rule M-51 said the opposite, and \texttt{M-54} records the correction.

Rule X6 was committed and pushed before its readings existed (ledger M-88; Appendix E gives its lead time) and is discharged in M-91. It is exploratory and re-opens no rule: rules M-51 and M-52 stand. It puts \S{}5.6's comparison to more units. The released checkpoint's ten episodes are cut into non-overlapping units of the model's history plus h forecast steps, inside each episode, by rule M-64's index builder, at h = 32 and h = 100, which governs. The statistic is rules M-51 and M-52's margin, r(disagreement, $|$error$|$) $-$ r(step size, $|$error$|$) on the applied scalar, a Pearson correlation over every (unit, step) cell, imported from their script; its 95\% interval resamples whole units, paired, 20,000 times (seed 0). The rule reads \textbf{DISAGREEMENT BEATS STEP SIZE} if the interval's lower bound is above zero, STEP SIZE BEATS DISAGREEMENT if its upper bound is below zero, and NOT RESOLVED otherwise. The minimum detectable effect was computed by rule M-51's method at this n before the reading, and governs nothing.

\begin{center}\small\resizebox{\ifdim\width>\linewidth\linewidth\else\width\fi}{!}{%
\begin{tabular}{lllllll}
\toprule
h & units (rows each) & r(disagreement, error) & r(step size, error) & margin [95\% interval] & minimum detectable margin (rule M-51's method) & partial r(disagreement given step size) [95\% interval] \\
\midrule
100 & 70 (132) & +0.6808 & +0.3992 & \textbf{+0.2816 [+0.1978, +0.3896]} & 0.1713 & +0.6361 [+0.5567, +0.7466] \\
32 & 150 (64) & +0.7415 & +0.5022 & +0.2393 [+0.0950, +0.3131] & 0.2214 & +0.6317 [+0.4691, +0.7250] \\
\bottomrule\end{tabular}}\end{center}

\textbf{Arena: all ten episodes, the released checkpoint's training data, as non-overlapping units inside each episode; n\_independent = the unit count at each horizon. Checkpoint: the released one.} Bold, the reading, which returns \textbf{DISAGREEMENT BEATS STEP SIZE}; the h = 32 row is alongside, and its margin would carry DISAGREEMENT BEATS STEP SIZE. Units from one episode are not independent of each other, and the interval resamples units, not episodes; the statistic is a correlation over (unit, step) cells, not a ranking of units.

\section{Per-horizon recalibration}

Detail behind \S{}5.7: the constant scalar, the per-horizon table, whether multipliers transfer between models, and the nine-episode fit.

\textbf{The constant scalar (\S{}5.7).} On the released checkpoint, each scalar fitted on one held-out episode's 2 trajectories and scored on the other's: fitting at one step works at one step and nowhere else. On the epistemic term --- the quantity the method uses --- a scalar of 5.08--5.82 brings h=1 coverage to 63--74\%, essentially calibrated against the 68.27\% target, and leaves h=368 at 17--21\%. On the aleatoric term a scalar of 593--611 gives 64--70\% at h=1 and 11--15\% at h=368. Fitting over the whole rollout instead drives one-step coverage to 100\%, an interval wide enough to be vacuous where the model is accurate, while at the far end its coverage depends on which episode fitted the multiplier. A constant multiplier cannot track an error that grows while $\sigma$ does not (\S{}5.7), so for a \emph{constant} scale "right shape, wrong scale" does not survive.

\textbf{The per-horizon table and its reading (\S{}5.7).} One multiplier per horizon is fitted on one held-out episode and evaluated on the other, in both directions, so no multiplier is ever scored on the episode that produced it. The two held-out episodes contribute 4 non-overlapping 400-step trajectories between them, so each direction fits on n\_independent = 2 and is scored on the other 2. The last column repeats the test on Arm A at ensemble size 5 and 2,500 iterations (3 seeds), whose model never saw either episode:

\begingroup\small
\begin{longtable}{>{\raggedright\arraybackslash}p{\dimexpr 0.1213\linewidth-2\tabcolsep\relax}>{\raggedright\arraybackslash}p{\dimexpr 0.2783\linewidth-2\tabcolsep\relax}>{\raggedright\arraybackslash}p{\dimexpr 0.1107\linewidth-2\tabcolsep\relax}>{\raggedright\arraybackslash}p{\dimexpr 0.1882\linewidth-2\tabcolsep\relax}>{\raggedright\arraybackslash}p{\dimexpr 0.1001\linewidth-2\tabcolsep\relax}>{\raggedright\arraybackslash}p{\dimexpr 0.2014\linewidth-2\tabcolsep\relax}}
\toprule quantity & released checkpoint: cells unseen by the multiplier within 10 points of 68.27\%, per-horizon c (unpowered) & same, constant c & per-horizon c fitted on the \emph{other} model, both models scored (unpowered) & range of fitted c & Arm A, its own per-horizon c: cells within the band, unseen by the model too \\ \midrule\endfirsthead
\toprule quantity & released checkpoint: cells unseen by the multiplier within 10 points of 68.27\%, per-horizon c (unpowered) & same, constant c & per-horizon c fitted on the \emph{other} model, both models scored (unpowered) & range of fitted c & Arm A, its own per-horizon c: cells within the band, unseen by the model too \\ \midrule\endhead
\bottomrule\endlastfoot
aleatoric & \textbf{12 /\allowbreak{} 12} & 2 /\allowbreak{} 12 & 0 /\allowbreak{} 72 & 592.6 -- 7782 (13.1$\times$) & 10 /\allowbreak{} 36 \\
epistemic & \textbf{12 /\allowbreak{} 12} & 2 /\allowbreak{} 12 & 12 /\allowbreak{} 72 & 5.082 -- 47.33 (9.31$\times$) & 17 /\allowbreak{} 36 \\
\end{longtable}\endgroup

On the released checkpoint every point estimate lands within 10 points of the 68.27\% target for both quantities. \textbf{That absolute test is UNPOWERED here, as in the \emph{different model} column:} \texttt{results/\allowbreak{}p4\_\allowbreak{}transfer\_\allowbreak{}power\allowbreak{}.json} scored this model under these very multipliers and found the 10-point band resolvable at 0 of 12 quantity-by-horizon pairs, with a binding minimum detectable effect of 12.55--40.62 points. So a cell inside the band is compatible with a true coverage well outside it, and nothing here shows any cell is outside it either. The largest deviation over all 24 held-out cells is aleatoric at h=100, fitted on episode 8 and scored on the other, at 77.48\%, 9.21 points off target. The two largest deviations are both on the aleatoric term and both above target --- 77.48\% at h=100 and 76.55\% at h=128 --- so the fitted multiplier is mildly \textbf{conservative} at the long horizons rather than unstable in both directions. The constant scalar manages 2 of 12, and those are the h=1 cells it was fitted at. \textbf{On Arm A, where the episodes are unseen by the model as well, its own multipliers manage 17 of 36 epistemic and 10 of 36 aleatoric cells}, so the recipe that lands every released-checkpoint cell does not carry over intact to a model that has not seen the test episodes.

\textbf{Rule M-69.} \textbf{The fourth column: the table is a property of the model, not of the horizon.} Under rule M-69 (Appendix E), multipliers fitted on Arm A and scored on the released checkpoint, and the reverse, on the same 4 held-out trajectories, return \textbf{DOES NOT TRANSFER --- A PROPERTY OF THE MODEL} on the paired change in coverage the arena can resolve: 52 of 72 governing cells, which are not independent tests, have an interval wholly outside the $\pm$10-point band, and Arm A's multipliers are 0.0069$\times$ to 1.36$\times$ the released checkpoint's at the same horizon (below). The \emph{different model} column is the absolute test, unpowered for the reason above, so it is reported and cannot move that verdict.

\textbf{The fourth column: the table is a property of the model, not of the horizon.} Everything else in \S{}5.7 is established across \emph{episodes}, on one model. Whether the same lookup table works on a \emph{different} model is a separate claim, and rule M-69 (Appendix E) fixed what an answer would look like --- criterion, arena, horizons and minimum detectable effect --- before any cross-model multiplier was computed. Multipliers were fitted on Arm A at ensemble size 5 (3 seeds) and scored on the released checkpoint, then the reverse, over the same 4 held-out trajectories and 6 horizons. The statistic the rule governs on is the \textbf{paired} change in held-out coverage, coverage under the other model's multiplier minus coverage under the same model's, on the same trajectories, because this arena can resolve that (largest minimum detectable effect 5.73 points against the $\pm$10-point band) and cannot resolve the absolute one. The verdict is \textbf{DOES NOT TRANSFER --- A PROPERTY OF THE MODEL}: of 72 governing cells, 52 have a 95\% interval on the paired change lying entirely outside the band, 3 entirely inside and 17 straddling an edge, and the largest paired change has magnitude 67.5 points. That is branch 1 of the rule's three. Per direction the verdict is the same: fitting on Arm A, \textbf{DOES NOT TRANSFER --- A PROPERTY OF THE MODEL} (29 of 36 cells outside the band); fitting on the released checkpoint, \textbf{DOES NOT TRANSFER --- A PROPERTY OF THE MODEL} (23 of 36). The plainest statement of the gap is the multipliers themselves: Arm A's are 0.0069$\times$ to 1.36$\times$ the released checkpoint's at the same horizon.

\textbf{The \emph{different model} column is the absolute test, and it is unpowered for the reason given above}, so a cell inside its band is not evidence the multiplier transferred. It is reported because dropping it would hide that this arena cannot resolve it; it is not the verdict and cannot move it. The caution above applies here unchanged and harder: the 72 governing cells are not 72 independent tests, the 3 Arm A seeds share their training data and differ only in initialisation and ordering, and no P-value attaches to any count here. The two models also differ on several axes at once, so this bounds transfer between these two models rather than attributing it to any one difference.

\textbf{X7's alongside readings.} Alongside, the aleatoric term returns WITHIN BAND AT 3 OF 6 HORIZONS, one constant multiplier lands within the band at one of the six horizons, and the fitted epistemic multipliers differ by at most 1.18$\times$ across the ten folds.

Rule X7 was committed and pushed before its readings existed (ledger M-89; Appendix E gives its lead time) and is discharged in M-92. It is exploratory and re-opens no rule: rules M-69 and M-70 stand. For each of the ten episodes, the per-horizon multiplier is fitted on the other nine episodes' non-overlapping 400-step trajectories with \S{}5.7's fitting function, imported, and scores $\pm$1$\sigma$ coverage on that episode's own; pooled, each of the 20 trajectories is scored once, by a multiplier not fitted on it. A horizon reads WITHIN BAND if the coverage's 95\% interval lies within 10 points of 68.27\%, OUTSIDE BAND if it lies wholly outside that band, and UNRESOLVED otherwise. The epistemic term governs, and the reading counts its horizons WITHIN BAND. The interval resamples the 20 trajectories 20,000 times (seed 0), one draw shared across horizons.

\begingroup\small
\begin{longtable}{>{\raggedright\arraybackslash}p{\dimexpr 0.0575\linewidth-2\tabcolsep\relax}>{\raggedright\arraybackslash}p{\dimexpr 0.2165\linewidth-2\tabcolsep\relax}>{\raggedright\arraybackslash}p{\dimexpr 0.1520\linewidth-2\tabcolsep\relax}>{\raggedright\arraybackslash}p{\dimexpr 0.1566\linewidth-2\tabcolsep\relax}>{\raggedright\arraybackslash}p{\dimexpr 0.1320\linewidth-2\tabcolsep\relax}>{\raggedright\arraybackslash}p{\dimexpr 0.2855\linewidth-2\tabcolsep\relax}}
\toprule h & epistemic, per-horizon multiplier (the reading) & aleatoric, per-horizon multiplier & epistemic, one constant multiplier & epistemic, no multiplier & epistemic multiplier across the ten folds (smallest -- largest) \\ \midrule\endfirsthead
\toprule h & epistemic, per-horizon multiplier (the reading) & aleatoric, per-horizon multiplier & epistemic, one constant multiplier & epistemic, no multiplier & epistemic multiplier across the ten folds (smallest -- largest) \\ \midrule\endhead
\bottomrule\endlastfoot
1 & \textbf{68.6 [63.9, 72.9]}, WITHIN BAND & 68.2 [57.4, 77.6], UNRESOLVED & 68.6 [63.9, 72.9], WITHIN BAND & 16.2 [13.4, 18.9] & 6.29 -- 6.81 \\
8 & \textbf{68.4 [61.2, 74.9]}, WITHIN BAND & 68.5 [57.6, 78.3], UNRESOLVED & 50.6 [44.7, 55.8], OUTSIDE BAND & 10.0 [8.6, 11.3] & 10.44 -- 12.35 \\
32 & \textbf{68.3 [61.3, 74.6]}, WITHIN BAND & 68.4 [57.2, 78.3], UNRESOLVED & 37.1 [32.4, 41.4], OUTSIDE BAND & 7.0 [6.0, 7.8] & 16.44 -- 18.84 \\
100 & \textbf{68.2 [64.5, 71.7]}, WITHIN BAND & 67.9 [59.4, 75.7], WITHIN BAND & 26.5 [24.0, 28.9], OUTSIDE BAND & 4.6 [4.1, 5.1] & 28.87 -- 31.90 \\
128 & \textbf{68.2 [65.3, 71.0]}, WITHIN BAND & 68.0 [60.2, 75.2], WITHIN BAND & 25.0 [23.1, 26.9], OUTSIDE BAND & 4.4 [4.0, 4.8] & 31.86 -- 34.20 \\
368 & \textbf{68.2 [65.3, 71.1]}, WITHIN BAND & 68.1 [60.9, 75.3], WITHIN BAND & 21.2 [19.7, 22.7], OUTSIDE BAND & 3.6 [3.3, 3.9] & 40.31 -- 43.02 \\
\end{longtable}\endgroup

\textbf{Arena: all ten episodes' 20 non-overlapping 400-step trajectories, n\_independent = 20, each scored by a multiplier fitted on the other nine episodes. Checkpoint: the released one.} Each cell is $\pm$1$\sigma$ coverage in percent with its 95\% interval and, for the multiplied columns, its label; bold, the reading, \textbf{WITHIN BAND AT 6 OF 6 HORIZONS}. The other columns are alongside: the aleatoric term returns WITHIN BAND AT 3 OF 6 HORIZONS, and one constant multiplier, fitted the same way, lands within the band at one of the six horizons. Every trajectory is unseen by its multiplier but perhaps not by the model, whose training rows are not recorded (\S{}3.2), so this says nothing about a model that has not seen the test episodes (the per-horizon table's Arm A column, above).

\textbf{$\pm$2$\sigma$ under the same fits (post hoc).} Scored post hoc on the same fits, arena and draws (X7's $\pm$1$\sigma$ cells are reproduced exactly first), $\pm$2$\sigma$ coverage under the per-horizon multiplier stays below the 95.45\% a calibrated Gaussian gives at every horizon, so \S{}5.7's "not two-sigma" holds at each one, not only at h = 100.

\begin{center}\small\resizebox{\ifdim\width>\linewidth\linewidth\else\width\fi}{!}{%
\begin{tabular}{llll}
\toprule
h & epistemic $\pm$2$\sigma$, per-horizon multiplier & epistemic $\pm$2$\sigma$, one constant multiplier & epistemic $\pm$2$\sigma$, no multiplier \\
\midrule
1 & \textbf{86.3 [82.1, 90.0]} & 86.3 [82.1, 90.0] & 30.1 [25.8, 34.2] \\
8 & \textbf{85.1 [79.1, 90.2]} & 71.1 [64.2, 77.3] & 19.8 [17.1, 22.2] \\
32 & \textbf{84.5 [78.2, 89.8]} & 57.9 [51.4, 63.7] & 13.7 [11.8, 15.4] \\
100 & \textbf{83.5 [80.9, 85.9]} & 43.8 [40.3, 47.2] & 9.3 [8.3, 10.2] \\
128 & \textbf{83.5 [81.5, 85.4]} & 41.8 [39.1, 44.5] & 8.8 [8.0, 9.5] \\
368 & \textbf{83.6 [81.6, 85.5]} & 36.2 [33.9, 38.6] & 7.2 [6.6, 7.8] \\
\bottomrule\end{tabular}}\end{center}

\textbf{Arena and checkpoint as in the table above.} Each cell is $\pm$2$\sigma$ coverage in percent with its 95\% interval.

\section{Defects in the released pipeline}

Detail behind \S{}6: the stale action across models and horizons, the spliced windows' cost, and the pipeline rows the ledger confirms.

\textbf{The held-out pair's per-trajectory values (\S{}6).} At h = 368 on the held-out pair the stale action changes the released checkpoint's error, trajectory by trajectory, by +0.8\%, +9.8\%, +14.8\%, +6.5\% on relative-L1 and +2.4\%, +5.8\%, +8.3\%, -0.3\% on nRMSE (\texttt{results/\allowbreak{}alignment\_\allowbreak{}defect\_\allowbreak{}ci\allowbreak{}.json}).

\textbf{Our models under the stale pairing (\S{}6).} For models trained under the causal pairing, the stale one also raises error, and how much depends on the arena. On their own training episodes' 16 trajectories it raises our 10,000-iteration Arm A's relative-L1 error by +124.6\% [+75.8, +201.9] at h = 1 and +209.3\% [+167.3, +257.6] at h = 100; on the held-out pair's 4 it changes it by -1.0\% [-6.1, +27.7] at h = 1, which those trajectories cannot resolve, and by +16.6\% [+1.9, +72.6] at h = 100 (three-seed means, computed post hoc, \texttt{results/\allowbreak{}n5\_\allowbreak{}stale\_\allowbreak{}armA\_\allowbreak{}ci\allowbreak{}.json}; the tables below give every model at every horizon). Whether the action matters to them at all is rule X2's question, which returns \textbf{RESPONDS TO THE ACTION} below, with its caveat that every action in the data comes from one policy.

\textbf{The forecast shift (\S{}6).} A percentage change depends on how accurate the model already is as much as on how much the action matters: measured as the shift in the forecast itself, the stale action moves our model's forecast at h = 8 by 0.200 and the released checkpoint's by 0.204 on their own training data (below), so at h = 8 the two respond about equally, and our model's larger percentages there reflect its smaller error, 0.086 against 0.158 relative-L1 (\texttt{results/\allowbreak{}n5\_\allowbreak{}stale\_\allowbreak{}armA\_\allowbreak{}ci\allowbreak{}.json}); at h = 100 the released checkpoint's forecast moves more, 0.373 against 0.246.

Rule X2 was committed and pushed before any of its readings existed (ledger M-84; Appendix E gives its lead time) and is discharged in M-85. It is exploratory and re-opens no rule. It asks whether our trained models condition their forecasts on the action they are given. Only the actions the forecast steps read are changed, so the history and the recurrent state that meets the first forecast action are untouched. E is the change in relative-L1 error, cumulative over forecast steps 1..h, against the true actions: the ratio err(I) / err(true), less one, in percent; for our arms, the mean over three seeds of each seed's E. Its 95\% interval resamples whole trajectories, one resample for all seeds and both interventions in each draw: exact over all 256 ordered resamples on the held-out pair, 20,000 Monte Carlo resamples (seed 0) otherwise. The interventions are the stale action (the released evaluation's offset), another trajectory's actions (a swap, each trajectory taking the next one's), each action dimension's mean over the model's training rows, and the true actions plus Gaussian noise at k times each dimension's training standard deviation (eight draws). \textbf{The readings} are Arm A's, under the swap at h = 8 on its in-sample arena: \textbf{RESPONDS TO THE ACTION} at 2,500 iterations (+227.8\% [+131.6, +398.2]) and at 10,000 (+676.3\% [+454.8, +1060.0]). The criterion was fixed in the rule before any reading existed (M-84), and the first outcome that matches applies: RESPONDS TO THE ACTION if the point estimate is at least +10\% and the interval's lower bound is above zero; ERROR FALLS WITH WRONG ACTIONS if the interval's upper bound is below zero; DOES NOT RESPOND MEASURABLY if its upper bound is below +10\%; otherwise UNRESOLVED. The bar is deliberately low: the one-step-stale action alone raises the released checkpoint's error at h = 8 by 54.4\% over all ten episodes. What the rule establishes is that our forecasts depend on the action. It does not establish that they depend on it correctly: every action in the data comes from one policy, so a forecast under an action that policy would not take has no recorded outcome to check it against. Arm B and the released checkpoint get the same statistic, alongside and not as readings.

\begin{center}\small\resizebox{\ifdim\width>\linewidth\linewidth\else\width\fi}{!}{%
\begin{tabular}{llllll}
\toprule
model, checkpoint & actions given & h = 1 & h = 8 & h = 32 & h = 100 \\
\midrule
Arm A, 2,500 & stale action & +27.8 [+15.6, +49.4] & +31.5 [+19.2, +53.0] & +53.8 [+34.6, +83.5] & +48.7 [+34.8, +65.7] \\
Arm A, 2,500 & another trajectory's actions & +73.2 [+33.7, +162.7] & \textbf{+227.8 [+131.6, +398.2]} & +376.0 [+248.7, +578.8] & +349.3 [+264.7, +454.0] \\
Arm A, 2,500 & training mean & +49.6 [+25.5, +101.3] & +169.5 [+104.6, +278.0] & +309.6 [+215.5, +450.5] & +300.5 [+234.8, +380.6] \\
Arm A, 2,500 & noise, k = 0.1 & +0.5 [+0.2, +1.1] & +0.6 [+0.4, +1.1] & +1.3 [+0.7, +2.2] & +1.3 [+0.8, +1.8] \\
Arm A, 2,500 & noise, k = 0.5 & +8.2 [+3.9, +17.9] & +12.3 [+7.1, +21.6] & +24.7 [+14.4, +41.4] & +23.6 [+16.6, +32.3] \\
Arm A, 10,000 & stale action & +124.6 [+75.8, +201.9] & +161.8 [+121.3, +231.1] & +205.9 [+156.7, +275.9] & +209.3 [+167.3, +257.6] \\
Arm A, 10,000 & another trajectory's actions & +278.9 [+145.8, +548.3] & \textbf{+676.3 [+454.8, +1060.0]} & +986.6 [+723.5, +1381.1] & +1081.0 [+896.2, +1301.0] \\
Arm A, 10,000 & training mean & +207.1 [+114.5, +385.7] & +531.7 [+380.0, +781.9] & +841.6 [+657.6, +1100.7] & +899.9 [+761.2, +1058.4] \\
Arm A, 10,000 & noise, k = 0.1 & +3.9 [+1.7, +8.6] & +6.0 [+3.3, +10.5] & +10.4 [+6.6, +16.1] & +11.4 [+8.6, +14.8] \\
Arm A, 10,000 & noise, k = 0.5 & +53.3 [+27.2, +106.9] & +74.8 [+47.5, +122.0] & +126.3 [+87.8, +181.8] & +141.2 [+115.1, +172.0] \\
Arm B, 2,500 (alongside) & stale action & +18.7 [+8.9, +36.1] & +15.9 [+7.9, +29.5] & +20.3 [+10.3, +36.8] & +9.3 [+5.1, +14.9] \\
Arm B, 2,500 (alongside) & another trajectory's actions & +63.6 [+31.1, +131.4] & +128.8 [+67.5, +241.8] & +173.4 [+101.6, +289.0] & +111.0 [+78.3, +155.4] \\
Arm B, 2,500 (alongside) & training mean & +41.4 [+22.7, +77.7] & +94.7 [+54.1, +164.4] & +145.9 [+89.1, +233.8] & +149.3 [+113.0, +195.4] \\
Arm B, 2,500 (alongside) & noise, k = 0.1 & +0.4 [+0.1, +0.8] & +0.3 [+0.1, +0.6] & +0.5 [+0.3, +1.0] & +0.4 [-0.2, +1.2] \\
Arm B, 2,500 (alongside) & noise, k = 0.5 & +7.2 [+3.7, +14.5] & +5.9 [+3.2, +11.4] & +9.8 [+5.2, +17.7] & +7.6 [+4.5, +11.5] \\
Arm B, 10,000 (alongside) & stale action & +75.1 [+44.2, +118.9] & +68.6 [+45.9, +105.2] & +60.0 [+39.9, +98.7] & +32.5 [+20.1, +50.3] \\
Arm B, 10,000 (alongside) & another trajectory's actions & +239.5 [+133.3, +439.8] & +358.2 [+220.9, +618.1] & +319.0 [+189.0, +556.8] & +218.4 [+154.0, +311.4] \\
Arm B, 10,000 (alongside) & training mean & +161.1 [+99.6, +272.2] & +295.7 [+193.3, +477.5] & +290.9 [+178.3, +485.9] & +242.2 [+172.3, +338.6] \\
Arm B, 10,000 (alongside) & noise, k = 0.1 & +2.6 [+1.3, +4.9] & +2.1 [+1.0, +4.3] & +2.5 [+1.1, +5.2] & +2.3 [+0.1, +4.7] \\
Arm B, 10,000 (alongside) & noise, k = 0.5 & +37.3 [+20.3, +69.3] & +30.8 [+18.0, +55.9] & +36.9 [+19.7, +69.5] & +34.7 [+21.4, +54.5] \\
released checkpoint (alongside) & stale action & +55.0 [+26.3, +92.7] & +54.4 [+37.5, +82.3] & +37.4 [+21.3, +65.4] & +8.2 [-10.4, +24.9] \\
released checkpoint (alongside) & another trajectory's actions & +515.3 [+295.2, +883.1] & +500.9 [+314.6, +824.2] & +363.5 [+219.0, +607.8] & +111.0 [+59.4, +192.7] \\
released checkpoint (alongside) & training mean & +296.3 [+176.8, +487.4] & +334.9 [+215.9, +538.9] & +267.3 [+168.0, +436.2] & +73.1 [+29.7, +141.7] \\
released checkpoint (alongside) & noise, k = 0.1 & +15.0 [+7.7, +27.1] & +14.7 [+8.0, +26.5] & +10.2 [+5.5, +18.2] & +2.6 [+0.5, +5.9] \\
released checkpoint (alongside) & noise, k = 0.5 & +143.0 [+80.7, +245.2] & +147.1 [+86.3, +251.7] & +109.9 [+61.2, +193.0] & +22.8 [+3.0, +51.8] \\
\bottomrule\end{tabular}}\end{center}

\textbf{Arena: each model's own training data. For Arm A and Arm B, the in-sample arena: the training episodes' 16 non-overlapping 400-step trajectories, n\_independent = 16. For the released checkpoint, all ten episodes: 20 trajectories, n\_independent = 20. Checkpoints: Arm A and Arm B at 2,500 and 10,000 iterations, three seeds each; the released checkpoint.} Each cell is E in percent with its 95\% interval; bold, the two readings. Arm B and the released checkpoint are alongside, not readings.

\begin{center}\small\resizebox{\ifdim\width>\linewidth\linewidth\else\width\fi}{!}{%
\begin{tabular}{llllll}
\toprule
model, checkpoint & actions given & h = 1 & h = 8 & h = 32 & h = 100 \\
\midrule
Arm A, 2,500 & stale action & +1.8 [-1.6, +11.6] & +3.6 [-0.6, +21.9] & +5.6 [-0.2, +30.8] & +6.9 [+2.6, +20.0] \\
Arm A, 2,500 & another trajectory's actions & +63.3 [+43.4, +125.0] & +114.3 [+54.8, +351.6] & +141.9 [+59.0, +348.3] & +154.5 [+114.0, +265.0] \\
Arm A, 2,500 & training mean & +44.4 [+36.6, +67.3] & +70.3 [+24.4, +208.9] & +88.5 [+31.5, +267.3] & +99.5 [+60.7, +218.1] \\
Arm A, 2,500 & noise, k = 0.1 & +0.6 [-0.2, +1.5] & +0.6 [+0.3, +1.5] & +0.7 [+0.2, +1.9] & +0.4 [+0.1, +1.0] \\
Arm A, 2,500 & noise, k = 0.5 & +9.3 [+5.3, +27.9] & +8.9 [+4.4, +28.3] & +11.5 [+5.1, +27.7] & +9.8 [+7.2, +17.6] \\
Arm A, 10,000 & stale action & -1.0 [-6.1, +27.7] & +4.7 [-5.0, +73.6] & +13.9 [-2.4, +104.1] & +16.6 [+1.9, +72.6] \\
Arm A, 10,000 & another trajectory's actions & +135.5 [+91.3, +448.8] & +179.4 [+78.7, +729.4] & +227.0 [+104.9, +660.2] & +242.7 [+176.8, +494.7] \\
Arm A, 10,000 & training mean & +82.4 [+57.0, +264.6] & +115.7 [+39.5, +459.8] & +152.0 [+63.9, +537.0] & +151.9 [+83.7, +411.1] \\
Arm A, 10,000 & noise, k = 0.1 & +2.0 [+0.7, +10.2] & +1.1 [+0.2, +6.4] & +2.1 [+0.9, +6.1] & +1.9 [+1.2, +4.5] \\
Arm A, 10,000 & noise, k = 0.5 & +25.6 [+9.1, +121.4] & +22.6 [+9.5, +93.3] & +35.1 [+15.8, +91.3] & +29.2 [+20.1, +63.9] \\
Arm B, 2,500 (alongside) & stale action & +1.1 [-0.8, +8.3] & +0.3 [-0.8, +4.5] & +3.4 [+0.4, +14.6] & +0.7 [-1.2, +5.4] \\
Arm B, 2,500 (alongside) & another trajectory's actions & +67.4 [+30.6, +124.8] & +77.3 [+41.9, +187.0] & +64.6 [+31.4, +140.4] & +43.2 [+20.6, +131.4] \\
Arm B, 2,500 (alongside) & training mean & +34.8 [+20.8, +57.8] & +39.9 [+15.8, +100.8] & +43.4 [+14.5, +149.5] & +43.3 [+20.5, +127.4] \\
Arm B, 2,500 (alongside) & noise, k = 0.1 & +0.6 [+0.2, +1.6] & +0.2 [-0.1, +0.8] & +0.0 [-0.5, +0.5] & -0.7 [-1.1, -0.2] \\
Arm B, 2,500 (alongside) & noise, k = 0.5 & +8.4 [+4.8, +22.2] & +4.4 [+2.6, +11.9] & +6.2 [+0.3, +11.3] & -1.8 [-3.5, +4.5] \\
Arm B, 10,000 (alongside) & stale action & -1.0 [-2.7, +9.3] & +1.4 [-0.5, +12.4] & +6.1 [+2.5, +25.5] & +6.2 [+3.8, +17.3] \\
Arm B, 10,000 (alongside) & another trajectory's actions & +79.7 [+56.8, +218.8] & +113.2 [+36.4, +405.8] & +83.6 [+27.8, +247.9] & +61.4 [+13.1, +215.3] \\
Arm B, 10,000 (alongside) & training mean & +51.8 [+32.2, +134.0] & +65.1 [+11.7, +215.3] & +64.6 [+17.1, +239.8] & +51.1 [+13.5, +212.4] \\
Arm B, 10,000 (alongside) & noise, k = 0.1 & +0.6 [-0.1, +4.3] & +0.4 [+0.0, +1.9] & -1.0 [-1.8, +0.1] & +2.0 [-2.0, +6.1] \\
Arm B, 10,000 (alongside) & noise, k = 0.5 & +11.3 [+3.6, +56.9] & +9.5 [+4.1, +32.0] & +8.6 [+2.6, +25.6] & +5.3 [-2.8, +31.0] \\
released checkpoint (alongside) & stale action & +34.2 [+10.6, +75.0] & +37.2 [+7.9, +95.6] & +47.7 [+5.8, +131.8] & +22.5 [+6.3, +41.7] \\
released checkpoint (alongside) & another trajectory's actions & +940.5 [+707.2, +1183.7] & +1196.9 [+1042.5, +1459.9] & +807.4 [+696.6, +868.5] & +355.7 [+218.5, +487.3] \\
released checkpoint (alongside) & training mean & +600.7 [+447.0, +710.2] & +784.0 [+704.1, +924.9] & +530.9 [+420.1, +752.8] & +302.6 [+176.3, +452.5] \\
released checkpoint (alongside) & noise, k = 0.1 & +44.0 [+30.6, +56.3] & +56.9 [+45.8, +76.5] & +29.7 [+25.1, +37.4] & +13.9 [+6.7, +23.0] \\
released checkpoint (alongside) & noise, k = 0.5 & +377.7 [+299.5, +445.6] & +483.1 [+428.7, +586.0] & +295.7 [+265.0, +351.2] & +116.1 [+79.6, +182.6] \\
\bottomrule\end{tabular}}\end{center}

\textbf{Arena: the held-out pair (episodes 1 and 8, 4 non-overlapping 400-step trajectories, n\_independent = 4), out-of-sample for our arms and possibly in-sample for the released checkpoint (\S{}3.2). Checkpoints as in the first table.} No reading is made on this arena; every row is alongside.

\begin{center}\small\resizebox{\ifdim\width>\linewidth\linewidth\else\width\fi}{!}{%
\begin{tabular}{lllllll}
\toprule
model, checkpoint & arena & stale action & another trajectory's actions & training mean & noise, k = 0.1 & noise, k = 0.5 \\
\midrule
Arm A, 2,500 & own training data & 0.139 & 0.525 & 0.417 & 0.014 & 0.071 \\
Arm A, 2,500 & held-out pair & 0.060 & 0.623 & 0.431 & 0.026 & 0.128 \\
Arm A, 10,000 & own training data & 0.200 & 0.654 & 0.528 & 0.023 & 0.110 \\
Arm A, 10,000 & held-out pair & 0.084 & 0.801 & 0.584 & 0.044 & 0.215 \\
Arm B, 2,500 (alongside) & own training data & 0.109 & 0.386 & 0.303 & 0.011 & 0.055 \\
Arm B, 2,500 (alongside) & held-out pair & 0.040 & 0.481 & 0.336 & 0.021 & 0.098 \\
Arm B, 10,000 (alongside) & own training data & 0.161 & 0.545 & 0.465 & 0.018 & 0.088 \\
Arm B, 10,000 (alongside) & held-out pair & 0.052 & 0.629 & 0.451 & 0.030 & 0.140 \\
released checkpoint (alongside) & own training data & 0.204 & 0.930 & 0.666 & 0.066 & 0.327 \\
released checkpoint (alongside) & held-out pair & 0.078 & 1.042 & 0.712 & 0.093 & 0.461 \\
\bottomrule\end{tabular}}\end{center}

\textbf{$\Delta$ at h = 8, the relative-L1 of the forecast made with the changed actions measured against the forecast made with the true ones (\texttt{rollout\_eval.relative\_error}), which is zero for a model that ignores its actions. Arenas: each model's own training data (16 trajectories for our arms, 20 for the released checkpoint) and the held-out pair (4). Checkpoints: Arm A and Arm B at 2,500 and 10,000 iterations, three seeds each; the released checkpoint. Descriptive, no reading.} How far the interventions move the actions: on the forecast rows, the action changes from one step to the next on 100\% of steps on the in-sample arena, 100\% on the held-out pair and 100\% over all ten episodes. A one-step shift moves the action by 0.32, 0.27 and 0.31 of its spread (the mean absolute change from one step to the next over the mean absolute deviation from the action's mean), and the swap by 1.44, 1.43 and 1.42 (\texttt{results/\allowbreak{}action\_\allowbreak{}sensitivity\allowbreak{}.json}).

\textbf{The stale action at every horizon (post hoc).} Section 6's figures for our models come from this table, computed after rule X2's readings with its machinery, imported (\texttt{results/\allowbreak{}n5\_\allowbreak{}stale\_\allowbreak{}armA\_\allowbreak{}ci\allowbreak{}.json}). Each cell is E under the stale action, in percent, with its 95\% interval; no cell is a reading.

\begin{center}\small\resizebox{\ifdim\width>\linewidth\linewidth\else\width\fi}{!}{%
\begin{tabular}{llllllll}
\toprule
model, checkpoint & arena (trajectories) & h = 1 & h = 8 & h = 32 & h = 100 & h = 128 & h = 368 \\
\midrule
Arm A, 2,500 & own training data (16) & +27.8 [+15.6, +49.4] & +31.5 [+19.2, +53.0] & +53.8 [+34.6, +83.5] & +48.7 [+34.8, +65.7] & +46.3 [+33.7, +62.0] & +38.2 [+27.4, +51.1] \\
Arm A, 2,500 & held-out pair (4) & +1.8 [-1.6, +11.6] & +3.6 [-0.6, +21.9] & +5.6 [-0.2, +30.8] & +6.9 [+2.6, +20.0] & +8.0 [+3.0, +19.9] & +7.9 [+1.4, +19.1] \\
Arm A, 10,000 & own training data (16) & +124.6 [+75.8, +201.9] & +161.8 [+121.3, +231.1] & +205.9 [+156.7, +275.9] & +209.3 [+167.3, +257.6] & +205.0 [+166.2, +247.5] & +184.1 [+150.5, +218.6] \\
Arm A, 10,000 & held-out pair (4) & -1.0 [-6.1, +27.7] & +4.7 [-5.0, +73.6] & +13.9 [-2.4, +104.1] & +16.6 [+1.9, +72.6] & +18.0 [+3.1, +71.8] & +22.7 [+4.0, +78.5] \\
Arm B, 2,500 & own training data (16) & +18.7 [+8.9, +36.1] & +15.9 [+7.9, +29.5] & +20.3 [+10.3, +36.8] & +9.3 [+5.1, +14.9] & +6.6 [+3.2, +11.2] & -1.1 [-7.2, +5.3] \\
Arm B, 2,500 & held-out pair (4) & +1.1 [-0.8, +8.3] & +0.3 [-0.8, +4.5] & +3.4 [+0.4, +14.6] & +0.7 [-1.2, +5.4] & +0.6 [-2.3, +5.3] & -7.4 [-8.5, -3.8] \\
Arm B, 10,000 & own training data (16) & +75.1 [+44.2, +118.9] & +68.6 [+45.9, +105.2] & +60.0 [+39.9, +98.7] & +32.5 [+20.1, +50.3] & +23.6 [+13.1, +37.5] & +10.8 [+2.0, +19.1] \\
Arm B, 10,000 & held-out pair (4) & -1.0 [-2.7, +9.3] & +1.4 [-0.5, +12.4] & +6.1 [+2.5, +25.5] & +6.2 [+3.8, +17.3] & +4.7 [+3.5, +10.0] & +19.0 [-3.1, +29.1] \\
released checkpoint & own training data (20) & +55.0 [+26.3, +92.7] & +54.4 [+37.5, +82.3] & +37.4 [+21.3, +65.4] & +8.2 [-10.4, +24.9] & -8.8 [-23.8, +4.9] & -4.6 [-13.4, +3.2] \\
released checkpoint & held-out pair (4) & +34.2 [+10.6, +75.0] & +37.2 [+7.9, +95.6] & +47.7 [+5.8, +131.8] & +22.5 [+6.3, +41.7] & +20.3 [+7.0, +35.7] & +7.9 [+3.1, +13.0] \\
\bottomrule\end{tabular}}\end{center}

\textbf{Arenas: each model's own training data (the in-sample arena's 16 trajectories for our arms, all ten episodes' 20 for the released checkpoint) and the held-out pair (4), n\_independent equal to the trajectory count. Checkpoints: Arm A and Arm B at 2,500 and 10,000 iterations, three seeds each, E the mean of each seed's; the released checkpoint.} The cells at h $\le$ 100 are the stale-action rows of the first two tables.

\textbf{The design.} We trained a contaminated arm on 7,882 windows, the clean 7,687 plus 195 splices, and, because that confounds \emph{content} with \emph{count}, a duplication control adding the same 195 windows as exact copies of windows already present.

The arm's contamination rate is 2.47\%, against the reference pipeline's 3.53\%. It is deliberately lower: we splice only the 5 boundaries whose \emph{both} sides are training episodes, because 4 of the 9 between the ten full episodes put held-out rows into training, a leakage problem rather than a physics one that would have invalidated our own comparison. So this experiment measures the cost of training on physically impossible transitions, not the reference's full exposure.

Training loss over the final 250 iterations: duplication costs 0.90\%, splicing costs 21.57\%. The bootstrap interval on duplicated $-$ clean is [-0.0061, 0.0215], including zero, so the rise is caused by splice content, not by dataset size.

In rollout, across 32 cells (the held-out pair and the training episodes, 200 and 400-step trajectories, the 500 and 2,500-iteration checkpoints, n\_independent 4 to 39, at two horizons on two metrics), contamination hurts in \textbf{0} of 32 and helps in 9 (Figure 8a). The control is inert, differing from clean in 2 cells.

\begin{figure}[!ht]\centering\includegraphics[width=\linewidth]{\detokenize{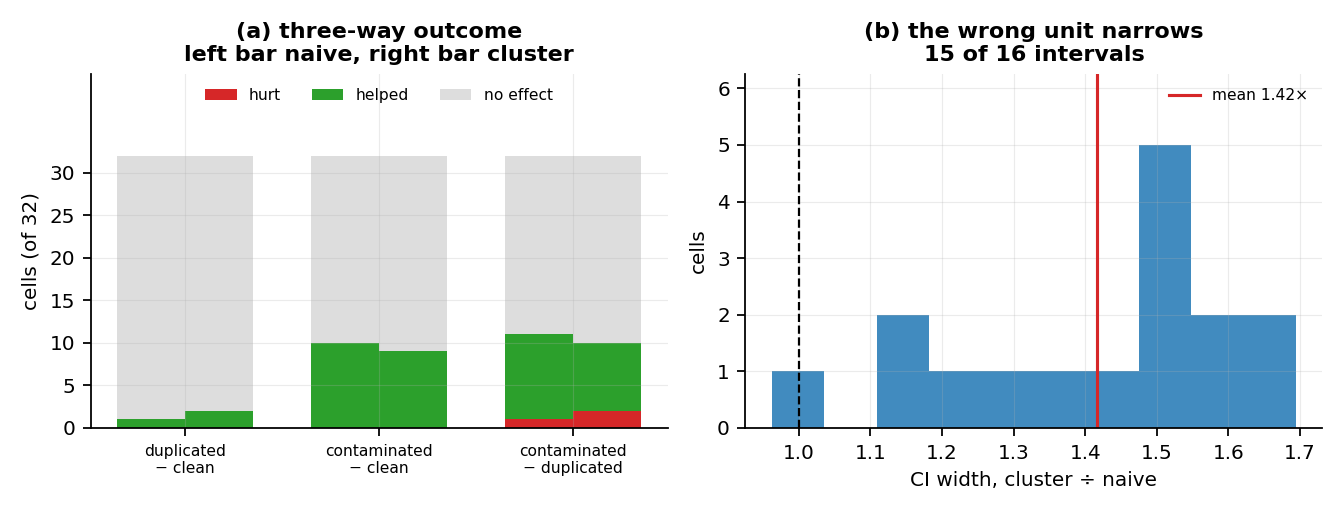}}\caption{The contamination control. (a) outcome across 32 cells for each arm pair (the held-out pair and the training episodes, 200 and 400-step trajectories, the 500 and 2,500-iteration checkpoints, n\_independent 4 to 39), naive bootstrap on the left of each position and cluster bootstrap on the right; the duplication control is inert. (b) a different family, the 16 cells of \S5's A/B gap at the 500 and 2,500-iteration checkpoints (\S3.4): distribution of the ratio of cluster to naive confidence-interval width, with the mean marked. Resampling pooled seed × trajectory values rather than whole trajectories narrows 15 of the 16 intervals.}\end{figure}

\textbf{So "costs nothing" is too strong.} Splicing raises training loss by 21.57\% against duplication's 0.90\%, and improves rollout in 9 of 32 cells: measured effects in opposite directions, the signature of regularisation. The spliced windows contain transitions the model cannot fit, it fits the rest less tightly as a result, and it rolls out slightly better. The defensible statement is that \textbf{at this rate the splices do not harm rollout and appear to help slightly. The unmarked boundaries remain a real defect on leakage grounds; the physically-impossible-transition component costs nothing detectable at this rate.}

\textbf{Pipeline rows.} Code facts are cited by file and line at the pinned upstream commits.

\begingroup\small
\begin{longtable}{>{\raggedright\arraybackslash}p{\dimexpr 0.5454\linewidth-2\tabcolsep\relax}>{\raggedright\arraybackslash}p{\dimexpr 0.1213\linewidth-2\tabcolsep\relax}>{\raggedright\arraybackslash}p{\dimexpr 0.3333\linewidth-2\tabcolsep\relax}}
\toprule finding & evidence (ledger ID, artifact) & bearing on the paper \\ \midrule\endfirsthead
\toprule finding & evidence (ledger ID, artifact) & bearing on the paper \\ \midrule\endhead
\bottomrule\endlastfoot
The reset guard tests index \emph{values} for truthiness, so a reset at the first row would be missed & B-02; \texttt{train\allowbreak{}.py:143} & Latent twice over: this dataset has no reset at its first row and marks none at all (\S{}6) \\
The train/\allowbreak{}test split draws windows at random, and adjacent windows share all but one of their 40 rows, so the released test loss is not held out & B-03; \texttt{model\_\allowbreak{}training\allowbreak{}.py:33} & The held-out arena (\S{}3) is built from whole episodes and does not use this split \\
\end{longtable}\endgroup

\section{Paper--code differences}

Where the released code departs from the papers' descriptions (\S{}5.1, \S{}5.3, \S{}6), each row checked against the code at the pinned commits.

The ledger tags some of its entries as contributions of this project. These are the confirmed ones the body does not state, or uses without stating the finding itself, each checked against its ledger entry. Code facts are cited by file and line at the pinned upstream commits; every measurement is read from the artifact named. Measurements are at 2,500 training iterations unless a row says otherwise or concerns the released checkpoint, and each row names its arena.

\begingroup\small
\begin{longtable}{>{\raggedright\arraybackslash}p{\dimexpr 0.5000\linewidth-2\tabcolsep\relax}>{\raggedright\arraybackslash}p{\dimexpr 0.1712\linewidth-2\tabcolsep\relax}>{\raggedright\arraybackslash}p{\dimexpr 0.3288\linewidth-2\tabcolsep\relax}}
\toprule finding & evidence (ledger ID, artifact) & bearing on the paper \\ \midrule\endfirsthead
\toprule finding & evidence (ledger ID, artifact) & bearing on the paper \\ \midrule\endhead
\bottomrule\endlastfoot
The papers name a forecast decay factor and set it to one in every version (RWM Table S9; RWM-U Table S8); the training loss averages every forecast step with equal weight, which is that setting & C-17 (which corrects C-09); \texttt{system\_\allowbreak{}dynamics\allowbreak{}.py:186-231} and both config files & A difference of naming, not of behaviour; our training does the same \\
The auxiliary heads (contact and termination) are trained on true next states, the state head on its own samples & M-13; \texttt{system\_\allowbreak{}dynamics\allowbreak{}.py:217, 261} & In a rollout the auxiliary heads read predicted states they never trained on; this paper compares states only (\S{}4.3) and does not measure them \\
Training feeds back a reparameterised sample, inference the mean & C-05; \texttt{system\_\allowbreak{}dynamics\allowbreak{}.py:115, 217} & Our arms train and roll out the same way; for the released checkpoint, whose per-member $\sigma$ has collapsed (\S{}5.5), the sample is numerically the mean \\
Actions enter the network raw (the config's action mean and standard deviation are zeros and ones), and no action scale is recorded in either repository; measured against joint motion, they are not joint targets in radians & C-07, D-07; \texttt{anymal\_\allowbreak{}d\_\allowbreak{}flat\_\allowbreak{}cfg\allowbreak{}.py} & A perturbation of one size means different things for actions and for states; no reported result perturbs either \\
The release's configuration, the follow-up's paper (its Table S8 in the first version, 2,500) and the checkpoint's own tag state three different training lengths; the first author believes the configuration's is a typo & C-13; \texttt{base\_\allowbreak{}cfg\allowbreak{}.py:97}, the checkpoint's \texttt{iter} field & \S{}6: none of them reaches the released variance state at a constant rate \\
\end{longtable}\endgroup

\textbf{The training tables.} RWM-U's training table fixes the iteration count, the forecast decay and the learning rate at the values our arms use (2,500 iterations), in every version; the released configuration and the checkpoint's tag state other iteration counts (\S{}6).

\textbf{Bootstrap ensembles.} RWM-U describes its members as a bootstrap ensemble. Per-head resampling exists in the online training path and not in the offline path that our arms and the released checkpoint follow (C-18).

\end{document}